\documentclass[journal]{IEEEtran}
\usepackage{graphicx}
\usepackage{times}
\usepackage{epsfig}
\usepackage{amssymb}
\usepackage{indentfirst}
\usepackage{array}
\usepackage{float}
\usepackage{multirow}
\usepackage{multicol}
\usepackage{color}
\usepackage{amsmath,bm}
\usepackage{subfig}
\definecolor{mypink}{rgb}{.99,.91,.95}
\emergencystretch=\maxdimen
\ifCLASSINFOpdf
\else
\fi
\begin{document}
%
\title{Scale-aware Pixel-wise Object Proposal Networks}
%
%
%

\author{Zequn~Jie, Xiaodan~Liang, Jiashi~Feng, Wen~Feng~Lu, Eng~Hock~Francis~Tay, Shuicheng~Yan
\thanks{Z. Jie, J. Feng, W. Lu and E. Tay are with National University of Singapore, Singapore. (e-mail: jiezequn@u.nus.edu)}
\thanks{X. Liang is with the School of Computer Science at Carnegie Mellon University, USA.}
\thanks{S. Yan is with Artificial Intelligence Institute of Qihoo/360.}
}

%
%

\markboth{Journal of \LaTeX\ Class Files,~Vol.~14, No.~8, August~2015}%
{Shell \MakeLowercase{\textit{et al.}}: Bare Demo of IEEEtran.cls for IEEE Journals}
%



\maketitle

\begin{abstract}
Object proposal is essential for current state-of-the-art object detection pipelines. However, the existing proposal  methods generally fail in producing results with satisfying localization accuracy. The case is even worse for small objects which however are quite common in practice. In this paper we propose a novel Scale-aware Pixel-wise Object Proposal (SPOP) network to tackle the challenges. The SPOP network can generate  proposals with high  recall rate and average best overlap (ABO), even for small objects. In particular, in order to improve the localization accuracy, a fully convolutional network is employed which predicts locations of object proposals  for each pixel. The produced ensemble of pixel-wise object proposals enhances the chance of hitting the object significantly without incurring heavy extra computational cost. To solve the challenge of localizing objects at small scale, two localization networks which are specialized for localizing objects with different scales  are introduced, following the divide-and-conquer philosophy. Location outputs of these two networks are then adaptively combined to generate the final proposals  by a large-/small-size weighting network. Extensive evaluations on PASCAL VOC 2007 show the SPOP network is superior over the state-of-the-art models. The high-quality proposals from SPOP network also significantly improve the mean average precision (mAP) of object detection with Fast-RCNN framework. Finally,  the SPOP network (trained on PASCAL VOC) shows great generalization performance when testing it on ILSVRC 2013 validation set.
\end{abstract}


%
\IEEEpeerreviewmaketitle

\section{Introduction}
%
%
%
%
\IEEEPARstart{I}{n} recent years, object proposal has become  crucial for modern object detection methods as an important pre-processing step~\cite{girshick2014rich,he2014spatial,girshick2015fast}. It aims to identify a small number (usually at the order of hundreds or thousands) of candidate regions that possibly contain class-agnostic objects of interest  in an image. Compared with the exhaustive search scheme such as sliding windows~\cite{felzenszwalb2010object},  object proposal methods can significantly reduce the number of candidates to be examined and  benefit  object detection in following two aspects: they can reduce computation time and  allow for applying more sophisticated classifiers.

Most of existing object proposal  methods can be roughly divided into two categories: the classic low-level cues based ones and the modern convolutional neural network (CNN) based ones. The former category of methods  mainly exploit low-level image features, including edge, gradient  and saliency~\cite{cheng2014bing,zitnick2014edge,alexe2010object,uijlings2013selective,manen2013prime,zhang2011proposal} to localize  regions possibly containing objects. Typically they either follow a bottom-up paradigm \emph{e.g.}, hierarchical image segmentation~\cite{uijlings2013selective,arbelaez2014multiscale} or
examine densely distributed windows~\cite{cheng2014bing,zitnick2014edge}. However, it is difficult for them to  balance well between localization quality and computation efficiency -- they cannot provide object proposals of high quality without incurring expensive computational cost. On the other hand, CNN-based methods either directly predict the coordinates of all the objects in an image~\cite{Erhan2013Scalable} or scan the image with a fully convolutional network (FCN)~\cite{ren2015faster,jieobject} to find the regions of high objectness\footnote{``Objectness" measures membership to foreground objects \emph{vs.} background}. Although they can achieve high recall rate w.r.t.\ relatively loose overlap criteria, \emph{e.g.} intersection over union (IoU) with a threshold value of $0.5$, this type of methods usually fails to provide high recall rate under more strict criteria (\emph{e.g.} IoU $>0.7$), suggesting their poor  localization quality.

Ideally, a generic object proposal generator should offer the following desired features:   high recall rate on objects of various categories with only a few proposals, good localization quality for each specific object instance 
and high computation efficiency. In this work, we make an effort to develop the object proposal method toward these targets.

\begin{figure}
	\centering
	\includegraphics[width=1\linewidth]{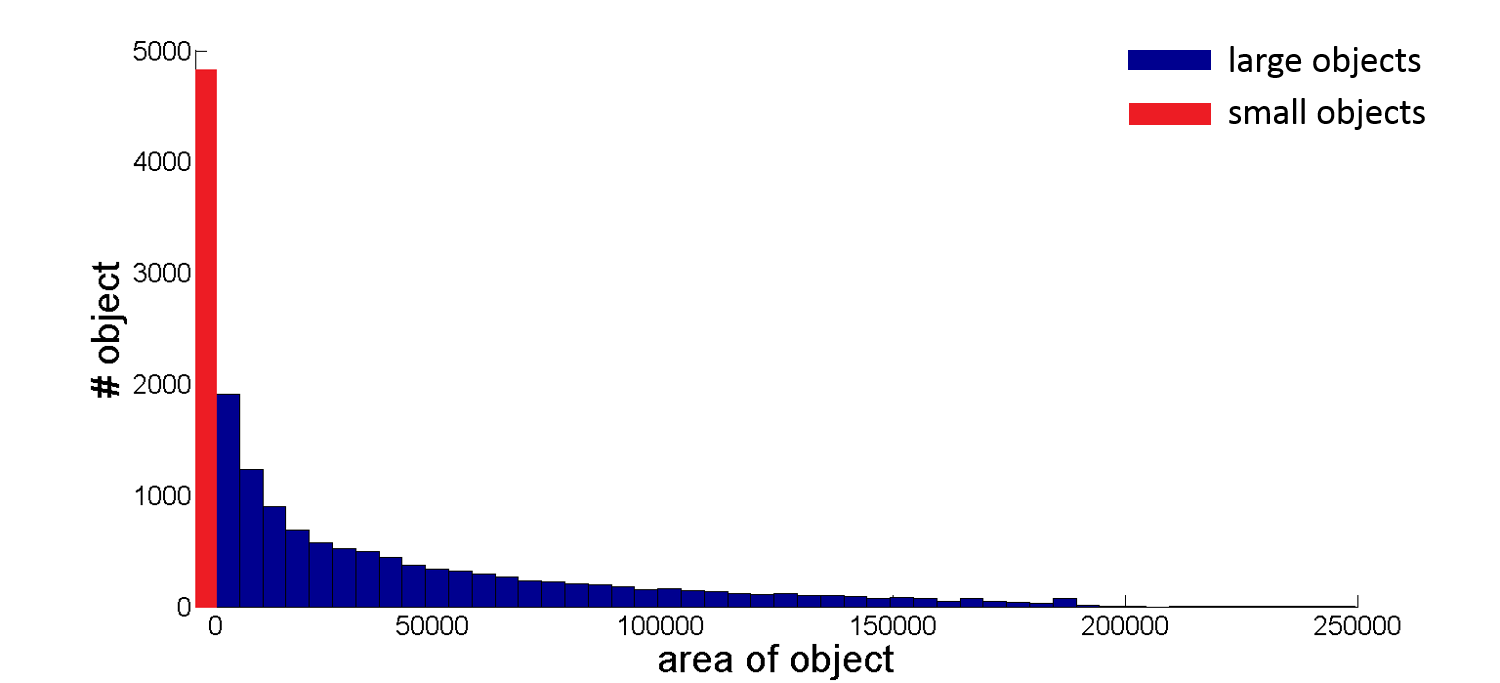}
	\caption{Distribution of objects w.r.t.\ their areas (measured by number of contained pixels) on the PASCAL VOC 2012 benchmark. It can be seen that small objects occupy a large proportion of the collection.
	}
	\label{fig:scale_stat}
\end{figure}

Our method is motivated by a statistical study on  the scale of objects in a collection of natural images. As shown in Figure~\ref{fig:scale_stat}, we plot the distribution of objects with varying scales (measured by number of pixels)   from the training and validation sets of the PASCAL VOC  detection benchmark~\cite{everingham2014pascal}. From the figure, one can  observe that the objects of small  scales (less than $2{,}000$ pixels) actually dominate the distribution. Similar observations also hold  in the ILSVRC 2013 and 2014  benchmark~\cite{russakovsky2014imagenet}.
Unfortunately, most of existing methods perform poorly in localizing  objects of such small sizes,  in terms of the best overlap\footnote{Best overlap of a particular ground-truth object is defined as the maximal intersection over union (IoU) among all the given proposals w.r.t.\ this object. Throughout the paper, Average Best Overlap (ABO) is obtained by averaging the best overlap of all the ground-truth objects}. Based on these empirical observations, we argue that  the quality of small objects localization is one main bottleneck for further improving the recall rate and average best overlap (ABO) for object proposal methods. Therefore, we focus on tackling such a challenging problem in this work.
\begin{figure}
	\centering
	\hspace{-0.3cm}
	\subfloat[Image]{\includegraphics[width=0.49   \linewidth,height=0.38  \linewidth]{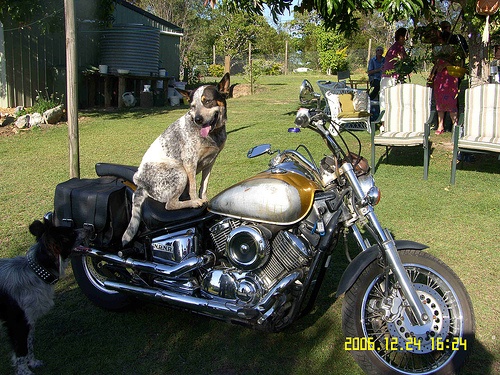}
	}
	\hspace{0cm}
	\subfloat[Objectness]{\includegraphics[width=0.49   \linewidth,height=0.38  \linewidth]{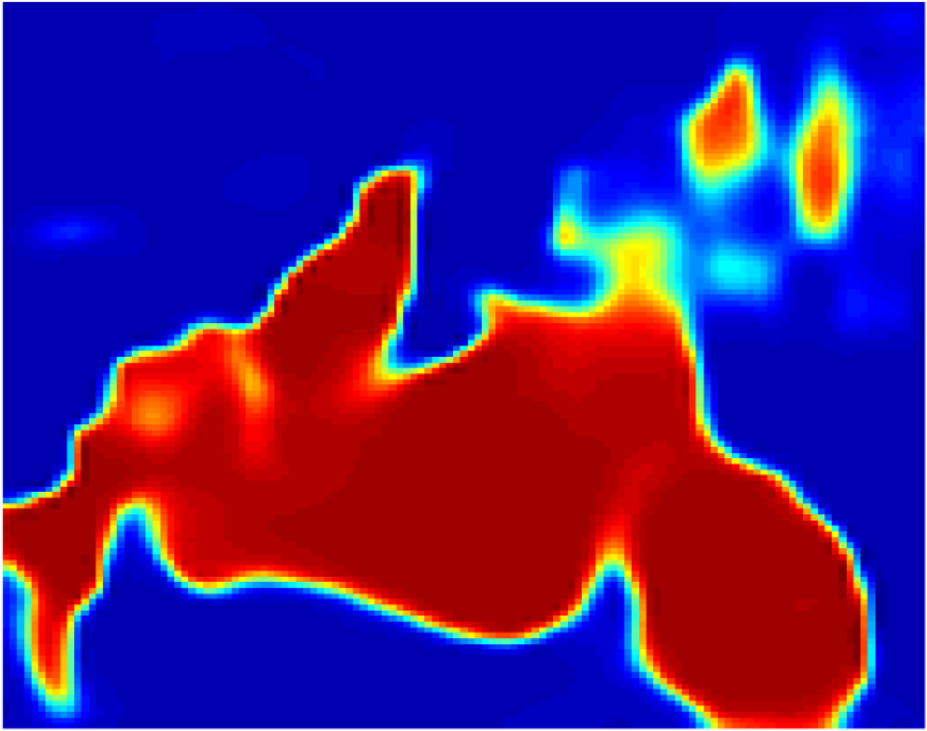}
	} 
	\\
	\hspace{-0.3cm}
    \subfloat[Offsets to object center]{\includegraphics[width=0.49   \linewidth,height=0.38  \linewidth]{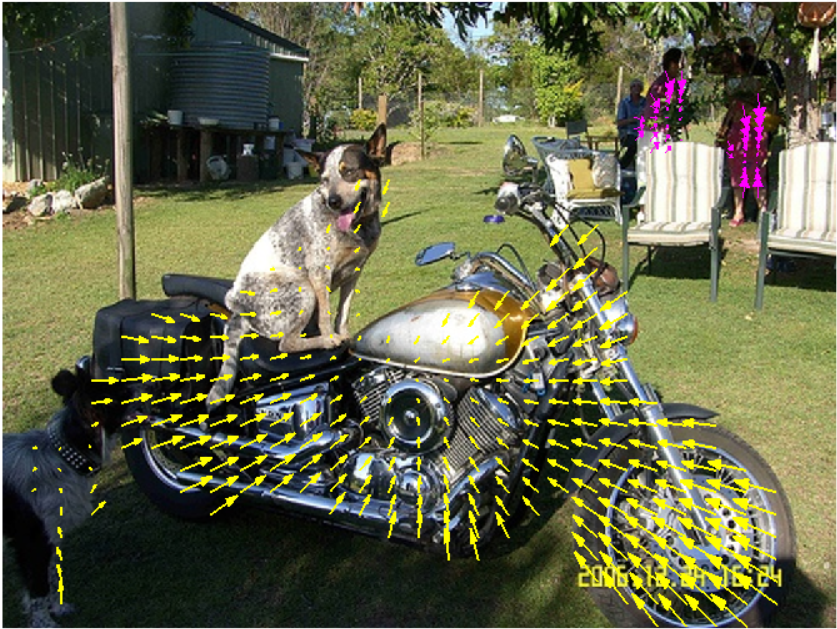}
	}
	\hspace{0cm}
	\subfloat[Object proposals]{\includegraphics[width=0.49   \linewidth,height=0.38  \linewidth]{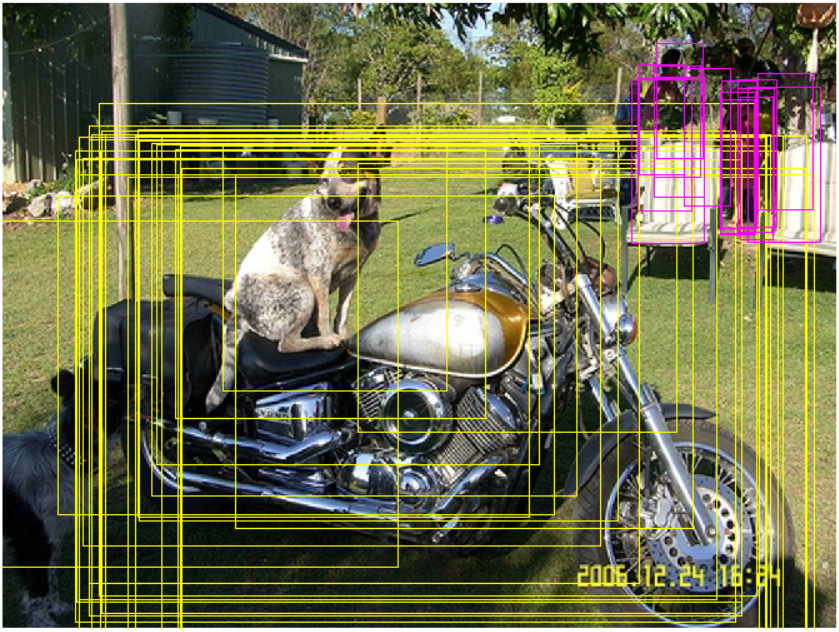}
	} 
	\\
	\caption{Examples of predicted ``objectness map" in (b), ``offsets to object center" after weighted combination in (c) and ``object proposals" in (d). ``Offsets to object center" is indicated by the arrows pointing to $((x_{\min}^{i}+x_{\max}^{i})/2-x_{i}, (y_{\min}^{i}+y_{\max}^{i})/2-y_{i})$ for each pixel $i$. Yellow and magenta colors in ``offsets to object center" and ``object proposals" indicate that the prediction is from a pixel with large-size confidence higher than $0.5$ or less than $0.5$. In the figure, only the predictions for the pixels with objectness higher than $0.5$ are shown.}
	\label{fig: overview}
\end{figure}

In particular, we develop a novel CNN based object proposal  method which contains a pixel-wise object proposal network, sharing the similar spirit with object segmentation networks~\cite{chen2014semantic,long2014fully,liang2015proposal}. Here the ``pixel-wise'' refers to: for \emph{every} pixel in an image, our proposed network model will predict a  bounding box of the object containing this pixel.  Such a pixel-level comprehensive object proposal strategy fully exploits  the  available annotations for object segmentation\footnote{The segmentation annotations can be readily collected from many public benchmark datasets.} and substantially improves the quality of object proposals through enhancing the opportunities of accurately hitting the ground-truth object. As the receptive field of each pixel in CNN is a local region around the pixel, directly predicting the coordinates of the bounding box is challenging due to the various spatial displacements of objects. We thus propose to predict the \emph{offset} of the bounding box  w.r.t this pixel, for each pixel.


We then take a further step to focus on enhancing the localization precision for small-scale objects. We propose  a new \emph{scale-aware} strategy for object proposal, which is inspired by the divide-and-conquer philosophy. Specifically, we train two independent networks, each of which 
predicts bounding box coordinates  for objects at different scales (small or large).  Then for each pixel, we will obtain two object proposals for choice. To adaptively fuse them, we introduce another object confidence network. The network consists of two branches -- one for predicting objectness confidence and the other one for weighting the large-/small-size\footnote{Throughout the paper, we use ``large-size network"/``small-size network" to refer to a localization network trained specifically for localizing objects of large/small sizes.} object localization networks. 
The objectness branch predicts the likelihood of each pixel coming from an object of interest, and the large-/small-size weighting branch trade-offs the contribution of the large-size and small-size networks to final prediction, by predicting the probability of the pixel  belonging to an object of a large size. In the training phase, the size of an object can be easily inferred from its annotated segmentation mask, which is used for training the proposed network.
 For a new image without annotation,  both the large-size and small-size object localization networks will predict the bounding box coordinates which are  combined according to the weights from the confidence network. An overview of the proposed network model is presented in Figure~\ref{fig: overview}.
 
 Therefore, the scale-aware coordinates prediction can achieve outperforming localization quality for a wide range of object sizes as for various object sizes, the final result can always considers and fuses the bounding boxes predicted by two localization networks robustly based on a reliable large-/small-size weighting mechanism.
 
To further improve the performance of localizing small objects, we employ a multi-scale  strategy for object proposal on a new image. This is inspired by the observation that by enlarging the challenging small object into a larger one, the coordinates prediction error of the small object will be scaled down, as in the case of zooming in on a small object to obtain a clearer view for humans or cameras.  Finally, a superpixel based bounding box refinement operation is applied to fine tune the proposals.

 
In short, we make the following contributions to object proposal generation. Firstly, we introduce a segmentation-like pixel-wise localization network to densely predict the object coordinates for each pixel.
Secondly, we develop a scale-aware object localization strategy which combines the predictions from a large-size and a small-size network with a weighting mechanism to boost the coordinates prediction accuracy for a wide range of object sizes. Thirdly, we conduct extensive experiments on the PASCAL VOC 2007 and ILSVRC 2013 datasets. The results demonstrate that our proposed approach outperforms the state-of-the-art methods by a significant margin, verifying the  superiority of the proposed scale-aware pixel-wise object proposal network.
 
 The remainder of this paper is organized as follows. In Section \ref{sec:related_works}, we review the related works on object proposal generation. In Section \ref{sec:network}, we describe our scale-aware pixel-wise localization network. After showing the experimental results in Section \ref{sec:experiment}, we draw the conclusion in Section \ref{sec:conclusion}.

\section{Related Work}
\label{sec:related_works}
The existing object proposal generation methods can be classified into three types: \emph{window scoring methods}, \emph{segment grouping methods} and \emph{CNN-based methods}.

\textbf{Window scoring methods} design different scoring strategies to predict the confidence of containing an object of interest for each candidate window. Generally, this type of methods first initializes a set of candidate window regions across scales and positions in an image, and then sorts them with a scoring model and selects the top ranked windows as proposals. Objectness~\cite{alexe2012measuring} selects the initial proposals from the salient regions in an image and sorts them based on multiple low-level cues, such as color, edges, location size, etc. \cite{zhang2011proposal} proposed a cascade of SVMs trained on gradient features to estimate the objectness. BING~\cite{cheng2014bing} trains a simple linear SVM on image gradients and applies it in a sliding window scheme to find the highest scored windows as object proposals. Edge Boxes~\cite{zitnick2014edge} is also performed in a sliding window manner, but relies on a carefully hand-designed scoring model which sums the edge strengths fully inside the window.  {Window scoring methods} are usually computationally efficient as they do not output segmentation masks for the proposals. However, it seems difficult for them to achieve high recall rate under  high overlap criteria (\emph{e.g.} IoU $>0.7$), which suggests the poor localization quality. This can probably be attributed to the discrete  sampling of the sliding windows which are all in the pre-defined scales and positions.


\begin{table*}[htbp]
	\centering
	\caption{Details of DeepLab-LargeFOV network architecture.}
	\begin{tabular}{rrrrrrrrr}
		\hline
		layer & \#channel & kernel size & stride & zero-padding size & hole size & training map size & receptive field size & \#weight \\
		\hline
		input image & 3     & -     & -     & -     & -     & 513*513 & 435*435 & - \\
		conv1\_1 & 64    & 3*3   & 1*1   & 1*1   & -     & 513*513 & 433*433 & 1.75K \\
		conv1\_2 & 64    & 3*3   & 1*1   & 1*1   & -     & 513*513 & 431*431 & 36K \\
		pool1 & 64    & 3*3   & 2*2   & 1*1   & -     & 257*257 & 215*215 & - \\
		conv2\_1 & 128   & 3*3   & 1*1   & 1*1   & -     & 257*257 & 213*213 & 72K \\
		conv2\_2 & 128   & 3*3   & 1*1   & 1*1   & -     & 257*257 & 211*211 & 144K \\
		pool2 & 128   & 3*3   & 2*2   & 1*1   & -     & 129*129 & 105*105 & - \\
		conv3\_1 & 256   & 3*3   & 1*1   & 1*1   & -     & 129*129 & 103*103 & 288K \\
		conv3\_2 & 256   & 3*3   & 1*1   & 1*1   & -     & 129*129 & 101*101 & 576K \\
		conv3\_3 & 256   & 3*3   & 1*1   & 1*1   & -     & 129*129 & 99*99 & 576K \\
		pool3 & 256   & 3*3   & 2*2   & 1*1   & -     & 65*65 & 49*49 & - \\
		conv4\_1 & 512   & 3*3   & 1*1   & 1*1   & -     & 65*65 & 47*47 & 1.13M \\
		conv4\_2 & 512   & 3*3   & 1*1   & 1*1   & -     & 65*65 & 45*45 & 2.25M \\
		conv4\_3 & 512   & 3*3   & 1*1   & 1*1   & -     & 65*65 & 43*43 & 2.25M \\
		pool4 & 512   & 3*3   & 1*1   & 1*1   & -     & 65*65 & 41*41 & - \\
		conv5\_1 & 512   & 3*3   & 1*1   & 2*2   & 2*2   & 65*65 & 37*37 & 2.25M \\
		conv5\_2 & 512   & 3*3   & 1*1   & 2*2   & 2*2   & 65*65 & 33*33 & 2.25M \\
		conv5\_3 & 512   & 3*3   & 1*1   & 2*2   & 2*2   & 65*65 & 29*29 & 2.25M \\
		pool5 & 512   & 3*3   & 1*1   & 1*1   & -     & 65*65 & 27*27 & - \\
		pool5a & 512   & 3*3   & 1*1   & 1*1   & -     & 65*65 & 25*25 & - \\
		fc6   & 1024  & 3*3   & 1*1   & 12*12 & 12*12 & 65*65 & 1*1   & 4.5M \\
		fc7   & 1024  & 1*1   & 1*1   & -     & -     & 65*65 & 1*1   & 1M \\
		\hline
	\end{tabular}%
	\label{tab:archi}%
\end{table*}%

\textbf{Segment grouping methods} are usually initialized with an oversegmentation to obtain superpixels for an image. Then different merging strategies are adopted to group the similar segments hierarchically to generate the object proposals of all scales. Generally, they follow a bottom-up scheme which relies on diverse low-level image cues including color, shape and texture. For example, Selective Search~\cite{uijlings2013selective} iteratively merges the most similar segments to form proposals based on several low-level cues. Randomized Prim~\cite{manen2013prime} learns a randomized merging strategy based on the superpixel connectivity graph.  Multiscale Combinatorial Grouping (MCG)~\cite{arbelaez2014multiscale} utilizes multi-scale hierarchical segmentations based on the edge strength and the obtained proposals are then ranked using features including size, location, shape and contour. Geodesic object proposal~\cite{krahenbuhl2014geodesic} also depends on superpixels as initialization, and then computes a geodesic distance transform and selects certain level sets of the distance transform as object proposals.~\cite{kk-lpo-15} proposes learning conditional random field (CRF) in multiscales to classify the superpixels into objects or background.  Generally, compared with \emph{window scoring methods}, \emph{segment grouping methods} achieve more consistent and acceptable recall under both loose and strict overlap criteria, indicating a better localization ability. Nevertheless, these methods produce high quality proposals often by multiple segmentations in different scales and color spaces, thus are quite computationally expensive and time-consuming.

\textbf{CNN-based methods} follow the great success of Convolutional Neural Network in other vision tasks,~\cite{krizhevsky2012imagenet,wei2015hcp,szegedy2014going,liang2015towards}, especially semantic segmentation~\cite{wei2015stc,wei2016learning,liang2015reversible}. They leverage the powerful discrimination ability of Convolutional Neural Network (CNN) to extract visual features as inputs of other techniques to produce proposals or directly regress the coordinates of all the object bounding boxes in an image. MultiBox~\cite{Erhan2013Scalable} trains a network to directly predict a fixed number of proposals and their confidences in an image and ranks them with the obtained confidences. RPN~\cite{ren2015faster} uses a Fully Convolutional Network (FCN) to densely generate the proposals in each local patch based on several pre-defined ``anchors" in the patch. DeepProposal~\cite{ghodrati2015deepproposal} hunts for the proposals in a sliding window manner by using the CNN features from the final to the beginning layers and training a cascade of linear classifiers to obtain the highest scored windows.
Current {CNN-based methods} typically achieve high recall with only a small number (usually $<1{,}000$) of proposals, under  loose overlap criteria (\emph{e.g.} $0.5$$<$IoU$<$$0.6$). But similar to {window scoring methods}, they can hardly achieve high recall rate under more strict overlap criteria (\emph{e.g.} IoU $>0.7$). To improve the object proposal localization quality, different from them, our approach predicts the object locations in a pixel-wise manner so that we have much more chances to localize each object with high precision. This also takes the full advantage of the publicly available segmentation masks annotations. This is similar to \cite{huang2015densebox} which deals with object detection task in the object coordinates prediction part. In addition, our scale-aware prediction strategy provides adaptive accurate  prediction for both large-size and small-size objects, which also distinguishes our method from others.

\section{Scale-aware Pixel-wise Proposal Network}
\label{sec:network}
The proposed Scale-aware Pixel-wise Object Proposal Network (SPOP-net) is based on a pixel-wise segmentation-like object coordinates prediction network, and includes a scale-aware localization mechanism for predicting the coordinates of objects of different sizes. In addition, a multi-scale prediction strategy is employed during testing to boost the small objects localization. Finally, a superpixel boundary based proposal refinement is introduced to further improve the proposal precision. We will elaborate all the components of SPOP-net in this section.

\subsection{Pixel-wise Localization Network}
The proposed Scale-aware Pixel-wise Object Proposal Network (SPOP-net) takes an image of \emph{any} size as input and predicts the location of the object w.r.t.\ each pixel in the image. More concretely, for each pixel, SPOP-net predicts the normalized coordinates of the bounding box of the object that contains the pixel. The predictions from the background pixels make no sense and will be ranked behind due to low objectness scores they obtain, thus making no  difference to the recall performance of top-ranked proposals, which will be detailed later. In this subsection, we first explain the architecture of SPOP-net and then elaborate on how to train and apply the SPOP-net.  

\paragraph{Architecture} Our SPOP-net is built upon a pre-trained DeepLab-LargeFOV segmentation network \cite{chen2014semantic}. Its architecture is shown in Table \ref{tab:archi}. 
The receptive field of our localization network in the last layer is $435 \times 435$. This large receptive field enables SPOP-net to ``see'' a large region of the image in its last layer and predict the object bounding boxes effectively.

\paragraph{Training}

For each pixel, the pixel-wise localization network aims to predict the bounding box coordinates $\mathbf{t}=(x_{\min}/w, y_{\min}/h, x_{\max}/w, y_{\max}/h)$ of the object that contains this pixel. Here ($x_{\min}$, $y_{\min}$) and ($x_{\max}$, $y_{\max}$) denote the coordinates of the top-left and bottom-right corners of the object bounding box containing the pixel; $h$ and $w$ represent the height and the width of the image plane respectively. Therefore, for a single object, all the pixels inside this object are given the same ground-truth values $(x_{\min}/w, y_{\min}/h, x_{\max}/w, y_{\max}/h)$.  
 We train the pixel-wise localization network
to  minimize the following localization error  $\mathcal{L}$ 
that is proportional to the Euclidean distance  
between the predicted coordinate vector $\mathbf{t}_{i}$ and the ground-truth coordinate vector $\mathbf{t}_{i}^{*}$ for all the foreground pixels. The loss function $\mathcal{L}$ is defined as
\begin{equation}
	\mathcal{L} 
	= \sum_{i} p_{i}^{*}\|\mathbf{t}_{i}-\mathbf{t}_{i}^{*}\|^{2},
\end{equation}
where $\mathbf{t}_i$ is the predicted 4-d object coordinate vector, and $p_{i}^{*}$ is a binary variable indicating whether the pixel $i$ is a foreground one: it takes $1$ if the pixel $i$ is from a foreground object and $0$ otherwise. Such a filtered loss (through $p_{i}^{*}$) enables the localization network to concentrate on  localizing foreground objects without being distracted by background pixels in the training phase. In the practical implementation, as the final layer has smaller size than the input image, we resize the ground-truth coordinate map to the same small size as the final layer.

However, due to the possible spatial displacement (\emph{e.g.} two  exactly the same objects could appear at different locations in an image), accurately predicting the absolute object bounding box coordinates is difficult. It is because these two objects have the same visual input for the model, but their locations the model needs to learn to predict are totally different. To solve this issue, for each pixel, we change its learning targets from the absolute object bounding box coordinates to the  offsets from the pixel  to the object bounding box. E.g. for object bounding box coordinate $x_{\min}/w$, we change the target from $x_{\min}/w$ to $(x_{\min}-x_{\text{self}})/w$, here $x_{\text{self}}$ is the $x$ coordinate of the pixel itself. Changing the coordinates to offsets can be conveniently achieved  by element-wisely summing the output of the 2nd last layer and the spatial coordinate map ($x$ or $y$ values of all the pixels themselves). Then  the absolute object bounding box coordinates can be used as learning targets for the final layer.
In this way, applying the absolute coordinates learning targets  to the final layer is equivalent to applying the following object coordinate offsets to the 2nd last layer. 
	\begin{equation*}
	\left(\frac{x_{\min}-x_{\text{self}}}{w}, \frac{y_{\min}-y_{\text{self}}}{h}, \frac{x_{\max}-x_{\text{self}}}{w}, \frac{y_{\max}-y_{\text{self}}}{h}\right)
	\end{equation*}
Then we can directly obtain the absolute object proposal coordinates from the predictions of the final layer. After obtaining the output map from the final layer having a smaller size than the input image, all the subsequent procedures (\emph{e.g.} refinement, ranking and NMS) are only based on the output map of smaller size.  Because we just leverage pixel-level prediction of proposals for having higher chance to hit the ground-truth objects accurately instead of doing pixel-level classification as DeepLab. If resizing the smaller output map back into the original size, the subsequent refinement, ranking and NMS steps will bring much higher computation burden but not significant performance improvement.

\begin{figure*}
	\vspace{0cm}
	\centering
	\includegraphics[width=150mm, height=60mm]{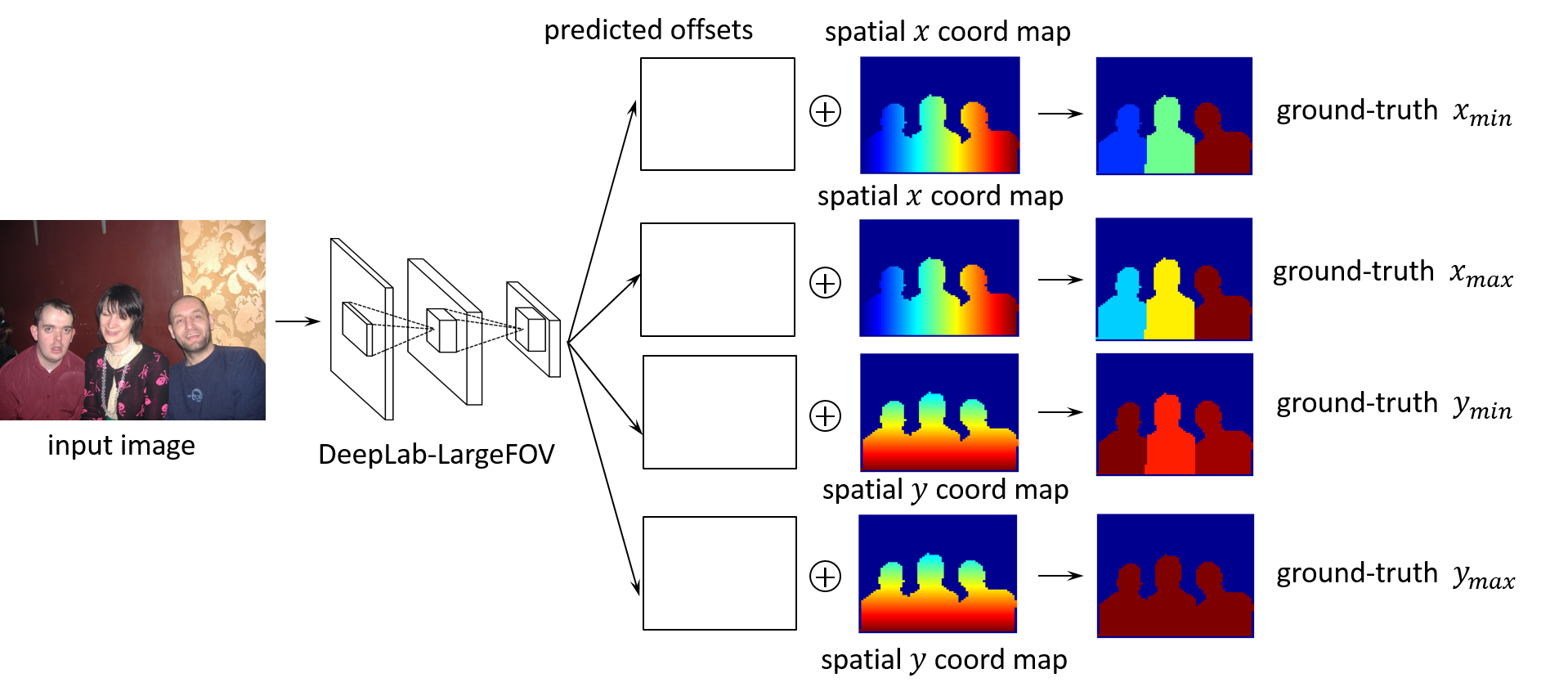}
	\caption{An image passes through several layers to obtain four maps in the second last layer. In the second last layer, two maps are element-wise summed with spatial $x$ coord map to produce the final prediction for the $x_{\min}$ and $x_{\max}$ of the corresponding objects for all the pixels, and the other two maps are element-wise summed with spatial $y$ coord map to produce the final prediction for the $y_{\min}$ and $y_{\max}$ of the corresponding objects for all the pixels. In this way, the four maps in the second last layer in our fully trained network actually predict the four offsets between each pixel position and its corresponding object location, which makes it easier for the network to predict the object coordinates in the final layer. Different colors in the ground-truth maps and spatial coord maps represent different values. Note that we only show the foreground regions of spatial $x$ and $y$ maps for better view. 
	}
	\label{fig: loc_net}
\end{figure*}

\begin{figure}
	\vspace{0cm}
	\centering
	\includegraphics[width=95mm, height=40mm]{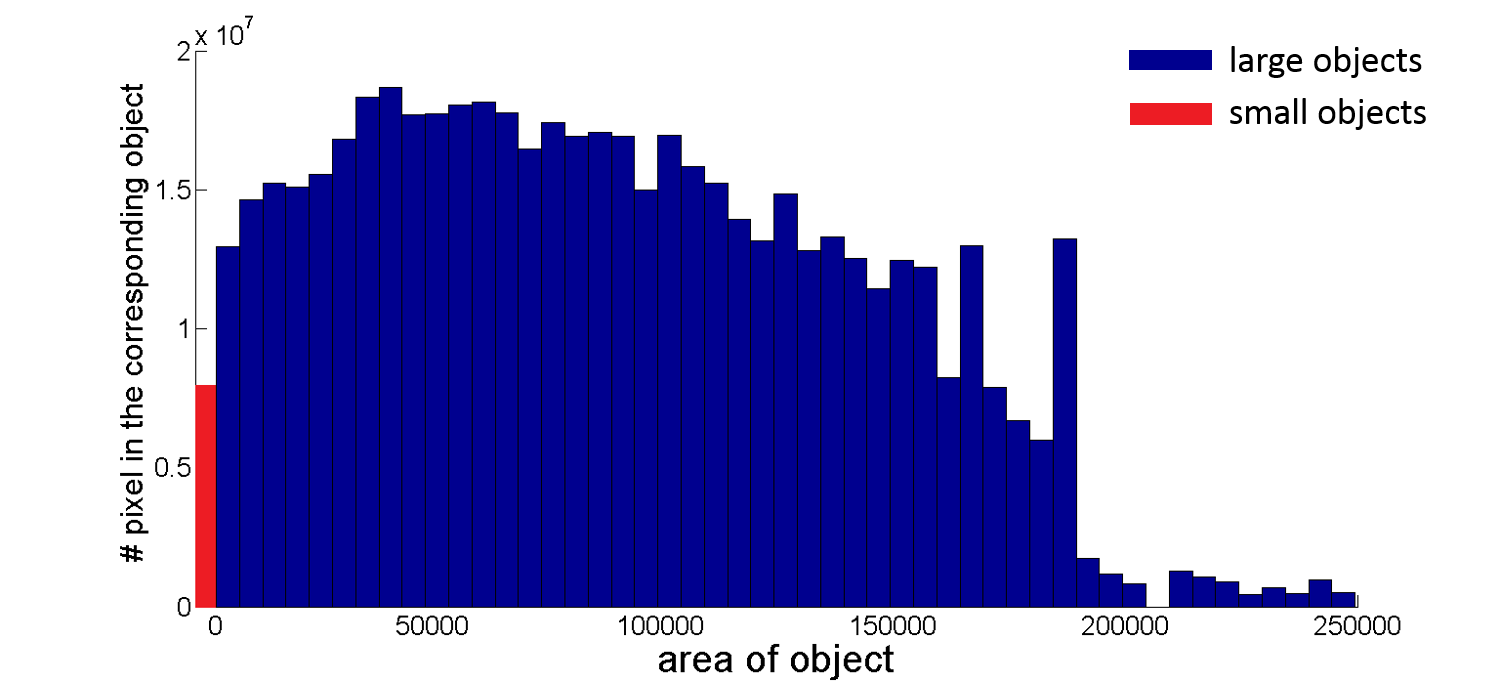}
	\caption{The distribution of all the pixels w.r.t.\ the area of the object each pixel belongs to. It is shown that although the number of small objects is large according to Figure ~\ref{fig:scale_stat}, the number of pixels belonging to small objects is still small, leading to the unbalanced pixel-level training samples.
	}
	\label{fig:pixel_distri}
\end{figure}

\subsection{Scale-aware Localization}


A fully trained pixel-wise localization network can predict the coordinates of object bounding boxes w.r.t.\ each pixel from an image. 
However, a single network model may not be able to well handle all the annotated objects that have quite diverse sizes and only offers inferior localization performance for objects of small sizes. To verify this point, we conduct the following preliminary experiments to evaluate the errors of bounding box prediction for  large and small objects, using a single pixel-wise localization network trained on the annotated objects of all sizes. The evaluation results are shown in Table \ref{tab: all}. 

From Table \ref{tab: all}, one can observe that the network trained on all the objects of different sizes produces an $L_2$ error for small objects that is about $3$ to $6$ times larger than the error for large objects. This demonstrates the poor localization ability of a single network model for small objects.

The difficulty of accurately localizing both large and small objects using a single network arguably lies in handling the  highly diverse offsets of large and small objects. Apart from this, another difficulty  comes from the extremely unbalanced training samples between the pixels from large and small objects. Such imbalance leads to the fact that training error of large objects dominates the training loss to minimize.

Also, we empirically verify the sample imbalance through statistics on the pixel-level distribution of the annotations in terms of the area of the object (see Figure~\ref{fig:pixel_distri}) since our pixel-wise localization network is trained on pixel-level annotations.
\begin{figure*}
	\hspace{1cm}
	\includegraphics[width=200mm, height=55mm]{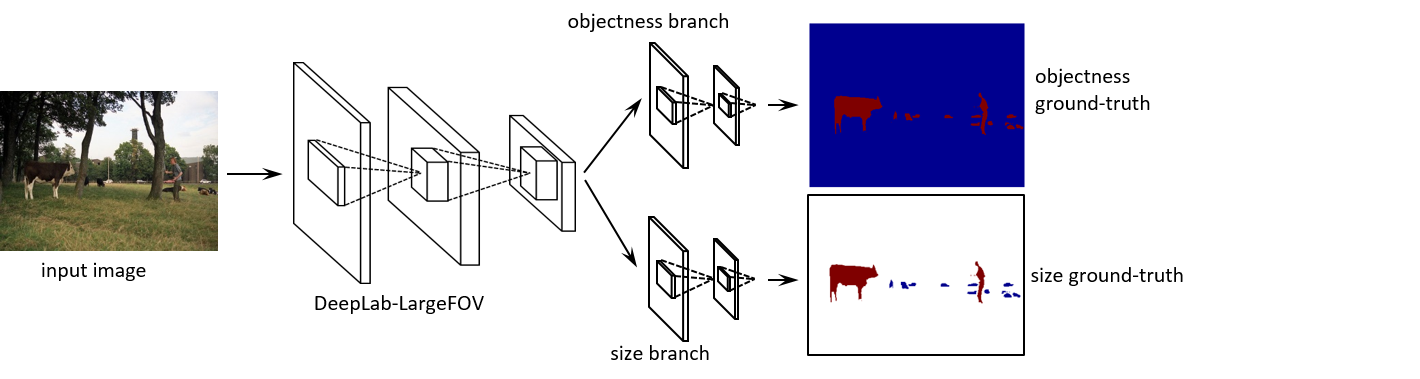}
	\caption{Illustration of the ``confidence network" which bifurcates into two branches to perform foreground/background classification and large/small object classification after all the layers of ``DeepLab-LargeFOV" network. Both the sub-networks contain two convolution layers with $3\times3$ kernel size. The first layer outputs $1{,}024$ feature maps while the second (also the last) layer produces two maps showing the final confidence of their own task. In the ground-truth map of the foreground/background classification branch, red pixels are in foreground objects while blue pixels are in background. In the ground-truth map of the large/small object classification branch, red pixels are in large objects, blue pixels are in small objects and white pixels are background pixels thus are not considered during training. 
    \label{fig:confidence_net}
	}
\end{figure*}
\begin{table}
	\caption{$L_2$ errors of normalized  coordinates prediction for both large ($\geq 2{,}000$ pixels) and small objects ($<2{,}000$ pixels) in VOC 2007 testing set, based on the network trained on the annotated objects of all sizes. }
	\centering
	\small
	\renewcommand{\arraystretch}{1.2}
	\begin{tabular}{c|c|c}
		\hline
		 & large objects  & small objects  \\
		\hline
		$x_{\min}^{err}/w$& $0.0090$ & $0.0270$ \\
		\hline
		$y_{\min}^{err}/h$& $0.0064$ & $0.0160$ \\
		\hline
		$x_{\max}^{err}/w$& $0.0080$ & $0.0412$ \\
		\hline
		$y_{\max}^{err}/h$& $0.0088$& $0.0476$ \\
		\hline		
	\end{tabular}
	\label{tab: all}
\end{table}

To improve the localization accuracy for small objects, we propose a \emph{scale-aware} localization strategy. Roughly, in the scale-aware strategy, two localization networks are trained -- which share the same architecture -- with two non-overlapped subsets of  the objects. The large-size network is only trained on the pixels belonging to large objects and the small-size network is only trained on the pixels belonging to small objects. The loss function to be optimized for the large-size and small-size network are shown in Eqn.~(2) and Eqn.~(3) below respectively:

\begin{equation}
L_{l}=\sum_{i}l_{i}^{*} \|\mathbf{t}_{i}-\mathbf{t}_{i}^{*}\|^{2}  
\end{equation}
\begin{equation}
L_{s}=\sum_{i}s_{i}^{*} \|\mathbf{t}_{i}-\mathbf{t}_{i}^{*}\|^{2} 
\end{equation}
where $l_{i}^{*}$ and $s_{i}^{*}$ are binary indicators showing whether the pixel $i$ belongs to a large object or a small object. The effectiveness of training such scale-aware networks is validated by  evaluating the $L_2$ errors of small objects location prediction with the small-size network. See Table \ref{tab: small}. During the testing phase,  the two networks work simultaneously to output their own  prediction  for an image. Then, the  predictions from two networks are combined with an adaptive weighting scheme. 

The weight is output by a network trained for classifying  large and small objects pixel-wisely and the weight  is equal to the  confidence of the pixel belonging to a large  object obtained in the last layer of the network. Such a classification network is termed as ``confidence network''. 

The  structure of the confidence network is illustrated in Figure~\ref{fig:confidence_net}. Apart from the large/small classification branch, the confidence network also outputs the objectness confidence in another branch aiming to classify all the pixels into two categories, \emph{i.e.}, foreground pixels and background pixels. 

In  the confidence network, the two branches share the convolutional features in the lower layers. The last  feature maps  shared are then fed into the two branches. The intuition for dividing the confidence network into two branches at the higher layer is that for different  tasks, the low-level features are usually common and can be shared~\cite{zeiler2014visualizing}, while the semantically high-level features  extracted by the higher layers may be totally different for different tasks. For example, the foreground/background classification task prefers the common features that are insensitive to different sizes of objects, but the large/small classification task aims to extract the discriminative features between large and small objects. The large receptive field (\emph{i.e.} $435\times435$) in the last layer of the ``confidence network" provides a sufficient large view enabling the prediction of both foreground/background and large/small classifications.

\begin{figure*}
	\vspace{0cm}
	\centering
	\includegraphics[width=160mm, height=85mm]{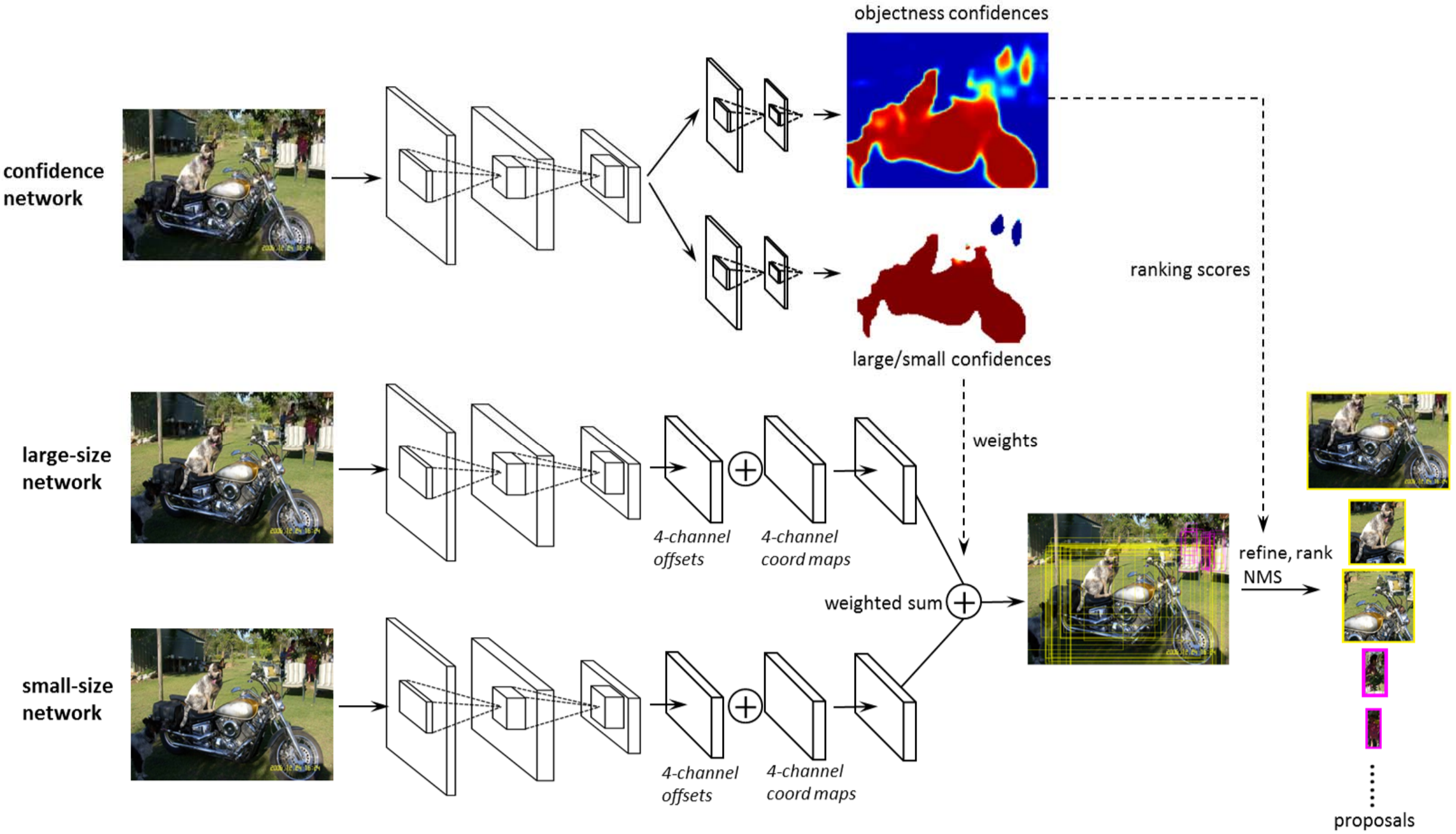}
	\caption{Overview of our approach. An image passes the confidence network to obtain the pixel-wise objectness confidences and large/small size confidences (red color represents higher values, \emph{e.g.}, high objectness and high large size confidences). The image also passes two localization networks to obtain the predicted pixel-wise large and small object coordinates $(x_{\min}, y_{\min}, x_{\max}, y_{\max})$, respectively. The feed-forward computation of the three networks are independent and can be run in parallel. Then the final predicted object coordinates are the sum of the predictions by large-/small-size networks weighted by the large/small size confidences. Using the objectness confidences as ranking scores, the final proposals are produced after refinement, ranking and NMS. For multi-scale inference, all the above procedures are run for the enlarged input image as well. Then the proposals obtained by both the original and enlarged scales are mixed in the ranking and NMS. 
	}
	\label{fig: pipeline}
\end{figure*}

\begin{table}
	\caption{$L_2$ errors of normalized coordinates prediction for small objects ($<2{,}000$ pixels) in VOC 2007 testing set, based on the network trained only on small objects.}
	\centering
	\small
	\renewcommand{\arraystretch}{1.2}
	\begin{tabular}{c|c}
		\hline
	       &  small objects \\
		\hline
		$x_{\min}^{err}/w$&  $0.00058$ \\
		\hline
		$y_{\min}^{err}/h$&  $0.00040$ \\
		\hline
		$x_{\max}^{err}/w$&  $0.00068$ \\
		\hline
		$y_{\max}^{err}/h$&  $0.00086$ \\
		\hline		
	\end{tabular}
	\label{tab: small}
\end{table}

The objective function to be optimized during training the confidence network is a multi-task cross-entropy loss:
\begin{equation}
\begin{aligned}
  L=\sum_{i}p_{i}^{*}\log(p_{i})+(1-p_{i}^{*})\log(1-p_{i})+ \\
  \sum_{i}p_{i}^{*}(z_{i}^{*}\log(z_{i})+(1-z_{i}^{*})\log(1-z_{i})).
\end{aligned}
\end{equation}
Here $p_{i}^{*}$ and $p_{i}$ are the ground-truth label of the foreground/background classification and the predicted confidence of being a foreground pixel for pixel $i$, respectively. $z_{i}^{*}$ and $z_{i}$ are the ground-truth label of the large/small object classification and the predicted confidence of being contained in a large object for pixel $i$, respectively. Note that the second term is only activated when $p_{i}^{*}$ equals 1, indicating that the pixel belongs to a foreground object. After the large object confidence $z_{i} $ for the pixel $i$ is obtained, the final predicted coordinates  of the object it belongs to are the weighted sum of the predictions by the large-size and small-size networks as follows. 
\begin{equation}
	\mathbf{t}_{i} = z_{i}\mathbf{t}_{l,i}+(1-z_{i})\mathbf{t}_{s,i}
\end{equation}
where $\mathbf{t}_{i,l}$ and $\mathbf{t}_{i,s}$ are the predictions by the large-size and the small-size network respectively. Then we treat the predicted object coordinates by each pixel as an initial proposal to be passed to the later proposal refinement and non-maximum suppression (NMS) steps to obtain the final object proposals.

\subsection{Multi-scale Inference}
To further enhance the accuracy of small objects localization, we propose to employ a multi-scale prediction strategy in the testing phase. The motivation is quite straightforward:  by enlarging the challenging small object into a larger one, the coordinates prediction error of the small object will be scaled down, which is similar to zooming in on a small object to improve the localization accuracy.   
At the enlarged scale, all the proposals in the enlarged image will be mapped back to their corresponding positions at the original scale.

Therefore, given a testing image, in addition to its original scale, we resize it into a larger scale and run the prediction process as well. Specifically, both on the original scale and the enlarged scale, we simultaneously run the two localization networks (i.e. large-size and small-size)  and the confidence network, and combine the both location predictions weighted by the large object confidence $z_{i}$ of its own scale. As all the feed-forward computation of the networks is independent and can be performed in parallel, the computation time cost can remain relatively low.

\subsection{Proposal Refinement}
We then refine the two sets of proposals obtained in both original and enlarged scales. An inherent weakness for object localization by regressing the four coordinates with CNN is that the objectness and coordinates ground-truths only permit determining the most discriminative foreground windows. Therefore, even though the windows decided by the localization networks are likely to overlap with target objects, it cannot be ensured that they are able to delineate object boundaries well.
\begin{figure}	
	\subfloat{\includegraphics[width=4.2cm,height=3.3cm]{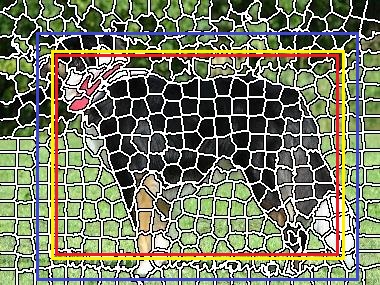}
	}  
	\subfloat{\includegraphics[width=4.2cm,height=3.3cm]{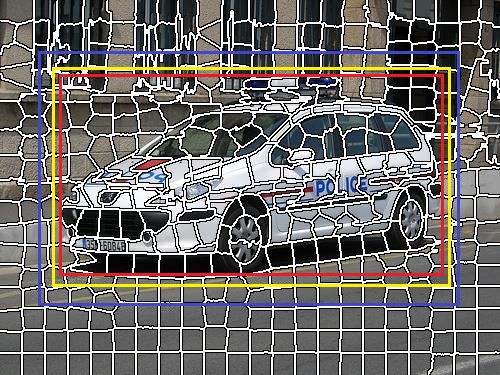}
	}
	\\
	\caption{Illustration of proposal refinement using superpixel boundary based expansion and shrinkage. Yellow boxes represent initial proposals; red boxes and blue boxes are the corresponding refined proposals after shrinkage and expansion respectively. In the left example, expansion finds a closer box to the ground-truth, but in the right example, shrinkage helps the proposal get close to the ground-truth.}  
	\label{fig:proposal_refine}
\end{figure}

To  take object boundaries into consideration, we utilize a superpixel boundary based window refinement method, similar to~\cite{Chen2015Improving}. The main idea is to expand or shrink the proposals to align the four sides of the proposals with the boundaries of the superpixels better. The reason for using superpixels is that the boundaries of superpixels are informative indicators of object boundaries and superpixels can be generated efficiently with off-the-shelf algorithms (\emph{e.g.} SLIC~\cite{achanta2012slic}). Specifically, for each proposal, we generate two versions of refined proposals, \emph{i.e.} the minimum bounding rectangle of all the superpixels entirely inside this proposal and the minimum bounding rectangle of all the superpixels entirely inside this proposal or straddling this proposal (see Figure~\ref{fig:proposal_refine}). As illustrated in Figure~\ref{fig:proposal_refine}, expansion and shrinkage offer two possible ways of getting close to the ground-truth box for the proposals with different location biases to the ground-truth. Therefore, we pass all the two versions of refined proposals as well as the initial proposals  to the later proposal ranking and NMS processing.

In the stage of proposal ranking , we sort  all the proposals (including the initial  and the two refined ones in both original and enlarged scale)  by their objectness confidence $p_{i}$. Recall $p_{i}$ is the output from foreground/background classification branch of the confidence network. For each initial proposal, its two versions of refined proposals are assigned with the same objectness confidence $p_{i}$ as itself. Finally, the standard non-maximum suppression (NMS) is employed to remove the highly overlapped redundant proposals.
 \begin{figure*}	
 	\subfloat[Recall vs \# proposal (IoU=0.5) ]{\includegraphics[width=0.248\linewidth]{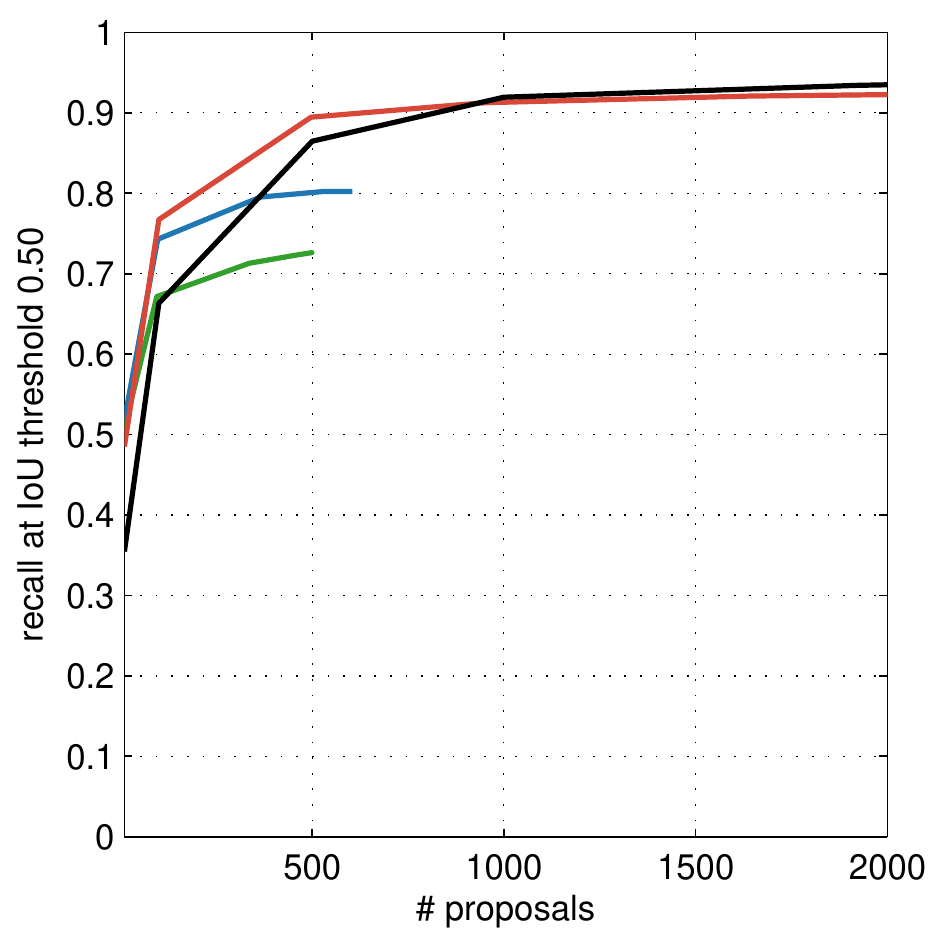}
 	}
 	\hspace{-0.3cm}
 	\subfloat[Recall vs \# proposal (IoU=0.7) ]{\includegraphics[width=0.248\linewidth]{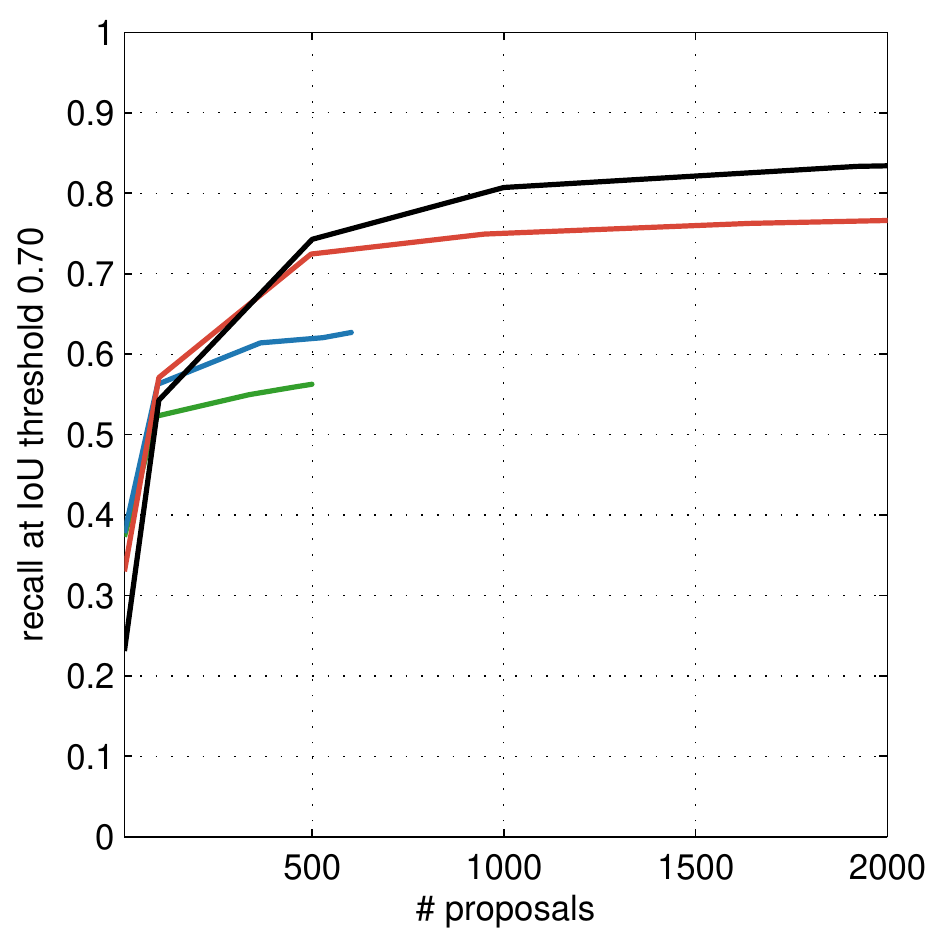}
 	} 
 	\hspace{-0.3cm}
 	\subfloat[AR vs \# proposal (0.5$<$IoU$<$1) ]{\includegraphics[width=0.248\linewidth]{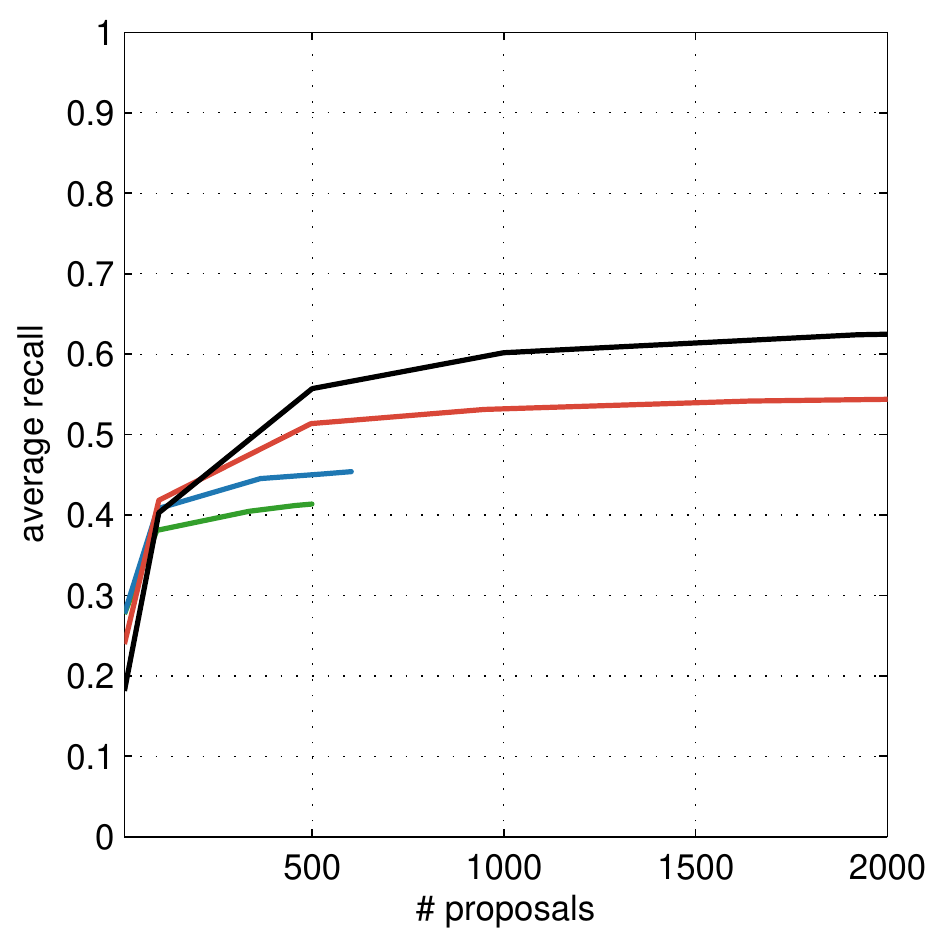}
 	}
 	\hspace{-0.3cm}
 	\subfloat[ABO vs \# proposal  ]{\includegraphics[width=0.248\linewidth]{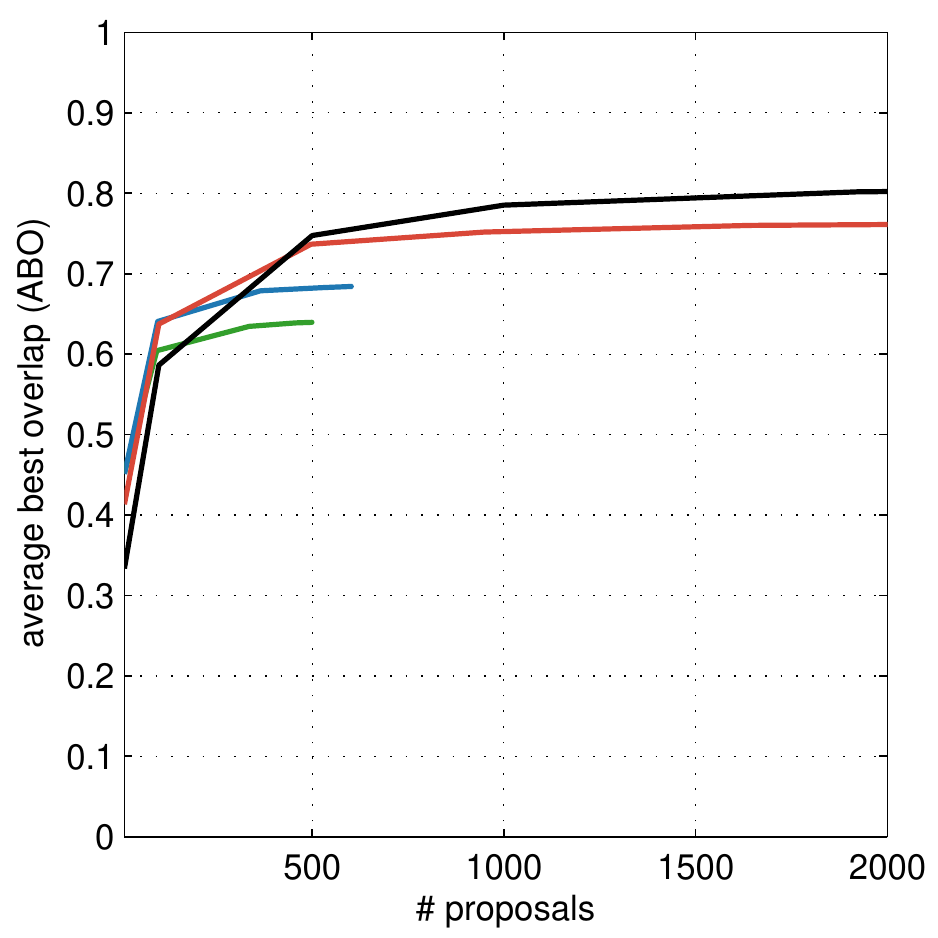}
 	}
 	\\	
 	\vspace{0.1cm}
 	\subfloat[Recall vs IoU (100 proposals) ]{\includegraphics[width=0.24\linewidth]{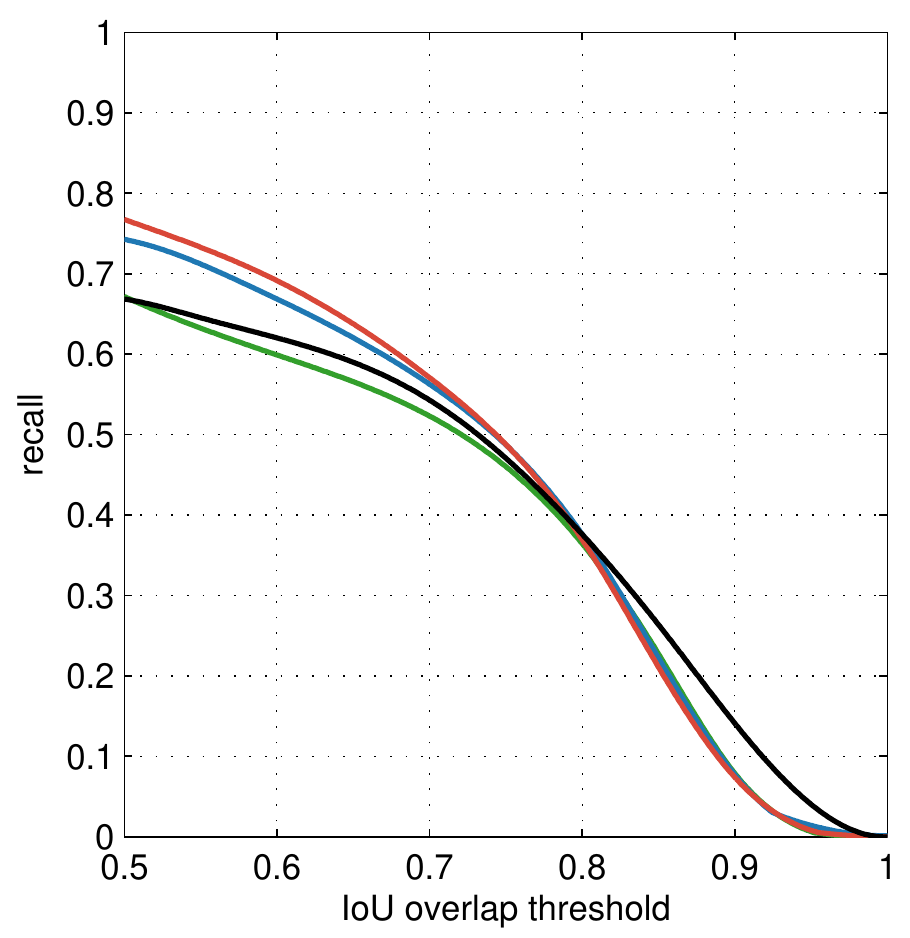}
 	}
 	\hspace{-0.3cm}  
 	\subfloat[Recall vs IoU (500 proposals) ]{\includegraphics[width=0.24\linewidth]{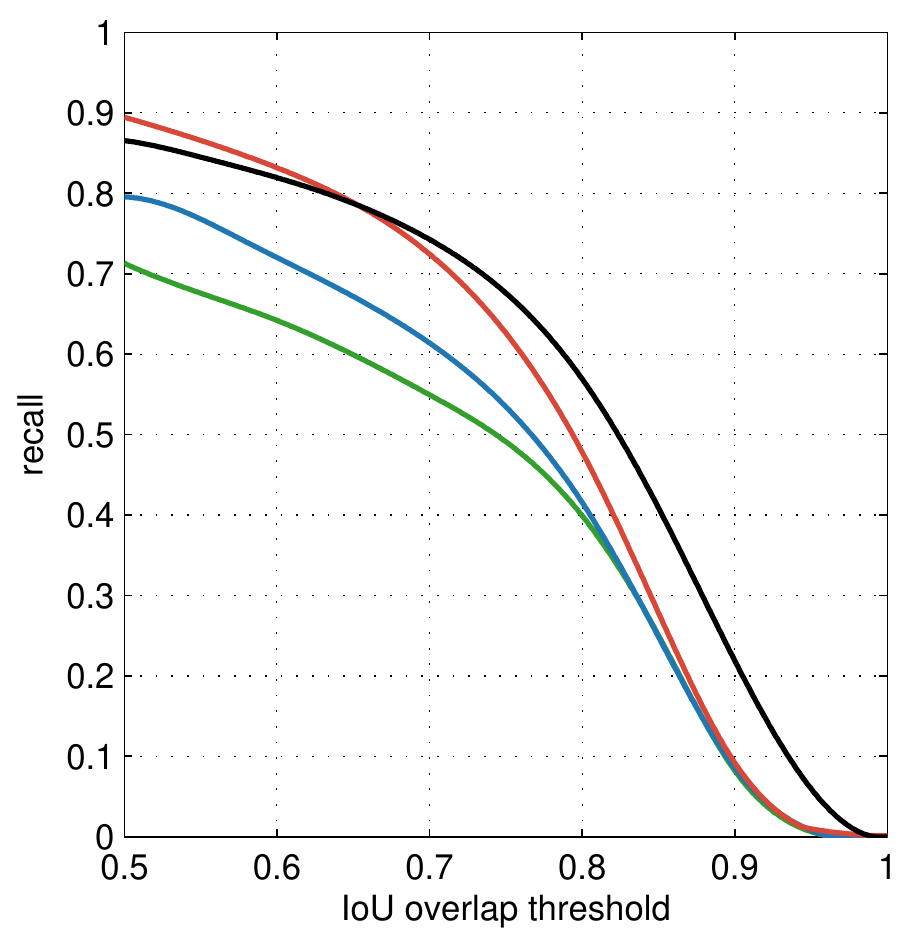}
 	}
 	\hspace{-0.3cm}
 	\subfloat[Recall vs IoU (1000 proposals) ]{\includegraphics[width=0.24\linewidth]{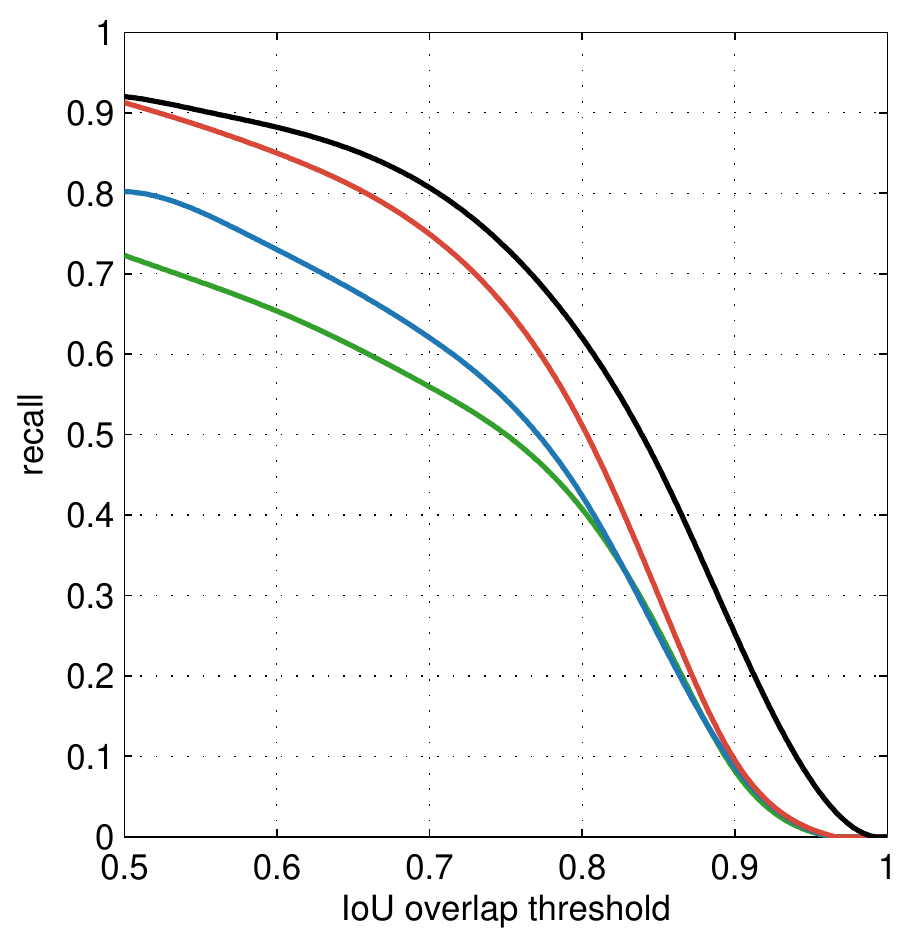}
 	} 
 	\hspace{0.3cm}
 	\subfloat{\raisebox{1.8cm}{\includegraphics[width=0.2\linewidth]{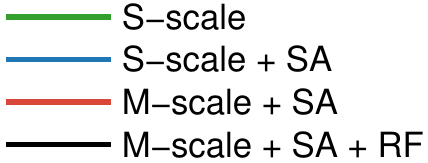}}
 	}		
 	\\    
 	\caption{Recall and average best overlap (ABO) comparison between different variants. S-scale, S-scale+SA, M-scale+SA, M-scale+SA+RF denote single-scale, single scale with scale-awareness, multi-scales with scale-awareness, multi-scales with scale-aware and refinement, respectively.  ``SA'' and ``RF'' denote ``scale-awareness'' and ``refinement'', respectively.}  
 	\label{fig:ablation_study}
 \end{figure*}
 
\section{Experiments and Discussion}
\label{sec:experiment}
\subsection{Experimental Setups}
The proposed Scale-aware Pixel-wise Object Proposal Network (SPOP-net) is trained on the SBD annotations~\cite{hariharan2011semantic} of PASCAL VOC 2012 trainval set, which provides $11{,}355$ images with fine segmentation masks annotations. We manually label the objects containing more than $2{,}000$ pixels as large objects and those containing less than $2{,}000$ pixels as small ones. Considering the unbalanced pixel samples when training the large-/small-size weighting branch, for each large object, we randomly sample $100$ pixels in it for training to balance the number of training pixels belonging to large and small objects.  Both the ``confidence network" and the two localization networks are trained  using the published DeepLab code~\cite{chen2014semantic}, which is based on the publicly available Deep Learning platform Caffe~\cite{jia2014caffe}. The weights in the newly added layers are all initialized with a zero-mean Gaussian distribution with the standard deviation $0.01$ and the biases are initialized with $0$.  The initial learning rate is $0.001$ for the pre-trained  layers in the DeepLab-LargeFOV network and $0.01$ for the newly-added layers. All of them are reduced by a scale of $10$ after every $20$ epochs. The mini-batch size is set as $8$. We train the network for about $60$ epochs. The overlap threshold for NMS in our experiments is set to 0.8 for a good trade-off between the recall at low IoU thresholds (\emph{e.g.} 0.5) and high IoU thresholds (\emph{e.g.} 0.8). The training images are all resized to 513*513. During testing, for original scale, all the images are directly fed into the networks without any scaling; for enlarged scale, all the images are enlarged by a factor of 2.

 The proposed SPOP-net is then extensively evaluated on PASCAL VOC 2007 testing set which is the most widely used in comparison of object proposal algorithms. It contains $4{,}952$ images with annotated objects (including ``hard'' objects) in bounding boxes. We are not able to evaluate on PASCAL VOC 2012 testing set because the ground-truths are not publicly released. Since the missed objects can never be recovered in the post-classification stage in a proposal-based object detection pipeline, object recall rate is naturally regarded as the standard evaluation metric for object proposal algorithms. Also, we evaluate the localization quality measured by Average Best Overlap (ABO). In addition, the object detection performance using our proposals in Fast-RCNN~\cite{girshick2015fast} detection pipeline is evaluated to validate the effectiveness of our proposals in the object detection task. Finally, we conduct the generalization ability evaluation by testing the recall rate on ILSVRC 2013 validation set using our network which is trained on PASCAL VOC 2012.

 \subsection{Ablation Studies}
 We first study the effectiveness of the four components in our method: pixel-wise localization network (basic setting), scale-aware localization, multi-scale inference and proposal refinement. Several simplified variants of the SPOP-net are tested in terms of the object recall rate on PASCAL VOC 2007 testing set. Specifically, we use the prediction only at the original scale without scale-awareness and proposal refinement as our baseline, which is referred to as single scale. Without scale-awareness, only one localization network is trained on all of the foreground pixels including both large-size and small-size ones. Then, we accumulatively add  scale-awareness, multi-scale inference, proposal refinement to the baseline to see the benefits of each component. Please note that multi-scale inference here indicates the prediction at two scales, namely the original image scale and the $2$-time enlarged scale.

 Figure~\ref{fig:ablation_study} shows the recall  and average best overlap (ABO) comparisons under different scenarios between the four variants, \emph{i.e.} single scale, single scale with scale-awareness, multi-scales with scale-awareness, multi-scales with scale-awareness and refinement.  The number of proposals of S-scale and S-scale+SA are around 500 due to that most proposals can be filtered after NMS as pixel-wise localization networks generate highly overlapped proposals (see Figure \ref{fig:visual}). From Figure \ref{fig:ablation_study}(a), \ref{fig:ablation_study}(b) and \ref{fig:ablation_study}(c), \ref{fig:ablation_study}(e), \ref{fig:ablation_study}(f) and \ref{fig:ablation_study}(g), we find that both scale-awareness and multi-scale inference improve the recall under both low IoU threshold (\emph{e.g.} $0.5$) and high IoU threshold (\emph{e.g.} $0.7$). As for proposal refinement, it is found that it harms the recall under low IoU thresholds (\emph{e.g.} $0.5$) when the number of proposals is less than $500$. The reason probably lies in the large number of proposals after refinement, which is $3$ times as big as that before refinement. Although this increases the opportunities of getting close to the ground-truths which can boost the recall for a large number of proposals, this also causes too many duplicate proposals to concentrate on a small area, which lowers down the recall under loose IoU criteria when only requiring a small number of proposals. For average best overlap, it shows a similar trend to the recall from Figure \ref{fig:ablation_study}(d), suggesting the benefits of all three components in terms of localization quality.

\begin{figure}
 	\centering
 	\includegraphics[width=1\linewidth,height=0.6\linewidth]{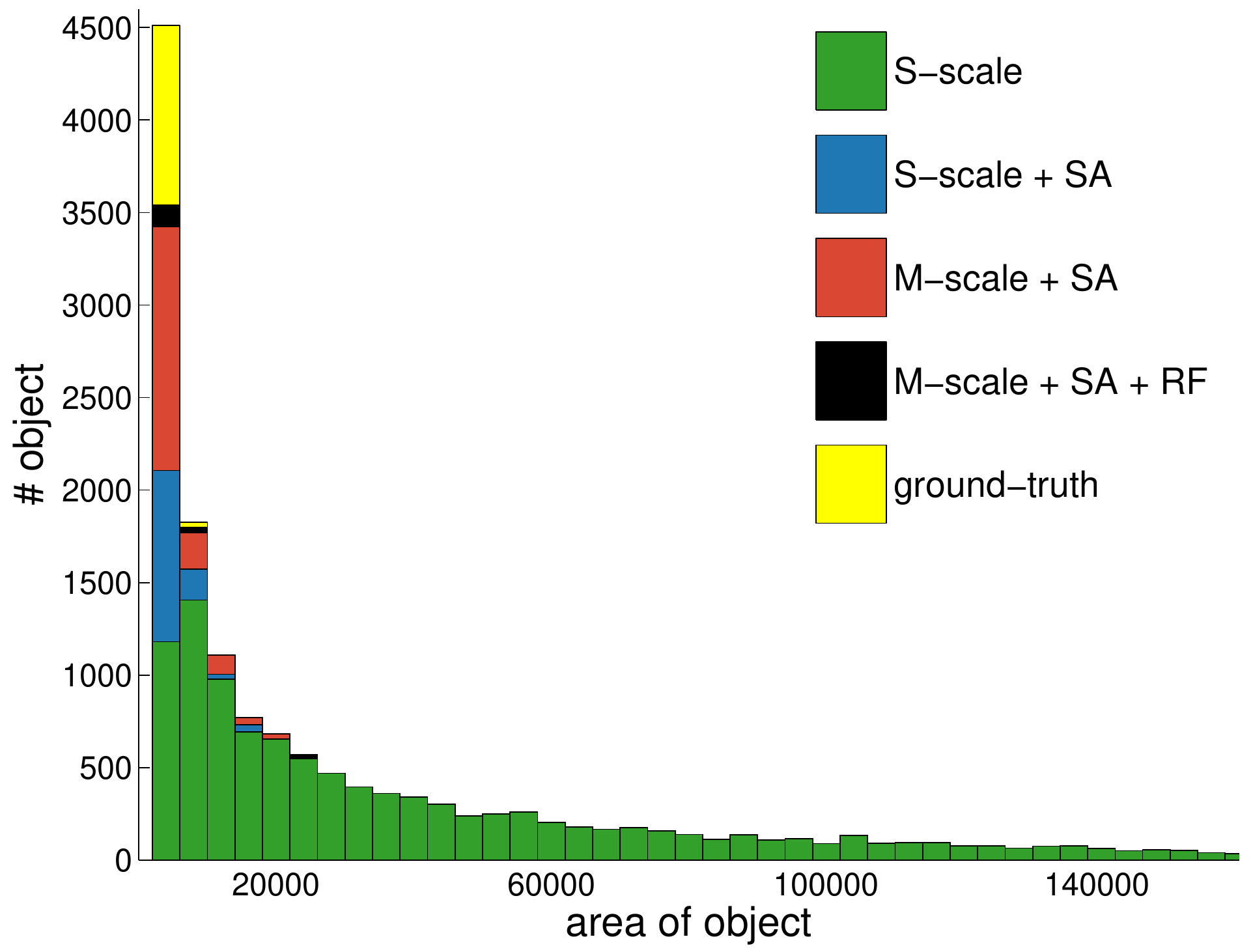}
 	\caption{Distribution of the detected objects w.r.t.\ the object areas (measured by number of contained pixels) on the PASCAL VOC 2007 testing set of the four variants of the SPOP-net. The IoU threshold is $0.5$. $2{,}000$ proposals are generated for each image.
 	}
 	\label{fig:error_analysis}
 	\vspace{-4mm}
 \end{figure}

 We then study the contributions of all the components for different object areas. Figure~\ref{fig:error_analysis} presents the distributions of the detected objects of both the four variants of SPOP-net and the ground-truths w.r.t the object areas. It is found that the baseline variant, \emph{i.e.} single scale without scare-awareness and refinement, can hit most of big objects but performs poor for small objects. Scare-aware weighted combination mechanism and multi-scale inference help improve the recall for small objects significantly, which shows the effectiveness of both the proposed scare-aware localization strategy and multi-scale inference.

\begin{figure}	
 	\subfloat{\includegraphics[width=0.99\linewidth,height=0.5\linewidth]{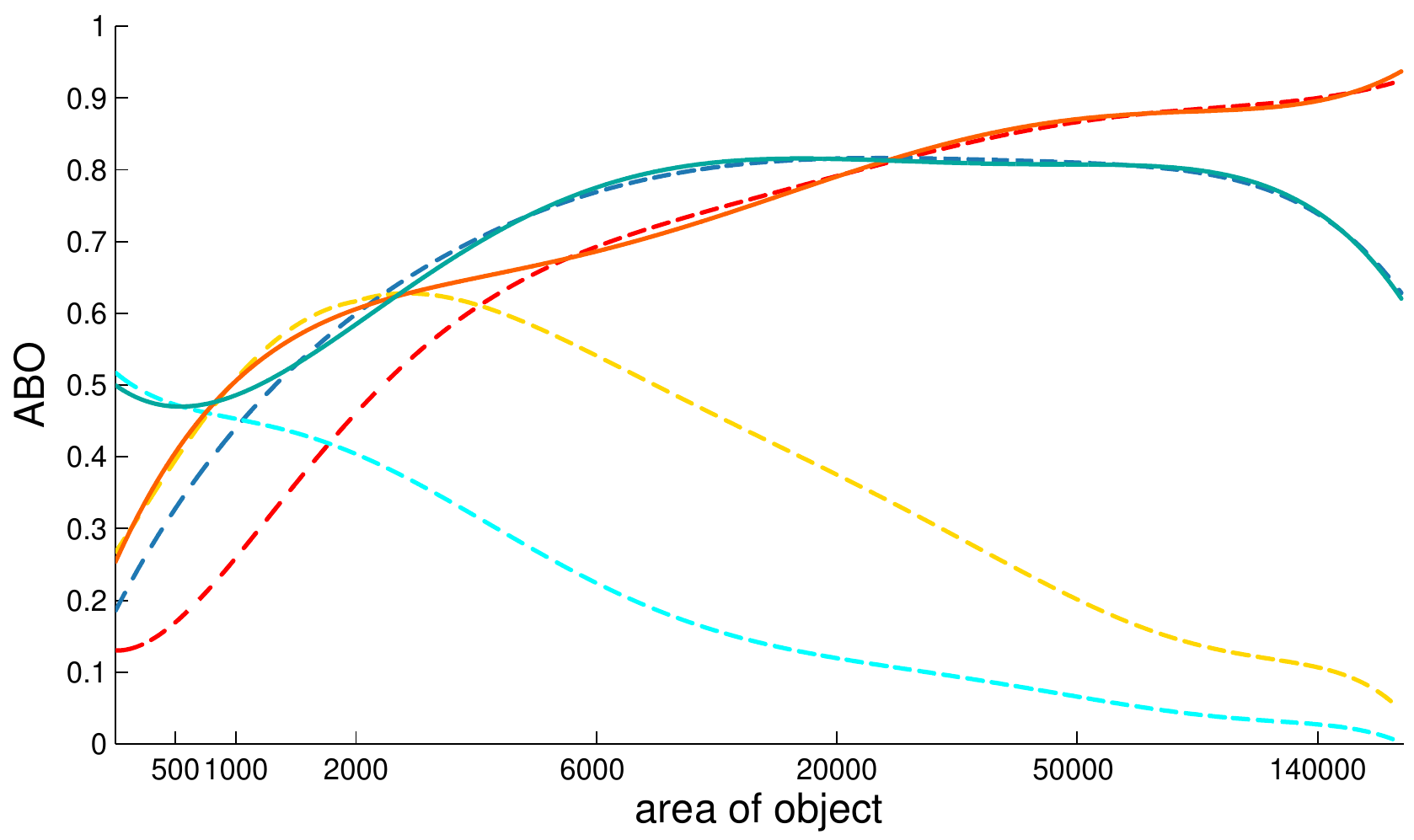}
 	}
 	\\	
 	\vspace{-0cm}
 	\subfloat{\includegraphics[width=1\linewidth,height=0.08\linewidth]{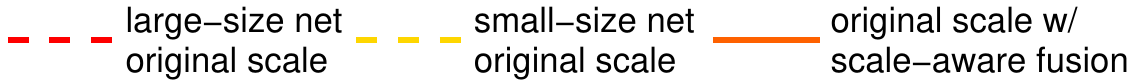}
 	}
 	\\    
 	\vspace{-0.1cm}
 	\hspace{-0.1cm}
 	\subfloat{\includegraphics[width=1\linewidth,height=0.08\linewidth]{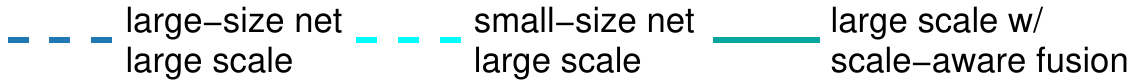}
 	}
 	\\    
 	\caption{Average best overlap (ABO) versus ground-truth object area for the four building blocks localization results: large-size network in original scale, small-size network in original scale, large-size network in enlarged scale and small-size network in enlarged scale. All the ABOs are computed given the top 1,000 proposals per image.}
 	\label{fig:ABO_area}
 \end{figure}

 To further verify the effectiveness of scale-awareness and multi-scale inference in small objects localization, we break up  the SPOP-net into four building blocks, \emph{i.e.} large-size network and small-size network in original scale, and large-size network  and small-size network in enlarged scale, in order to investigate their respective contributions to the final localization. We evaluate the average best overlap (ABO) of the four building blocks for the ground-truth objects with different areas. Figure~\ref{fig:ABO_area} shows the ABO versus object area curves of the four building blocks. It can be seen that when the object becomes larger, the large-size network in original scale predicts more accurate localization results. The small-size network in original scale achieves the highest ABO when the object area is around $2{,}000$, but it also performs poorly for those too small objects. Fortunately, the small-size network in enlarged scale covers this shortage, and gives the best performance for very small objects due to the enlarged view of small objects. As for the large-size network in enlarged scale, it performs the best for those middle-size objects containing $2{,}000$ to $20{,}000$ pixels, serving as the bridge between the large-size network in original scale and the small-size networks in both scales. The reason for the behavior of the large-size network in enlarged scale is probably that when the small objects are enlarged, they become ``large objects" such that it becomes easier for the large-size network to predict, but original large objects become even larger which cannot be covered by the receptive field, making it difficult to precisely localize them. In both original scale and enlarged scale, the result after \emph{scale-aware} fusion can achieve the maximal ABO among the two ABOs obtained by large-size and small-size networks, validating the effectiveness of the adaptive \emph{scale-aware} fusion strategy.

 By investigating the building blocks of the proposed SPOP-net, it is found that they can complement each other in localizing the objects with different areas and ensures the SPOP-net to perform well for a wide range of object sizes.
 
 \subsection{Comparisons on Object Recall }
 We compare our SPOP-net with the following state-of-the-art object proposal methods: BING~\cite{cheng2014bing}, Edge Boxes~\cite{zitnick2014edge}, Geodesic Object Proposal~\cite{krahenbuhl2014geodesic}, MCG~\cite{arbelaez2014multiscale}, Objectness~\cite{alexe2012measuring}, Selective Search~\cite{uijlings2013selective} and Region Proposal Network (RPN with VGG-16)~\cite{ren2015faster}. We first evaluate object recall on PASCAL VOC 2007 testing set, which contains $4{,}952$ images with about $15{,}000$ annotated objects. Proposals of most state-of-the-art methods were provided by Hosang et al.~\cite{Hosang2015arXiv} in a standard format. As for DeepProposal approach, we directly downloaded the pre-computed proposals from the official website\footnote{https://github.com/aghodrati/deepproposal}.

\begin{figure*}	
	\subfloat[Recall vs \# proposal (IoU=0.5) ]{\includegraphics[width=0.248\linewidth]{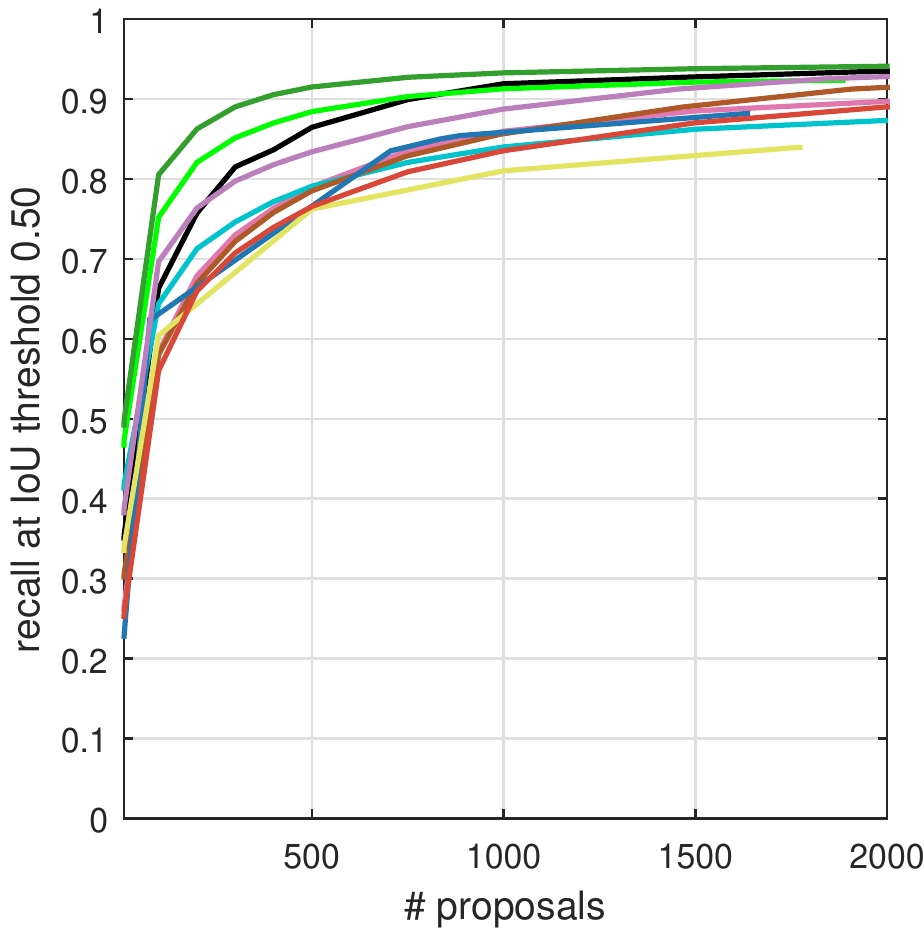}
	}
	\hspace{-0.3cm}
	\subfloat[Recall vs \# proposal (IoU=0.7) ]{\includegraphics[width=0.248\linewidth]{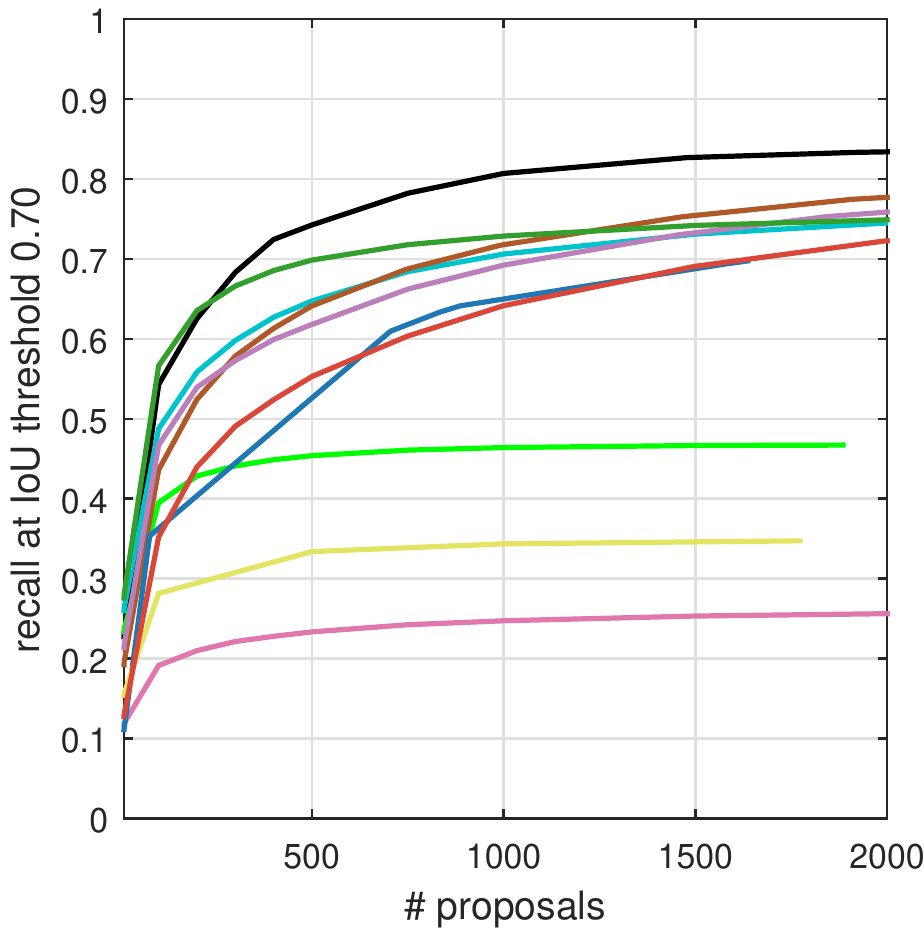}
	} 
	\hspace{-0.3cm}
	\subfloat[AR vs \# proposal (0.5$<$IoU$<$1) ]{\includegraphics[width=0.248\linewidth]{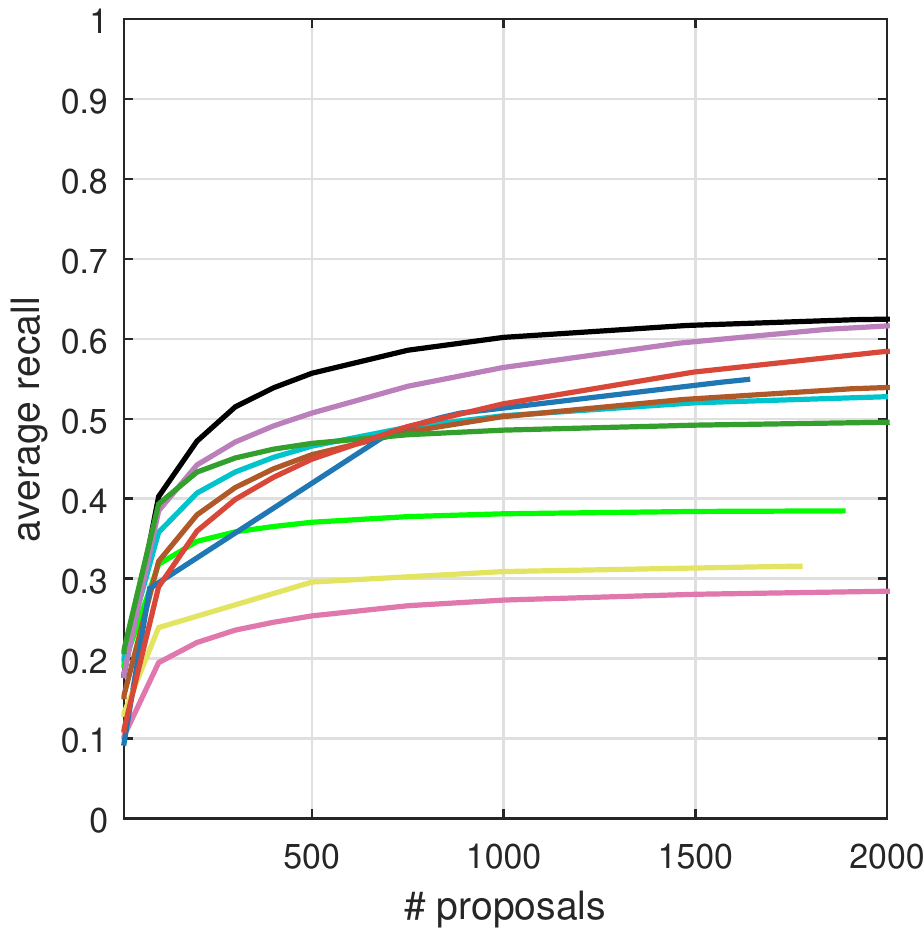}
	}
	\hspace{-0.3cm}
	\subfloat[ABO vs \# proposal  ]{\includegraphics[width=0.248\linewidth]{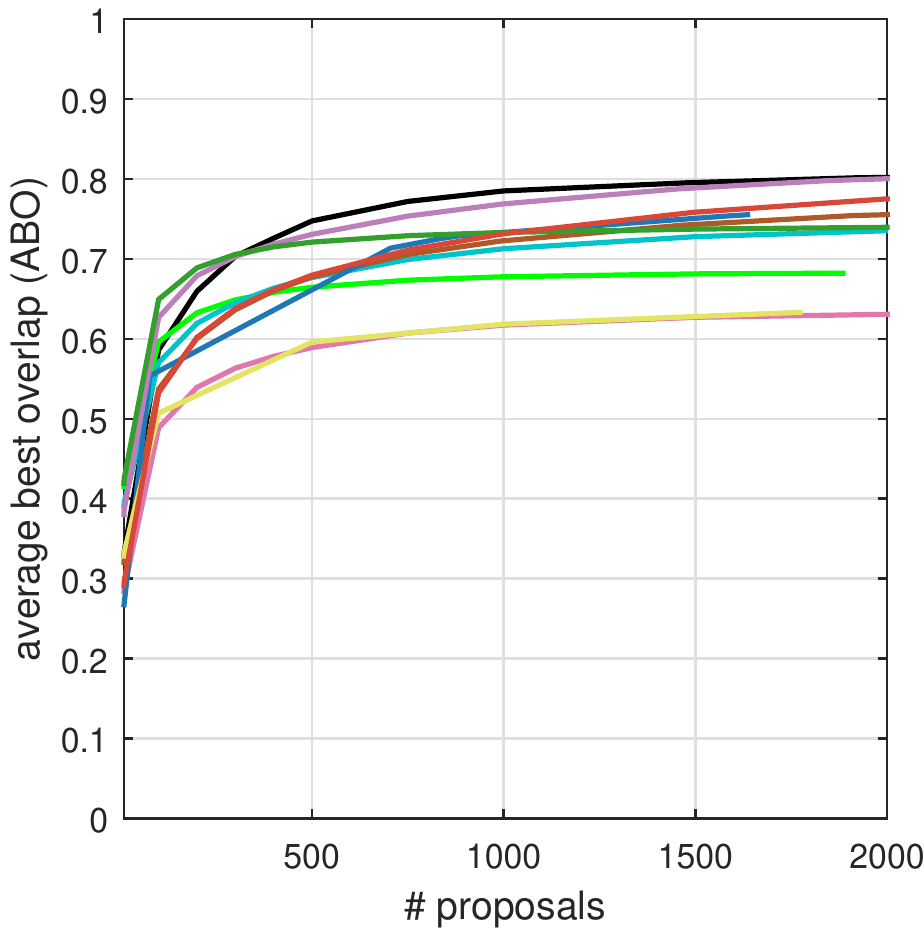}
	}
	\vspace{0.1cm}
	\\	
	\vspace{0.1cm}
	\subfloat[Recall vs IoU (100 proposals) ]{\includegraphics[width=0.24\linewidth]{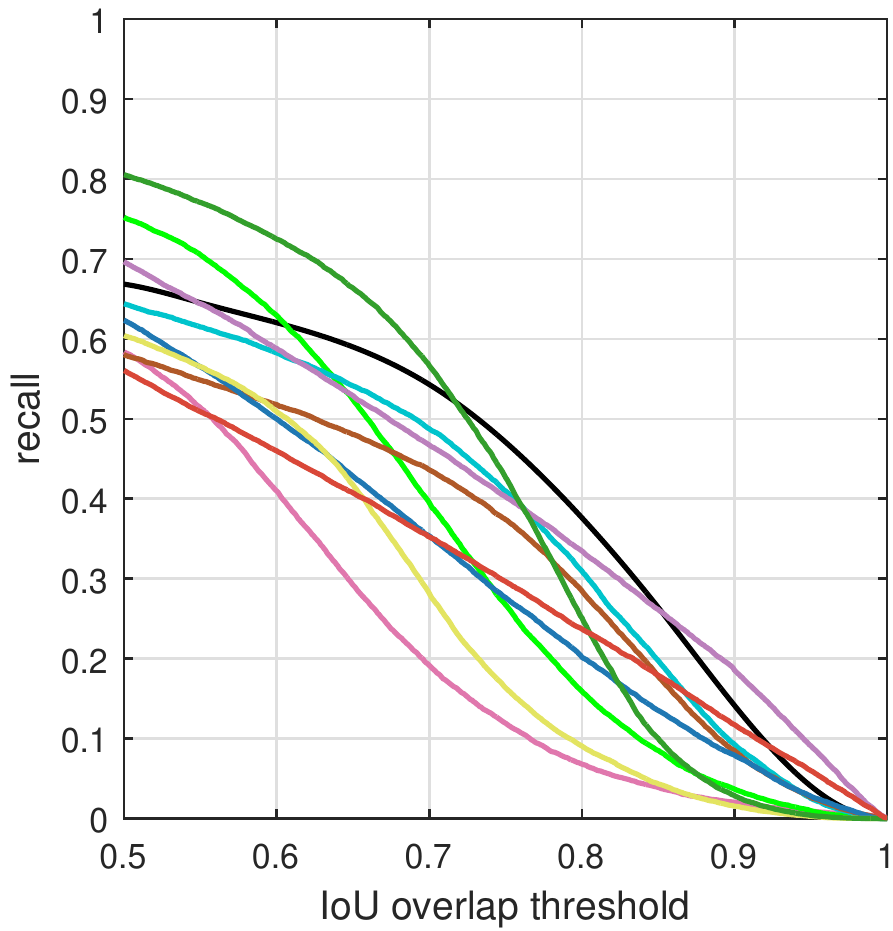}
	}
	\hspace{-0.1cm}  
	\subfloat[Recall vs IoU (500 proposals) ]{\includegraphics[width=0.24\linewidth]{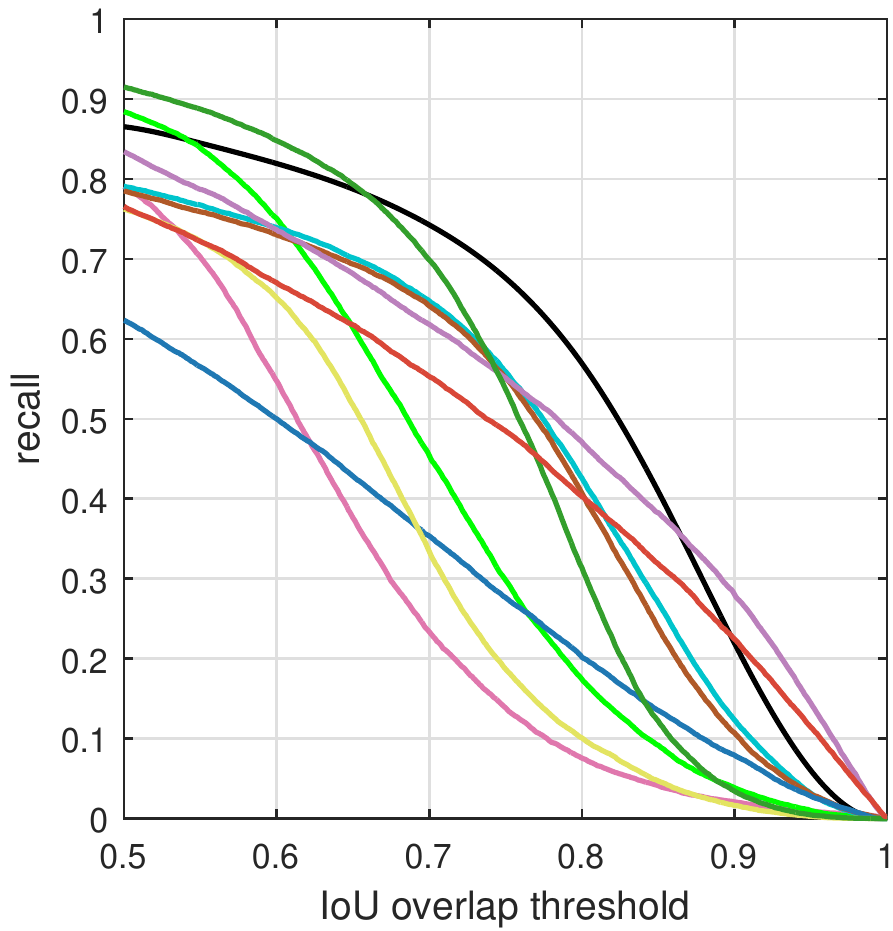}
	}
	\hspace{-0.1cm}
	\subfloat[Recall vs IoU (1000 proposals) ]{\includegraphics[width=0.24\linewidth]{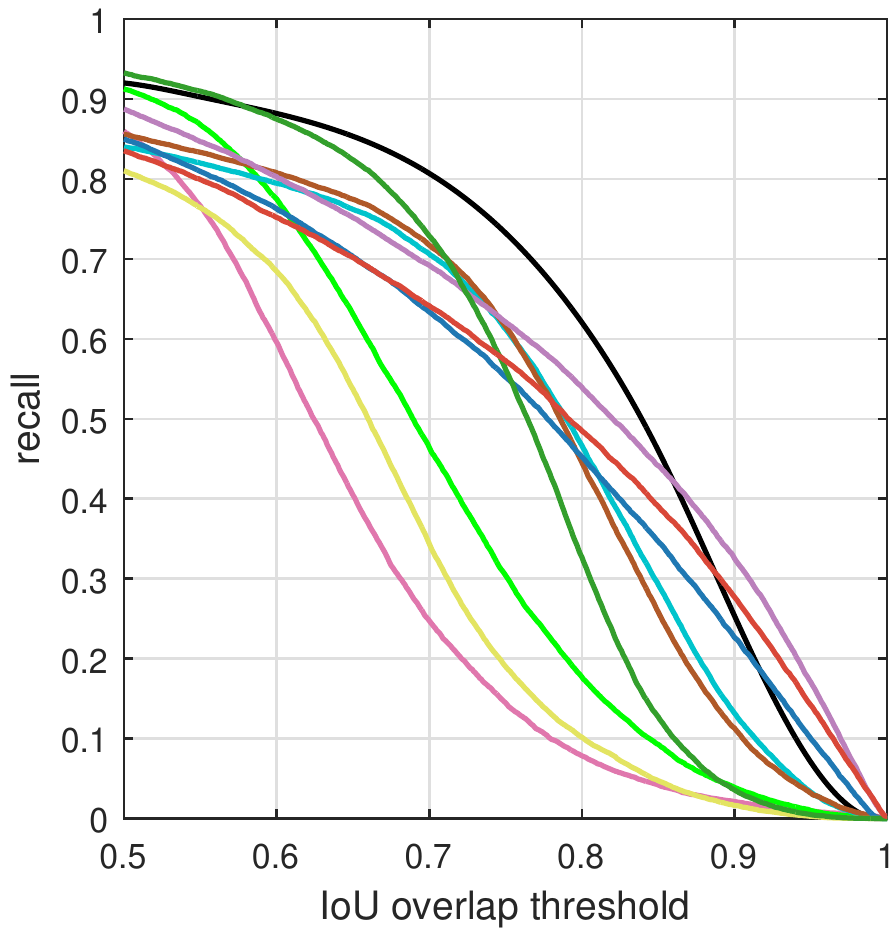}
	} 
	\hspace{0.3cm}
	\subfloat{\raisebox{0.9cm}{\includegraphics[width=0.17\linewidth,height=0.19\linewidth]{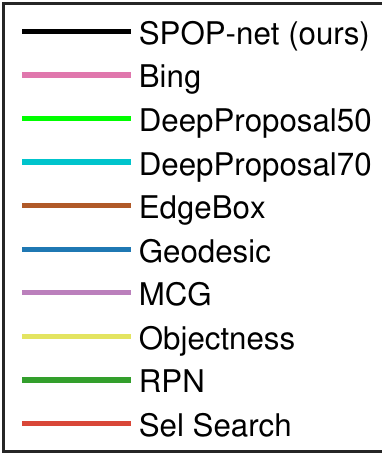}}
	}		 \\    
	\caption{Recall and average best overlap (ABO) comparison between our SPOP-net and other state-of-the-arts on PASCAL VOC 2007 testing set.}  
	\label{fig:recall_SOA}
\end{figure*}

\begin{figure*}	
	\subfloat[Recall with 100 proposals ]{\includegraphics[width=0.19\linewidth,height=0.2\linewidth]{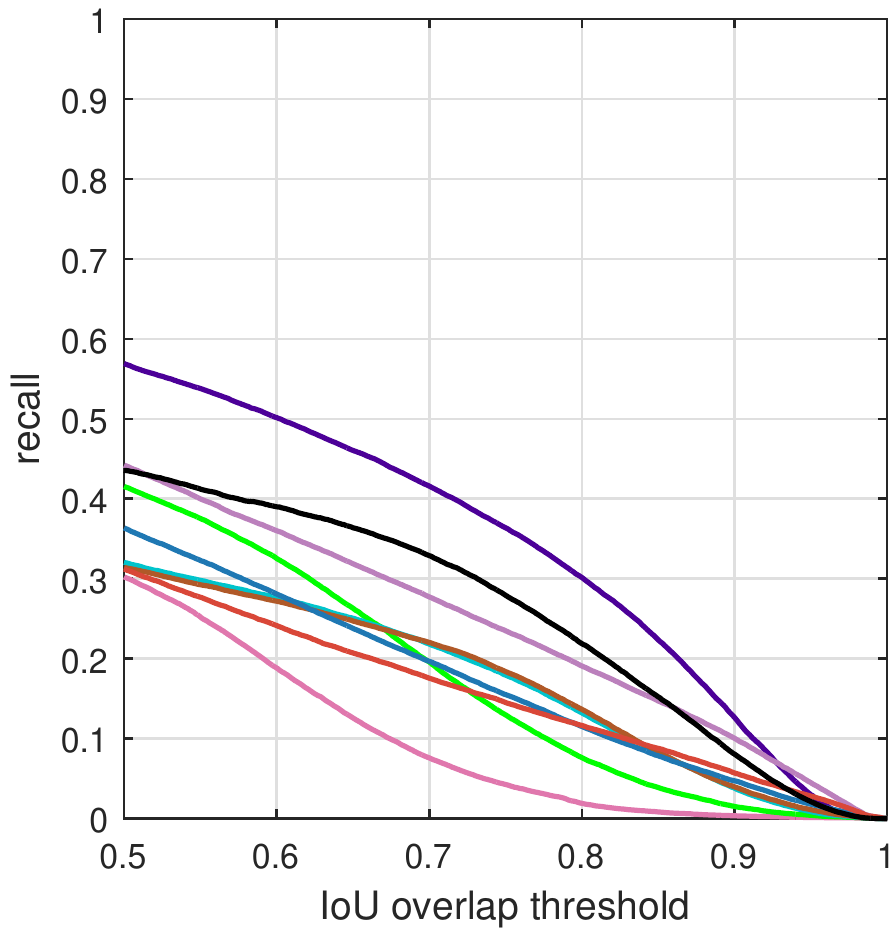}
	}
	\hspace{-0.1cm}
	\subfloat[Recall with 500 proposals ]{\includegraphics[width=0.19\linewidth,height=0.2\linewidth]{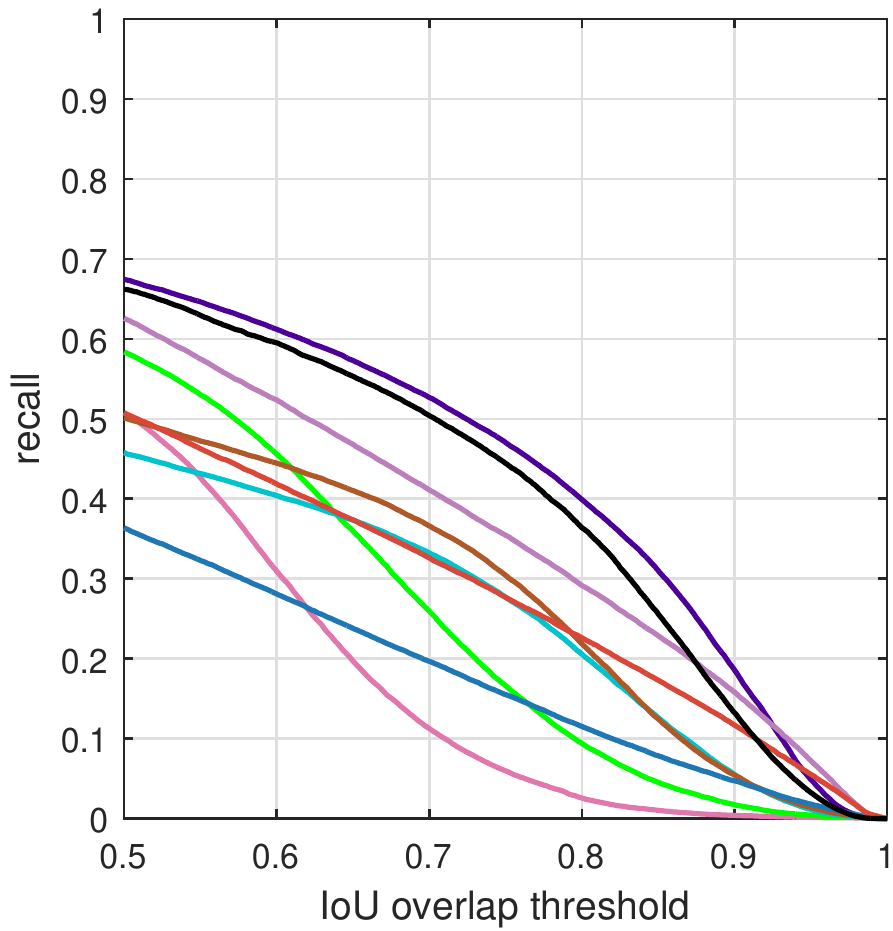}
	} 
	\hspace{-0.1cm}
	\subfloat[Recall with 1000 proposals ]{\includegraphics[width=0.19\linewidth,height=0.2\linewidth]{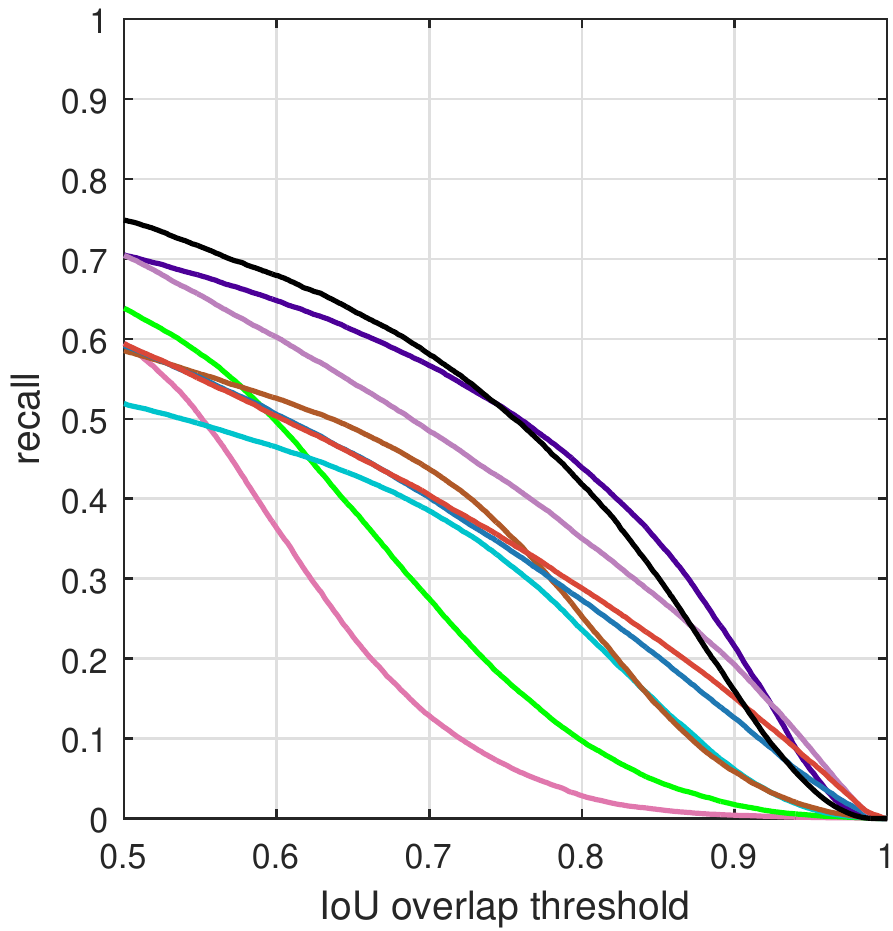}
	}
	\hspace{-0.1cm}
	\subfloat[Recall at IoU 0.5  ]{\includegraphics[width=0.19\linewidth,height=0.2\linewidth]{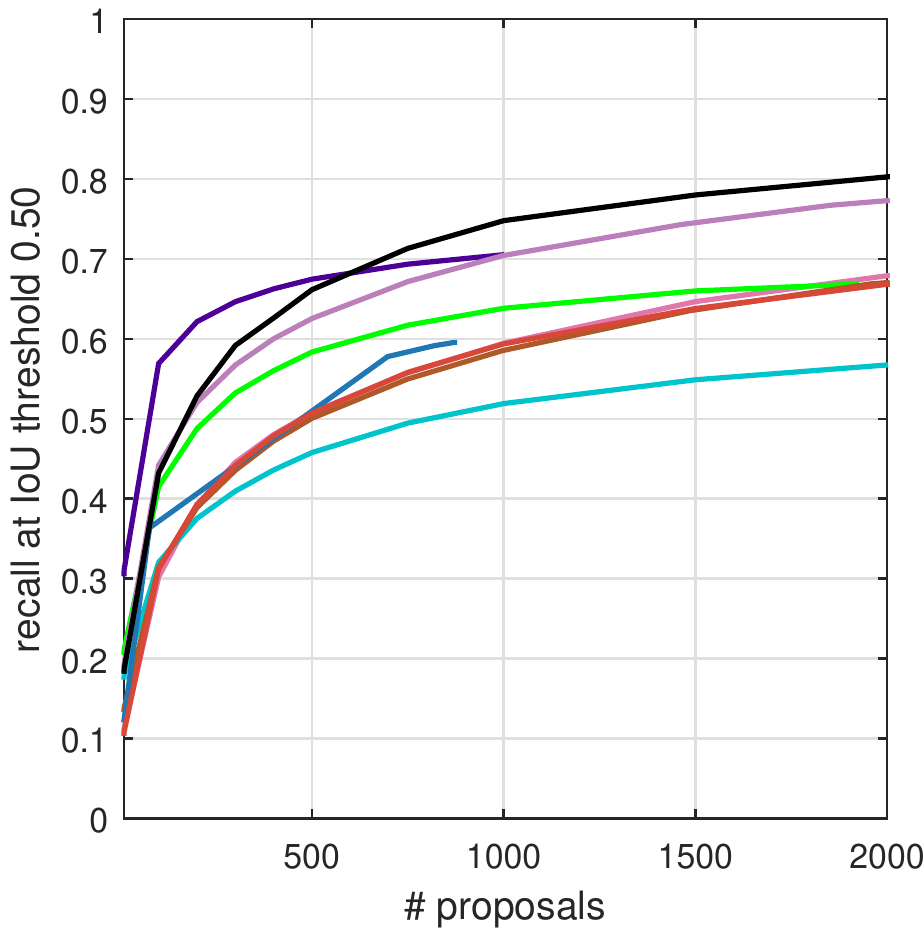}
	}
	\hspace{-0.1cm}
	\subfloat[Recall at IoU 0.7  ]{\includegraphics[width=0.19\linewidth,height=0.2\linewidth]{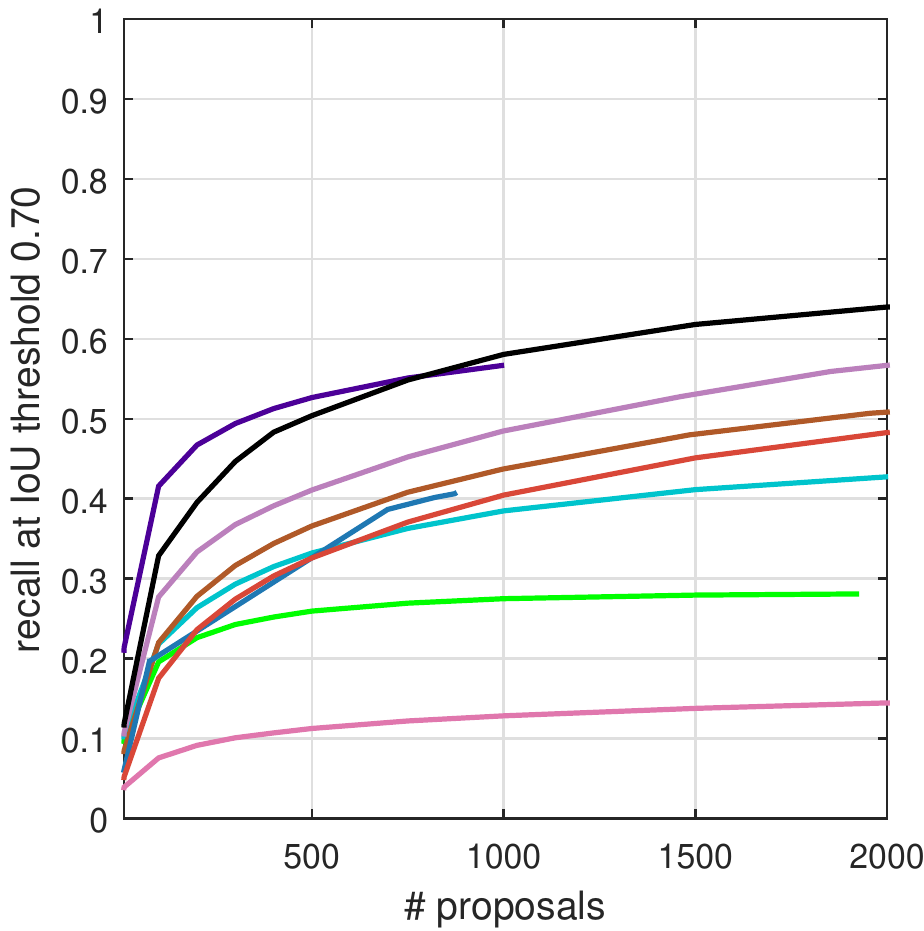}
	}
	\vspace{0.1cm}
	\\	
	\vspace{0.1cm}
	\subfloat[ABO vs \# proposal ]{\includegraphics[width=0.19\linewidth,height=0.2\linewidth]{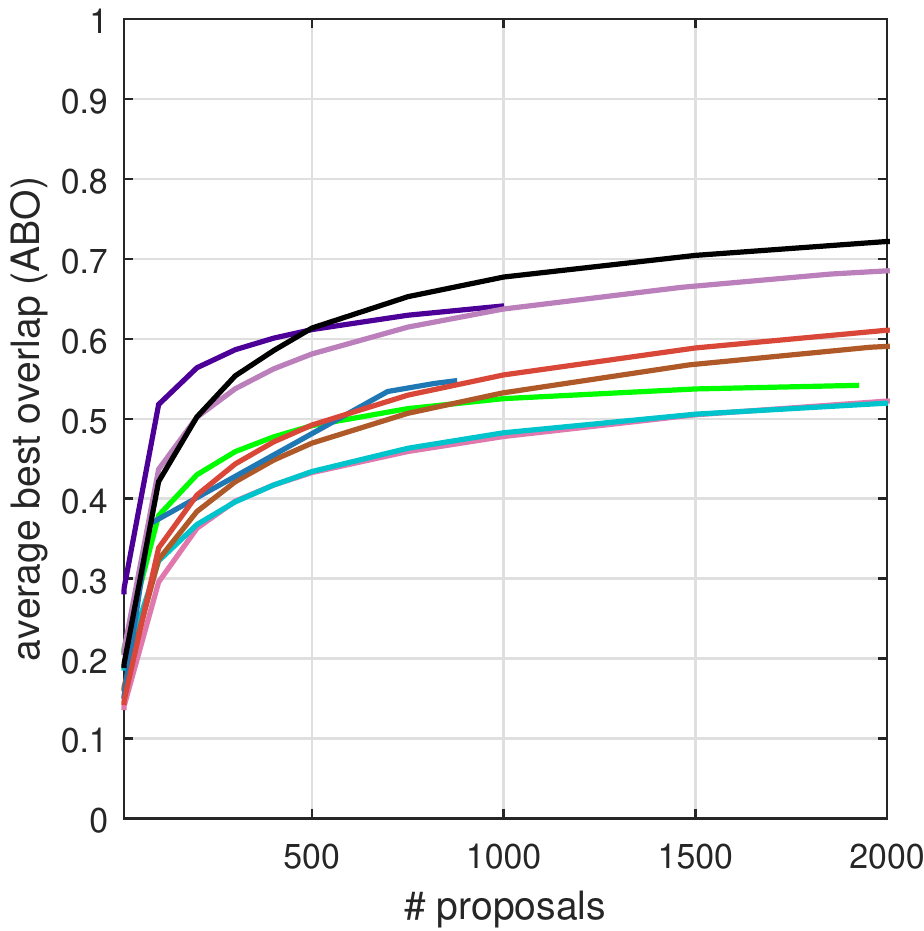}
	}
	\hspace{-0.1cm}  
	\subfloat[AR vs \# proposal  ]{\includegraphics[width=0.19\linewidth,height=0.2\linewidth]{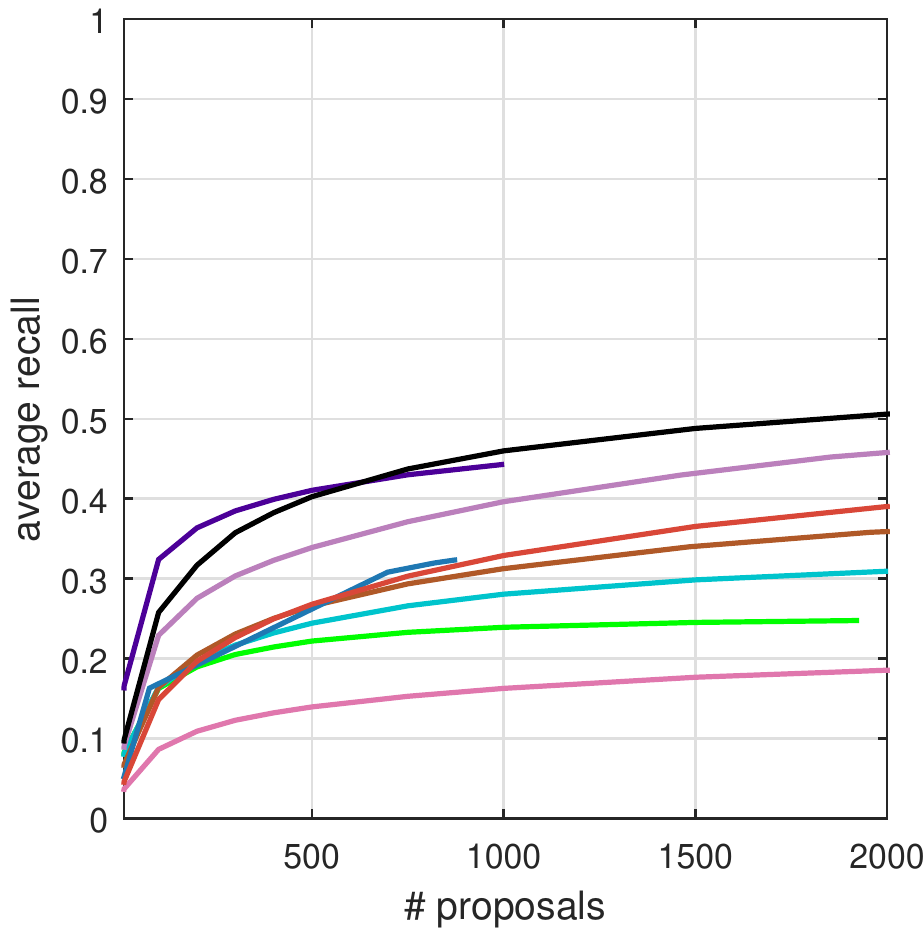}
	}
	\hspace{-0.1cm}
	\subfloat[AR vs \# proposal (large) ]{\includegraphics[width=0.19\linewidth,height=0.2\linewidth]{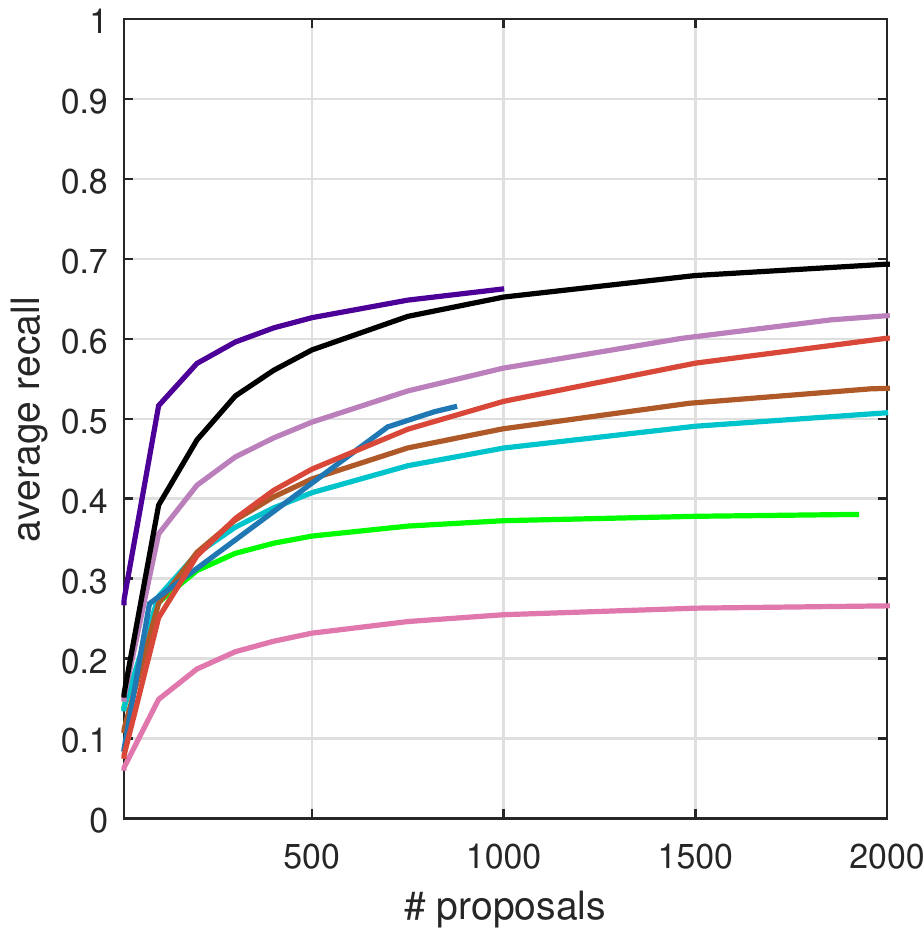}
	} 
    \hspace{-0.1cm}
	\subfloat[AR vs \# proposal (small) ]{\includegraphics[width=0.19\linewidth,height=0.2\linewidth]{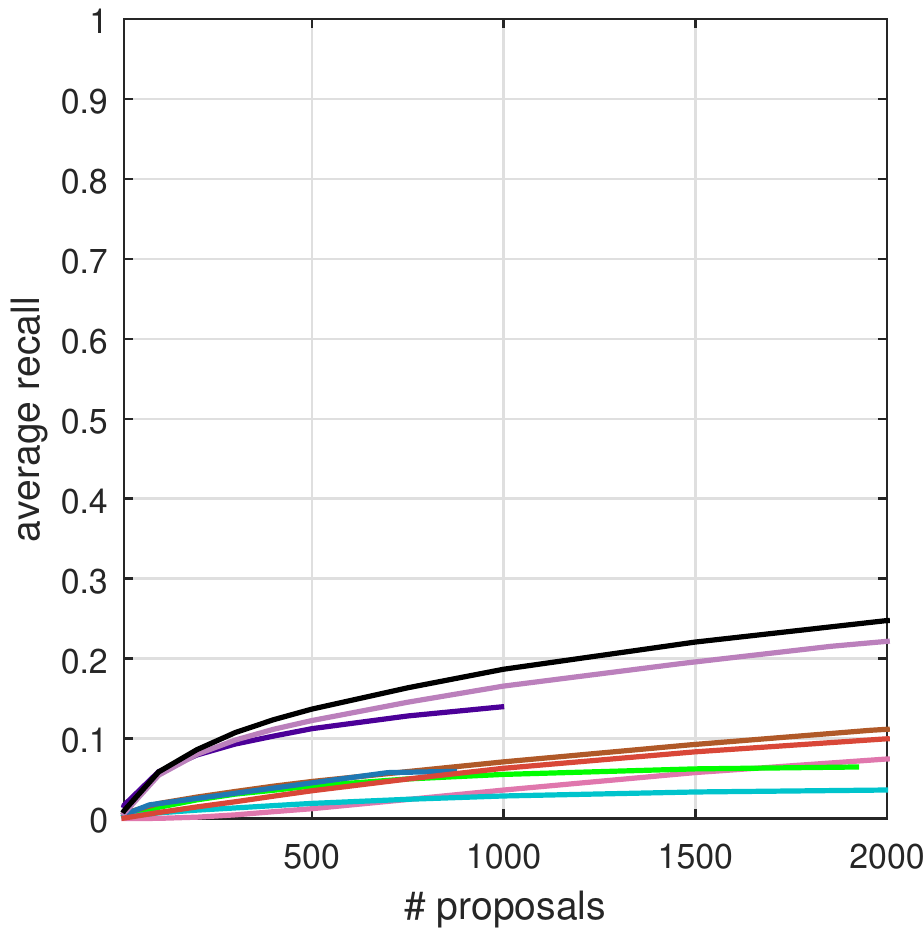}
	} 
	\hspace{-0.1cm}
	\subfloat{\raisebox{0.6cm}{\includegraphics[width=0.15\linewidth]{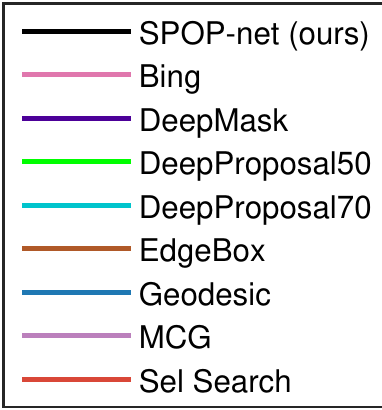}}
	}		 \\    
	\caption{Recall and average best overlap (ABO) comparison between our SPOP-net and other state-of-the-arts on MS COCO 2014 validation set.}  
	\label{fig:recall_coco}
\end{figure*}
 
  Figure~\ref{fig:recall_SOA}(a)  and \ref{fig:recall_SOA}(b) show the recall when varying the number of proposals for different IoU thresholds. As can be seen, under a loose $0.5$ IoU threshold, RPN takes the lead all the time for both a small and a large number of proposals.DeepProposal 50 also performs well under low IoU thresholds (\emph{e.g.} 0.5). Given a more strict IoU threshold $0.7$, our SPOP-net almost keeps the best consistently. We also plot the average recall (AR) versus the number of proposals curves for all the methods in Figure \ref{fig:recall_SOA}(c). This is because AR summarizes proposal performance across IoU thresholds and correlates well with object detection performance~\cite{Hosang2015arXiv}. The proposed SPOP-net also takes the first place all the time regarding the number of proposals. Figure \ref{fig:recall_SOA}(d) shows the average best overlap (ABO) when changing the number of proposals. The proposed SPOP-net shows good localization quality, especially when the number of proposals is more than $500$. Figure \ref{fig:recall_SOA}(e), \ref{fig:recall_SOA}(f) and \ref{fig:recall_SOA}(g) demonstrate the recall when the IoU threshold changes within the range [$0.5$, $1$] for different numbers of proposals. It is found that RPN performs well with a small number of proposals when setting a low IoU threshold ($<0.7$). When increasing the number of proposals from $100$ to $1{,}000$, our SPOP-net gradually shows its advantage. Especially for the top $1{,}000$ proposals, the SPOP-net performs superiorly across a wide range of IoU thresholds from $0.5$ to $0.85$, which have the strongest correlation to object detection performance thus are typically desired in practice~\cite{Hosang2015arXiv}.

\begin{figure}	
               	\subfloat{\includegraphics[width=0.99\linewidth,height=0.5\linewidth]{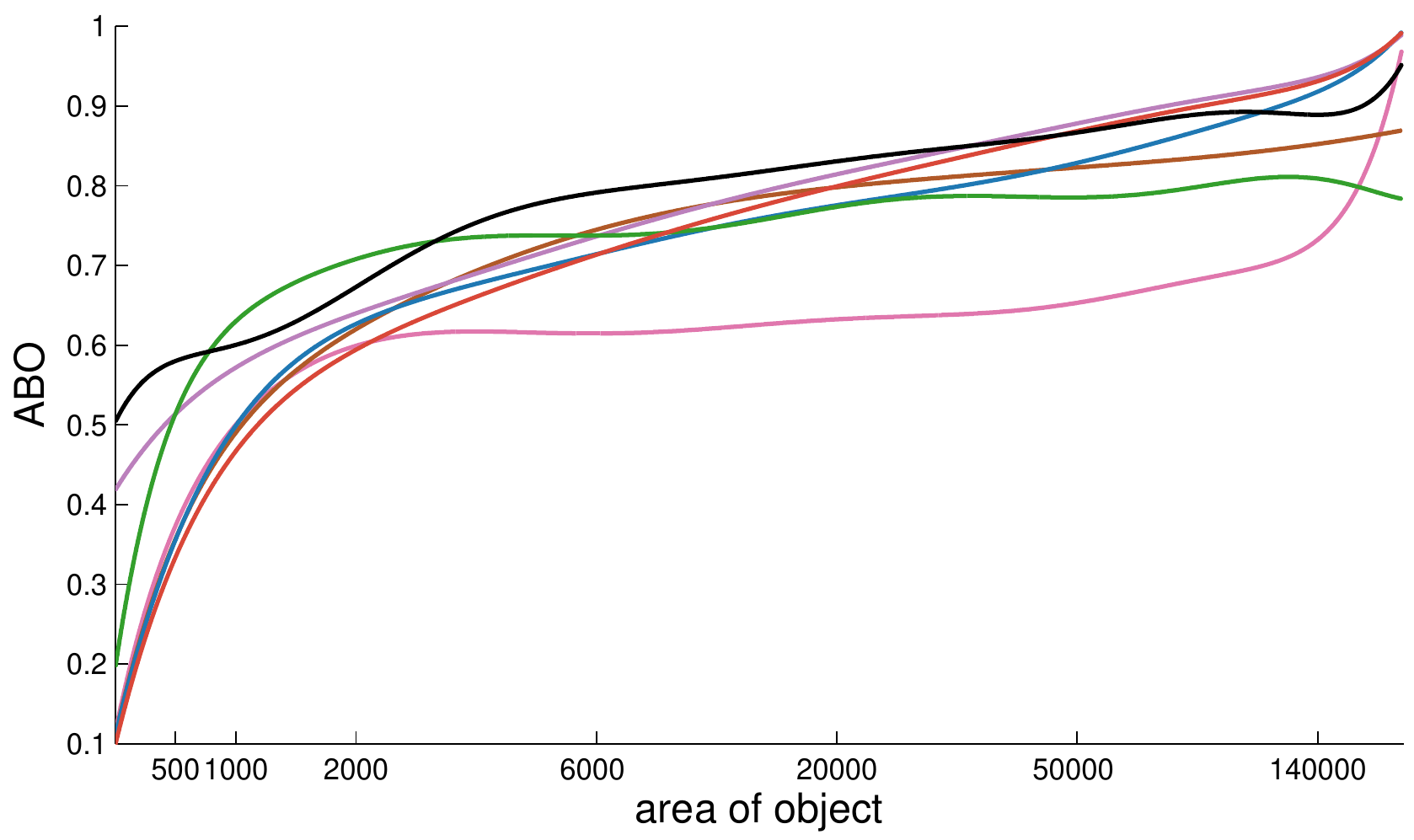}
               	}
               	\\	
               	\vspace{-0.1cm}
               	\subfloat{\includegraphics[width=1\linewidth]{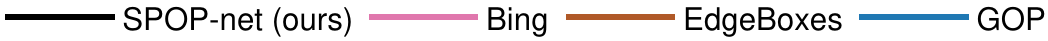}
               	}
               	\\    
               	\vspace{-0.1cm}
               	\hspace{0.8cm}
               	\subfloat{\includegraphics[width=0.74\linewidth]{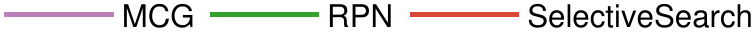}
               	}
               	\\
               	\caption{Average best overlap (ABO) versus ground-truth object area for the SPOP-net and other state-of-the-art methods. All the ABO are computed given the top $1{,}000$ proposals per image.}
               	\label{fig:ABO_area_SOA}
               	\vspace{-5mm}
               \end{figure}

  Figure~\ref{fig:ABO_area_SOA} shows the average best overlap (ABO) of the proposed SPOP-net as well as several state-of-the-art methods for the ground-truth objects with different areas. For most object sizes, the SPOP-net shows outstanding performance. Especially for small objects whose area is less than about $1{,}000$, the SPOP-net takes the first place, achieving an ABO higher than $0.5$. RPN can achieve a good ABO around $0.7$ for the objects whose areas are more than $2{,}000$ pixels, but can hardly reach a higher ABO even if the object is large. This may explain why the recall of RPN is very high when setting a loose IoU threshold (\emph{e.g.} $0.5$) but decreases sharply with the increasing of IoU threshold when it exceeds $0.7$. The classic low-level cues based methods (\emph{e.g.} Selective Search, MCG, GOP) perform very well for large objects but have inferior performance for small ones compared with two CNN-based methods (\emph{i.e.} SPOP-net, RPN).

\begin{figure*}
               	\captionsetup[subfigure]{labelformat=empty}	
               	\vspace{-0.3cm}
               	\hspace{0cm}
               	\subfloat[]{\includegraphics[width=0.23   \linewidth,height=0.17  \linewidth]{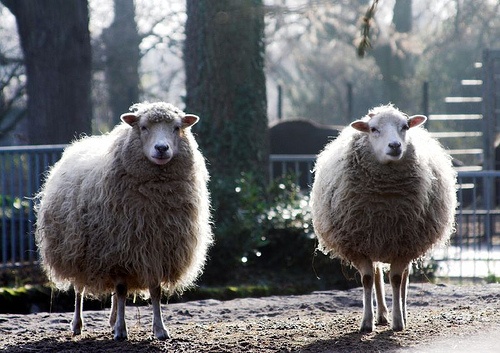}
               	}
               	\hspace{0cm}
               	\subfloat[]{\includegraphics[width=0.23   \linewidth,height=0.17  \linewidth]{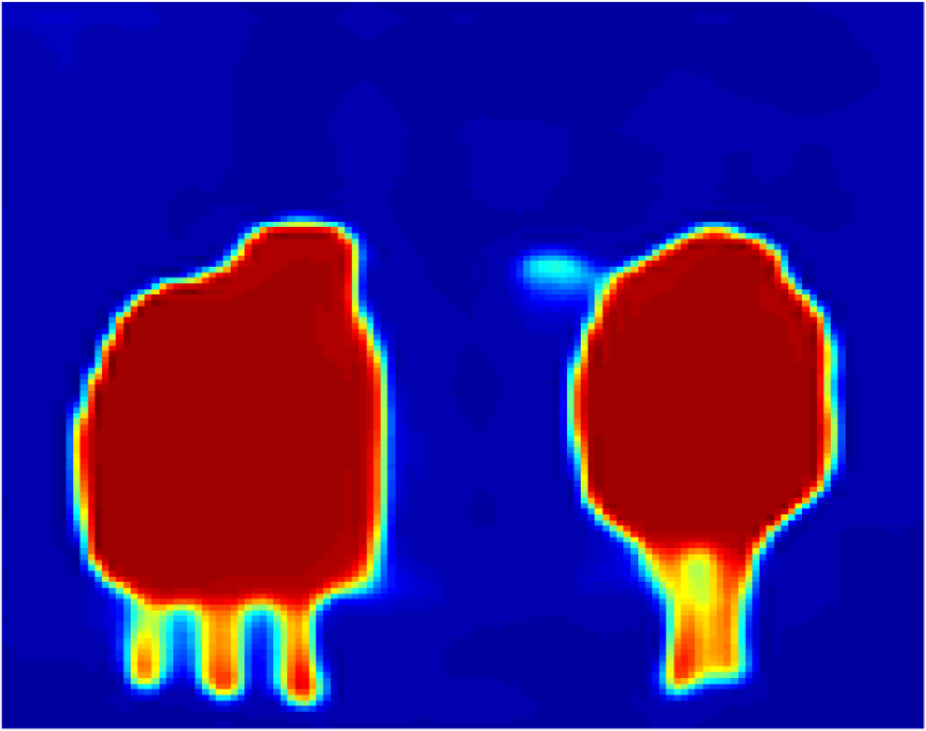}
               	} 
               	\hspace{0cm}
               	\subfloat[]{\includegraphics[width=0.23   \linewidth,height=0.17  \linewidth]{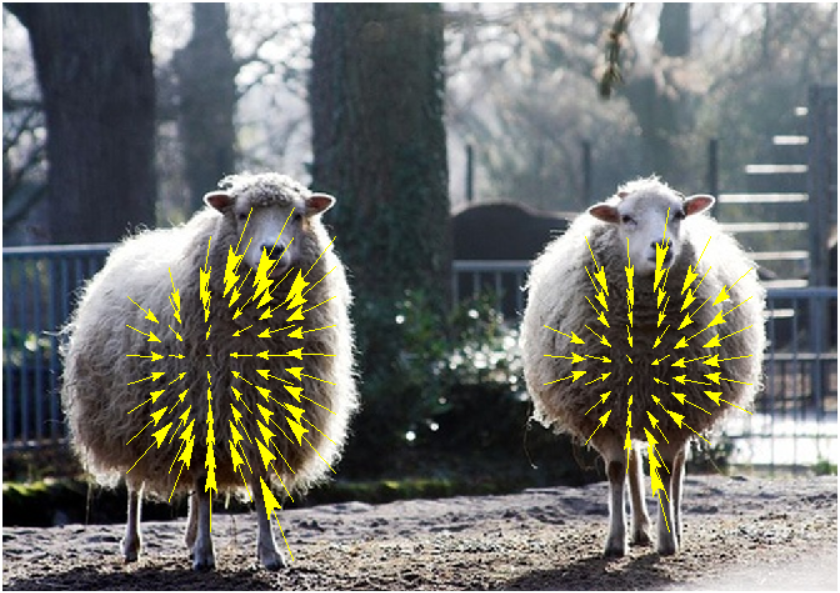}
               	}
               	\hspace{0cm}
               	\subfloat[]{\includegraphics[width=0.23   \linewidth,height=0.17  \linewidth]{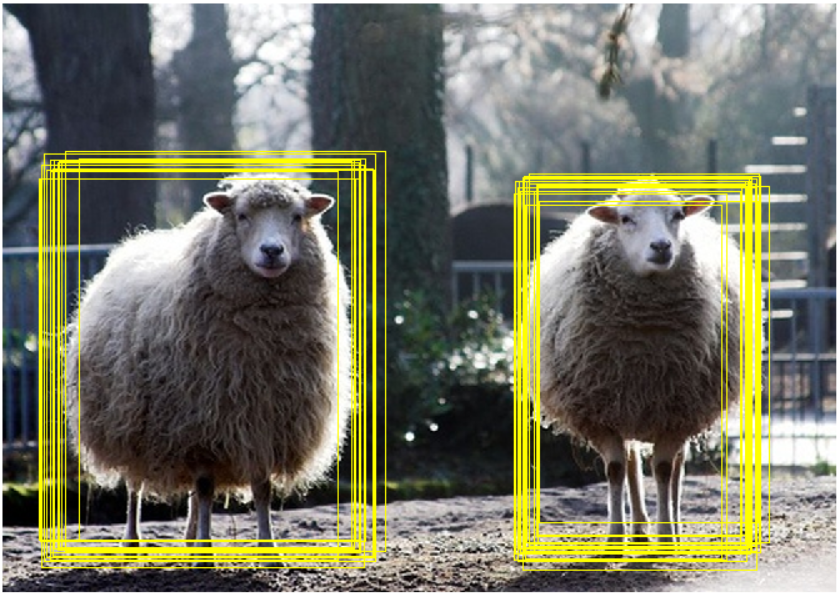}
               	} 
               	\vspace{-0.2cm}
               	\\
               	\vspace{-0.2cm}
               	\hspace{-0.1cm}
               	\subfloat[]{\includegraphics[width=0.23   \linewidth,height=0.17  \linewidth]{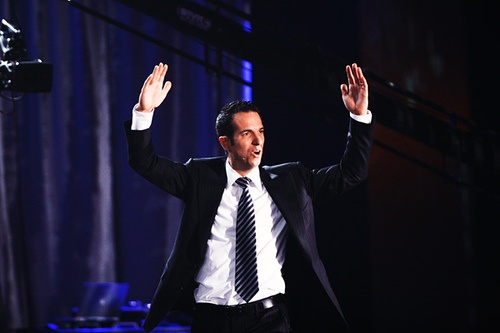}
               	}
               	\hspace{0cm}
               	\subfloat[]{\includegraphics[width=0.23   \linewidth,height=0.17  \linewidth]{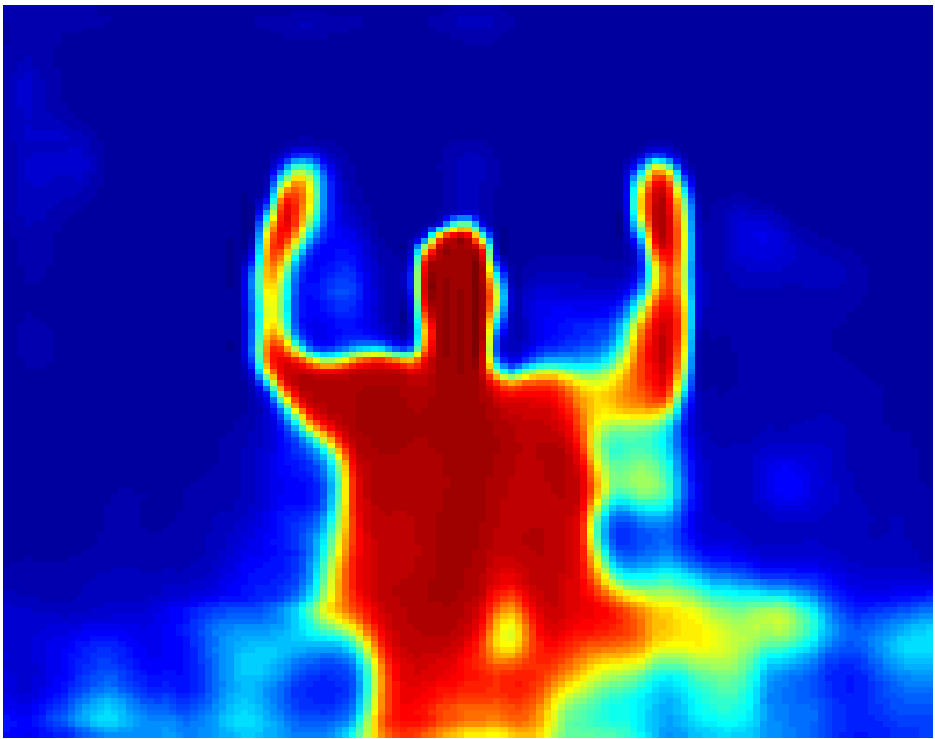}
               	} 
               	\hspace{0cm}
               	\subfloat[]{\includegraphics[width=0.23   \linewidth,height=0.17  \linewidth]{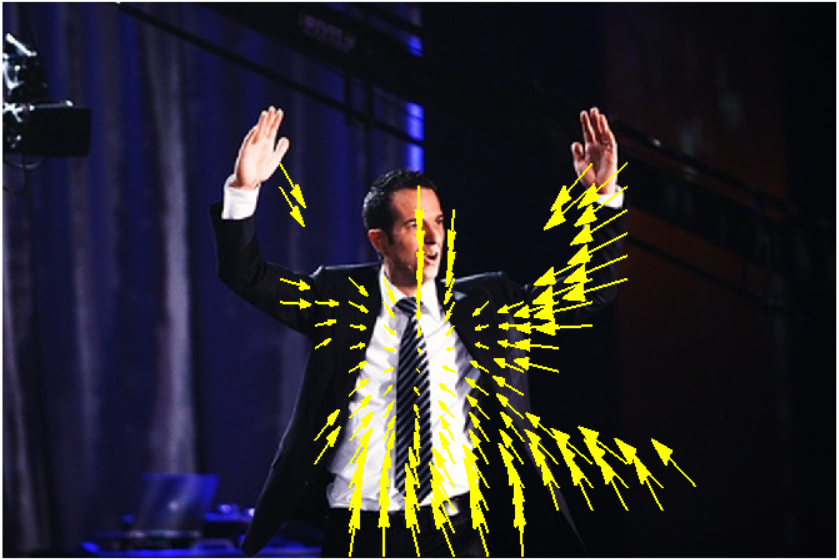}
               	}
               	\hspace{0cm}
               	\subfloat[]{\includegraphics[width=0.23   \linewidth,height=0.17  \linewidth]{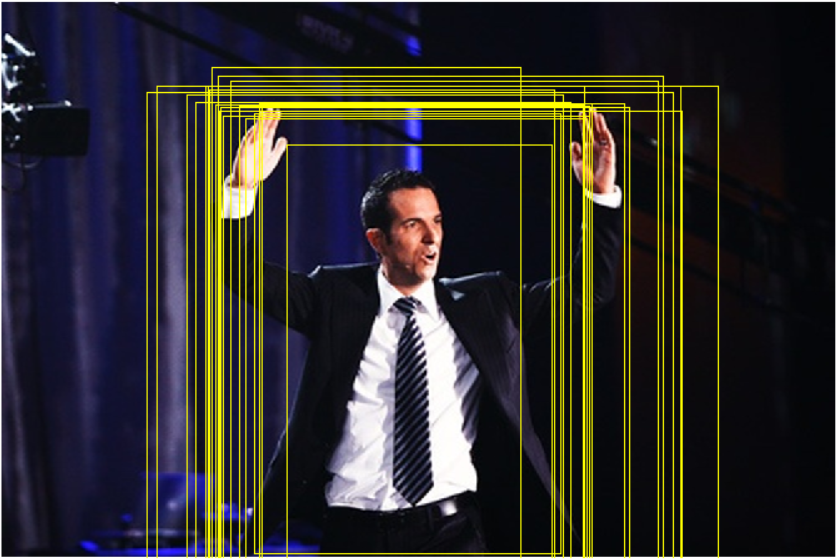}
               	} 
               	\\
               	\vspace{-0.2cm}
               	\hspace{-0.1cm}
               	\subfloat[]{\includegraphics[width=0.23   \linewidth,height=0.17  \linewidth]{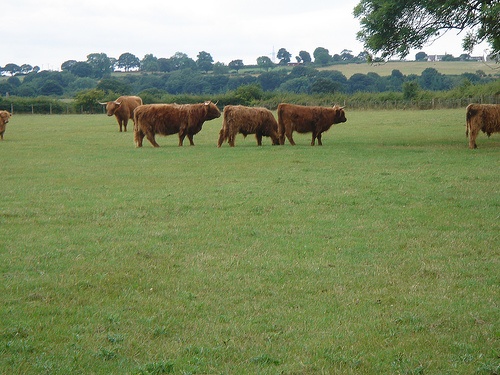}
               	}
               	\hspace{0cm}
               	\subfloat[]{\includegraphics[width=0.23   \linewidth,height=0.17  \linewidth]{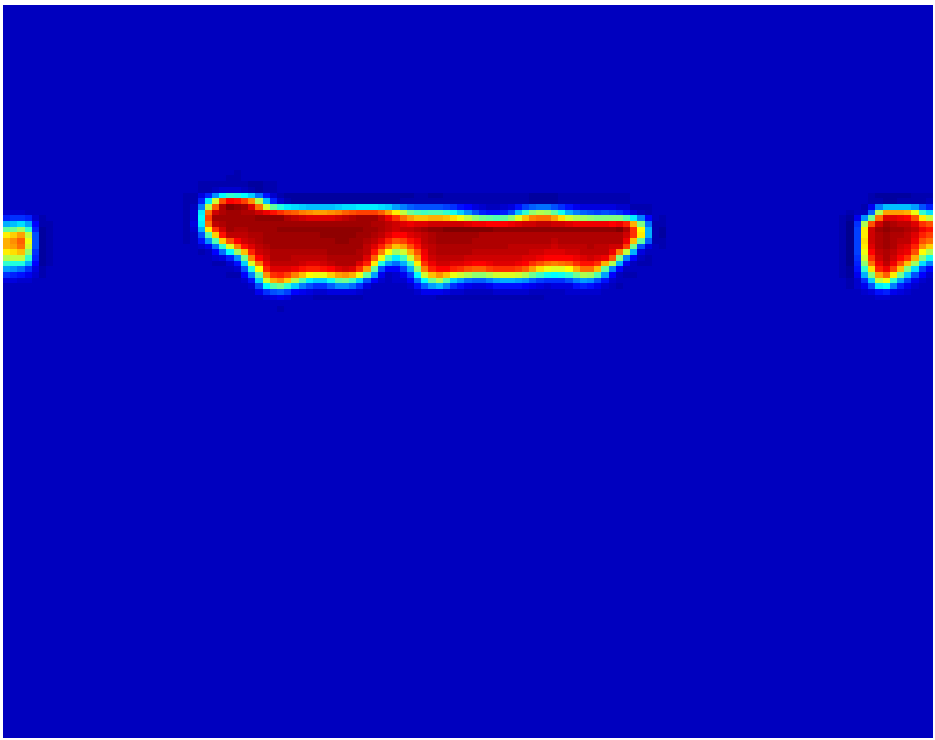}
               	} 
               	\hspace{0cm}
               	\subfloat[]{\includegraphics[width=0.23   \linewidth,height=0.17  \linewidth]{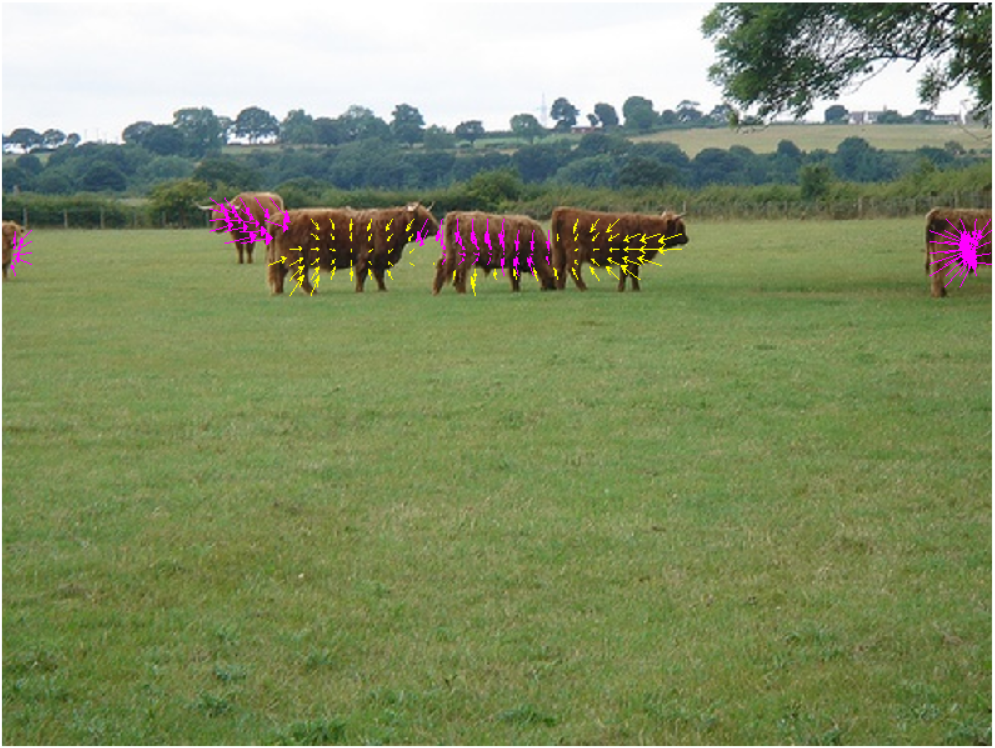}
               	}
               	\hspace{0cm}
               	\subfloat[]{\includegraphics[width=0.23   \linewidth,height=0.17  \linewidth]{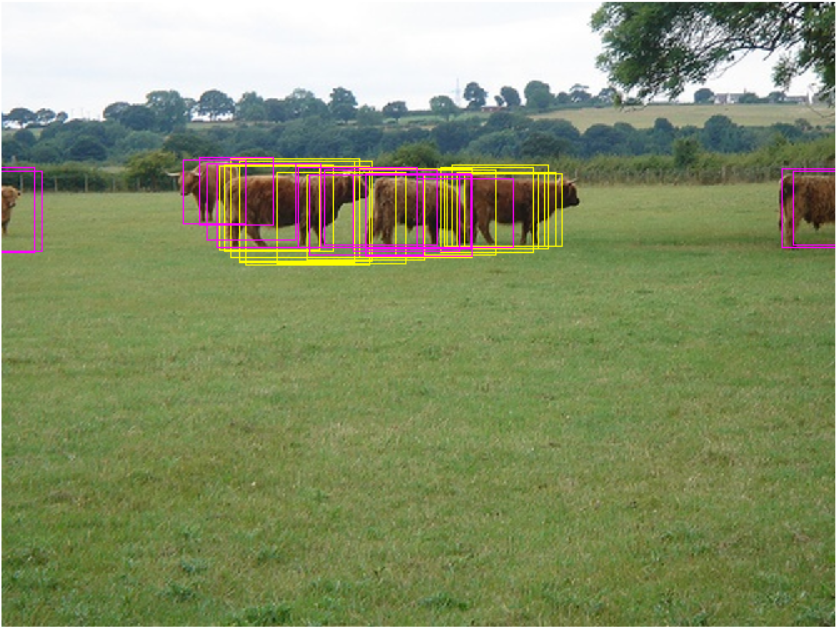}
               	} 
               	\\	
               	\vspace{-0.2cm}
               	\hspace{-0.1cm}
               	\subfloat[]{\includegraphics[width=0.23   \linewidth,height=0.17  \linewidth]{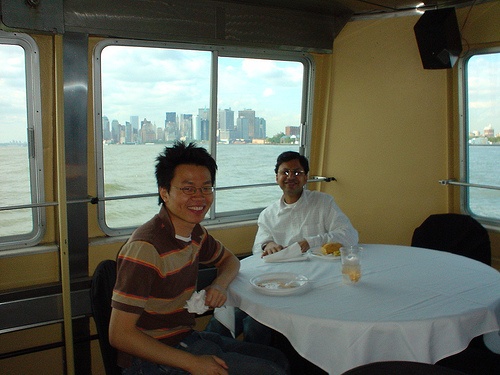}
               	}
               	\hspace{0cm}
               	\subfloat[]{\includegraphics[width=0.23   \linewidth,height=0.17  \linewidth]{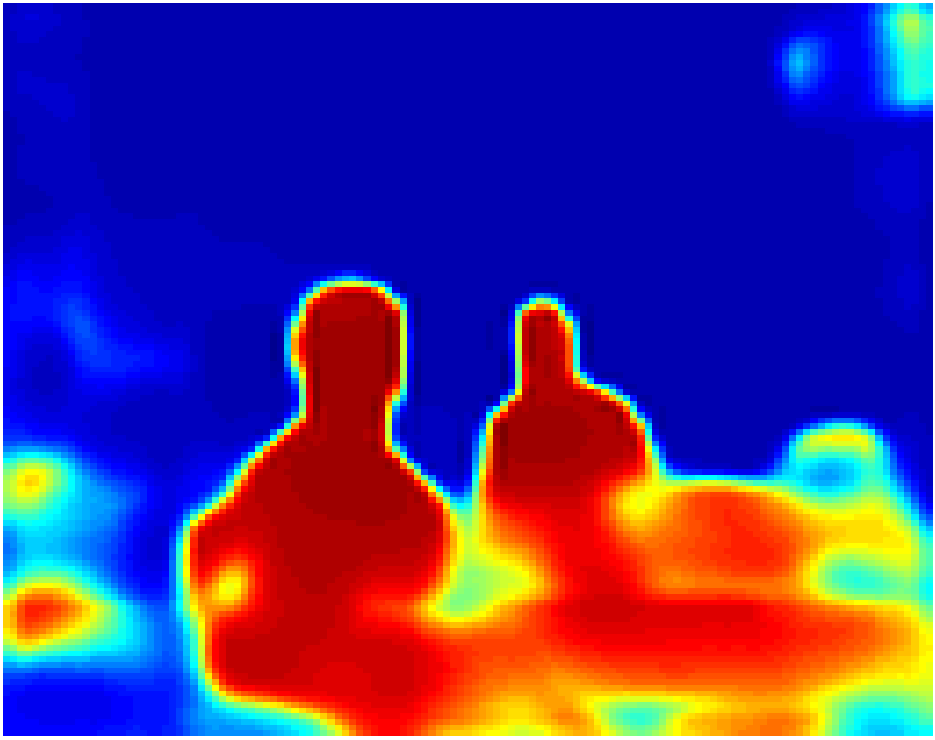}
               	} 
               	\hspace{0cm}
               	\subfloat[]{\includegraphics[width=0.23   \linewidth,height=0.17  \linewidth]{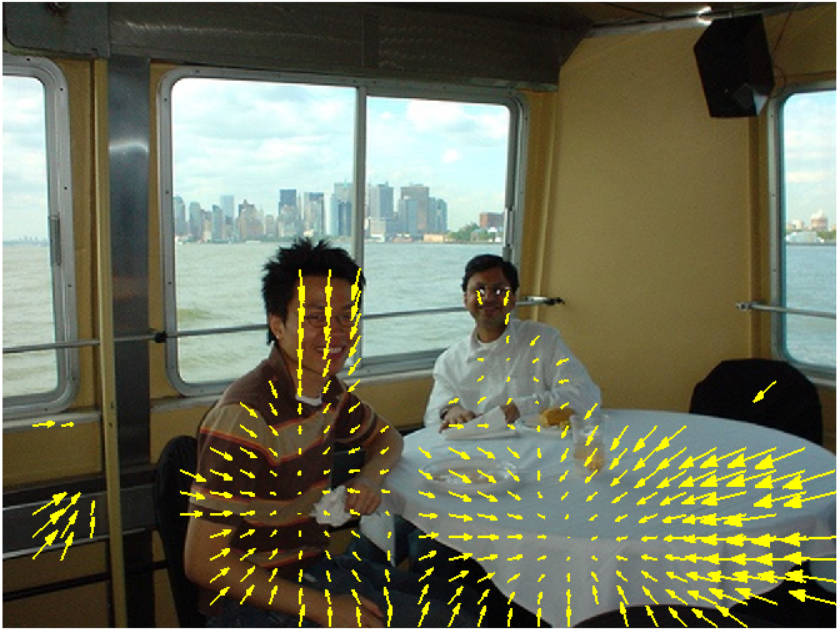}
               	}
               	\hspace{0cm}
               	\subfloat[]{\includegraphics[width=0.23   \linewidth,height=0.17  \linewidth]{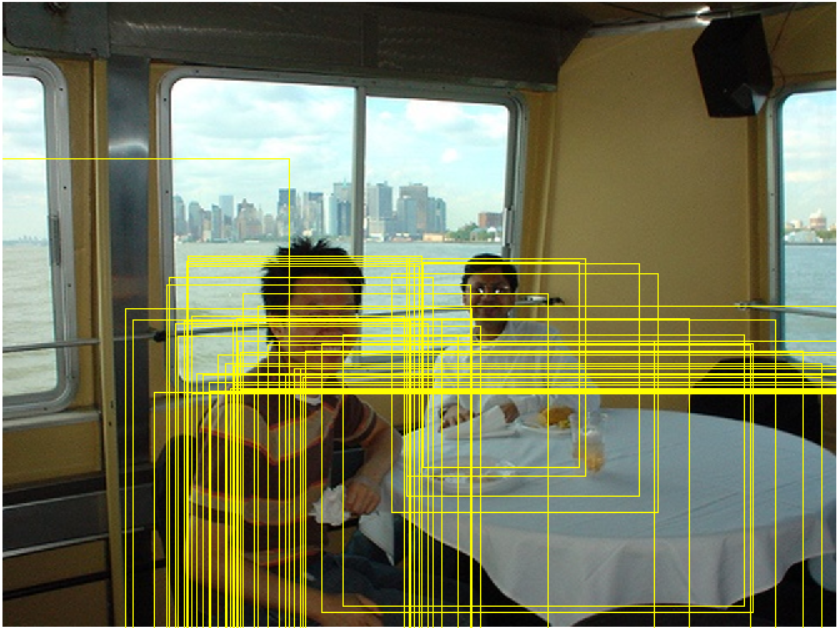}
               	} 
               	\\
               	\vspace{0.4cm}
               	\hspace{-0.1cm}
               	\subfloat[\normalsize Image]{\includegraphics[width=0.23   \linewidth,height=0.17  \linewidth]{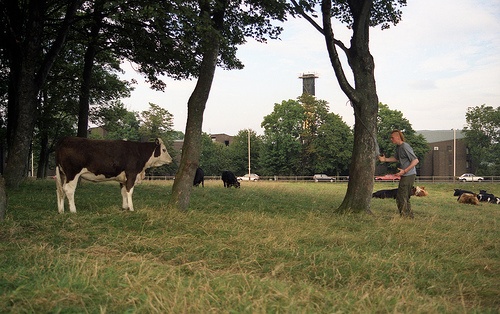}
               	}
               	\hspace{0cm}
               	\subfloat[\normalsize Objectness]{\includegraphics[width=0.23   \linewidth,height=0.17  \linewidth]{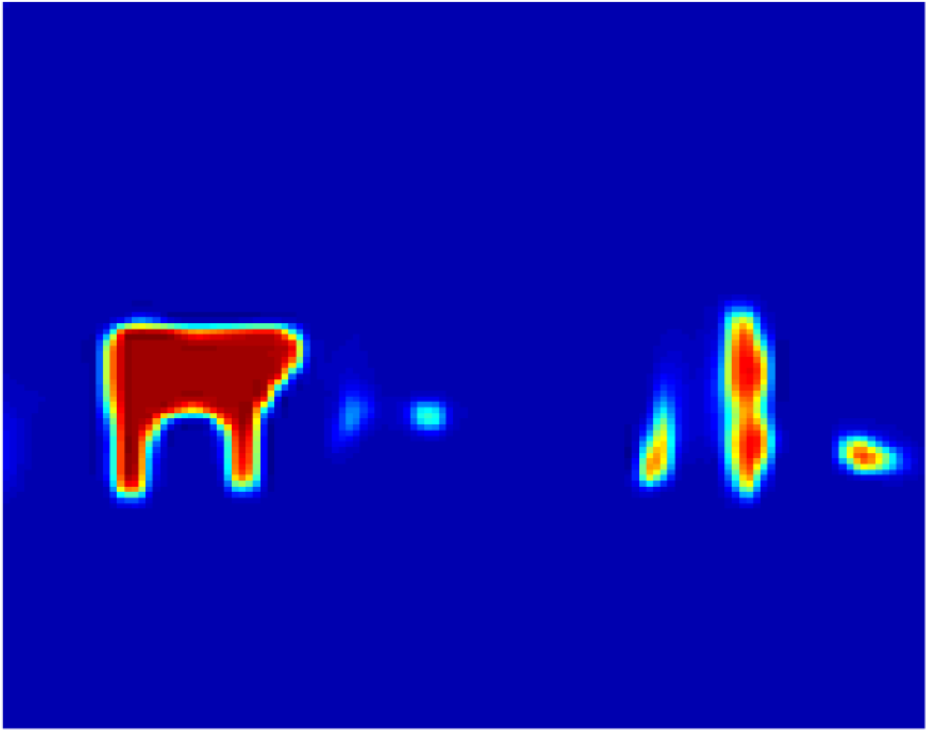}
               	} 
               	\hspace{0cm}
               	\subfloat[\normalsize Offsets to object center]{\includegraphics[width=0.23   \linewidth,height=0.17  \linewidth]{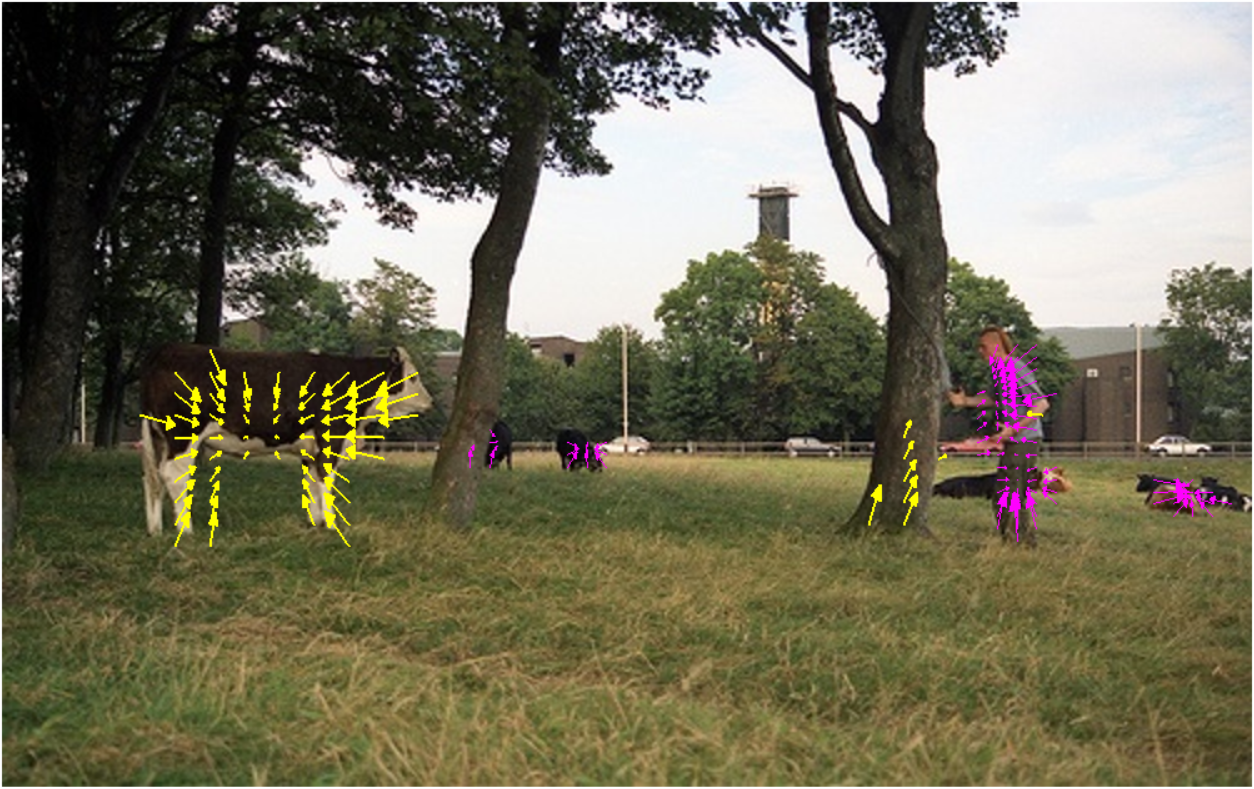}
               	}
               	\hspace{0cm}
               	\subfloat[\normalsize Object proposals]{\includegraphics[width=0.23   \linewidth,height=0.17  \linewidth]{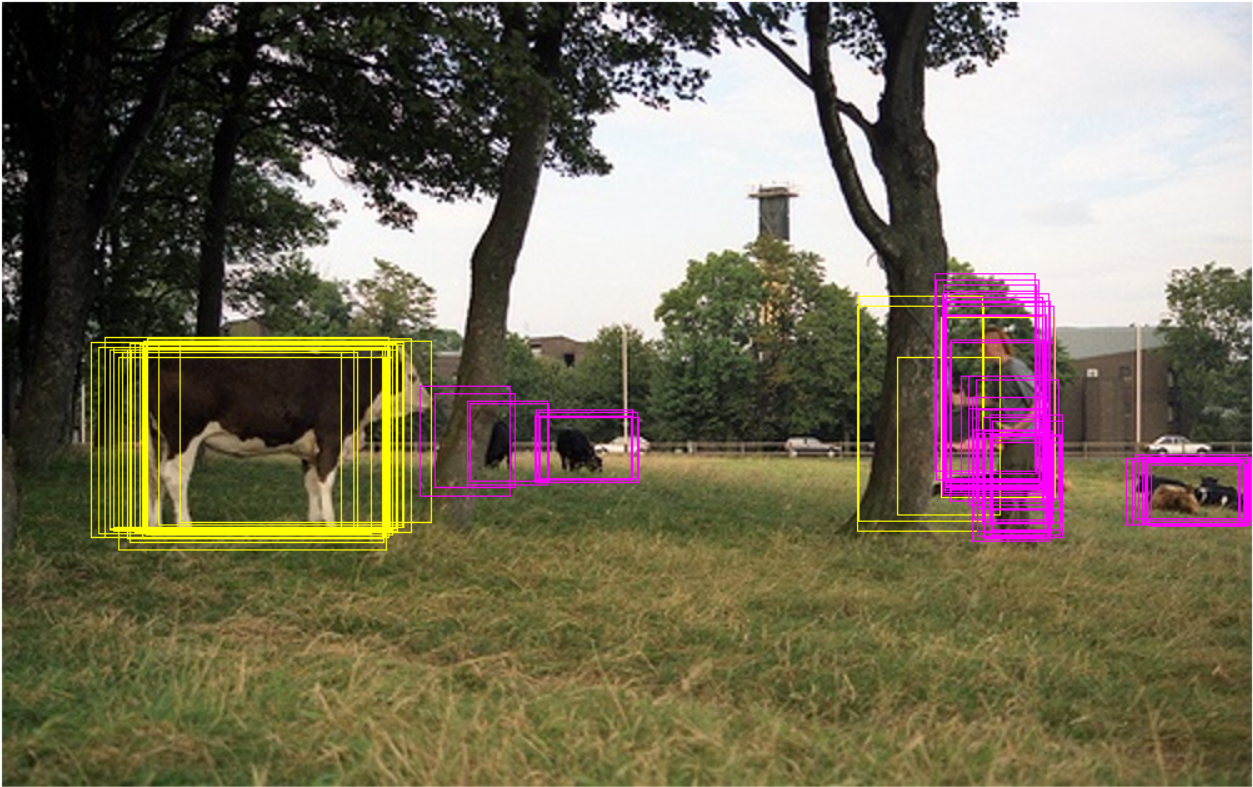}
               	} 
               	\\
               	\caption{Example results of predicted ``objectness map" (second column), ``offsets to object center" after weighted combination (third column) and ``object proposals" (fourth column) for the input images (first column).}  
               	\label{fig:visual}
               	\vspace{-4mm}
               \end{figure*}

  For better understanding of the keys of enabling the SPOP-net to work well, we show the intermediate output maps of both the localization and confidence networks for visualization in Figure~\ref{fig:visual}. For each image, we show its ``objectness confidence map",  ``offsets map" pointing to the object center, and its proposals. We argue that the first key is the reliable objectness prediction as the proposals predicted by the pixels obtaining low objectness confidence will be ranked behind. Based on an accurate objectness confidence, for each ground-truth object, each pixel inside it predicts its own location of this object, as shown in the ``offsets maps", thus greatly increasing the chances of precise localization. Another advantage of this pixel-wise prediction is that most of predicted bounding box locations from the pixels within the same object are heavily overlapping, which can be easily filtered by NMS. Normally only a few proposals are remained after NMS, thus improving the recall of the top-ranked proposals. For small objects, to overcome the inherent weakness that less chances are available to propose the correct locations, a scale-aware prediction is adopted by relying on an accurate estimation of the object size (\emph{i.e.} large or small) and combining the predictions of two networks.
   \begin{table}
   	\caption{Time cost of the state-of-the-arts and our method.}
   	\centering
   	\begin{tabular}{c|c}
   		& Time cost per image \\
   		\hline
   		BING& $0.01$s \\
   		\hline
   		Edge Boxes& $0.3$s \\
   		\hline
   		Geodesic& $1$s \\
   		\hline
   		MCG& $30$s \\
   		\hline
   		Objectness& $3$s \\
   		\hline
   		Selective Search& $10$s \\
   		\hline
   		RPN& $0.15$s \\
   		\hline
   		SPOP-net (ours)& $1.03$s \\
   	\end{tabular}
   	\label{tab: speed}
   \end{table}
  
  The detailed running speed of the SPOP-net as well as other state-of-the-art methods is presented in Table \ref{tab: speed}. The detailed setting of parameters for each method is as follows. We choose the single color space (\emph{i.e.} RGB) proposal computation for BING, and the "Fast" version for selective search. For the rest methods, we directly run their default codes.  As can be seen, \emph{window scoring methods} and \emph{CNN-based methods} are much faster than \emph{segment grouping methods}.  Inference for an image of PASCAL VOC size (\emph{e.g.} $300$*$500$) takes $1.03$s for our SPOP-net on a single TITAN X CPU. Specifically, testing one network of the original scale and the enlarged scale takes $0.11$s and $0.23$s on a single TITAN X GPU, respectively. However, as  the  computation within different CNNs of SPOP-net are independent of each other, training and testing SPOP-net can be accelerated by parallel computation over multiple GPUs. Although it is not one of the fastest object proposal methods (compared to BING, RPN and Edge Boxes), our approach is still competitive in speed among the proposal generators. We do, however, require use of the library Caffe~\cite{jia2014caffe} which is based on GPU computation for efficient inference like all CNN based methods. To further reduce the running time, some CNN speedup methods such as FFT, batch parallelization, or truncated SVD could be used in the future.

We also evaluate the proposed SPOP-net on MS COCO \cite{lin2014Microsoft} 2014 validation set and the results are shown in Figure \ref{fig:recall_coco}. The SPOP-net model here is trained on MS COCO training set which contains more than $80,000$ pixel-level annotated images. To conduct fair comparisons with the state-of-the-art segmentation annotations based approach, \emph{i.e.,} DeepMask, we only evaluate on the first $5,000$ images. Note that we  directly used the public DeepProposal model to extract proposals on MS COCO images. It is observed that DeepMask performs well, especially for the cases with low IoU thresholds (\emph{e.g.} $0.5$) and a few proposals (\emph{e.g.} $100$ proposals). The performance of the proposed SPOP-net gradually increases and SPOP-net demonstrates its superiority  as the number of proposals increases. Specifically, SPOP-net outperforms DeepMask in terms of recall at IoU $0.5$ (Figure \ref{fig:recall_coco}(d)), recall at IoU $0.7$ (Figure \ref{fig:recall_coco}(e)), ABO (Figure \ref{fig:recall_coco}(f)) and average recall (Figure \ref{fig:recall_coco}(g)) when the number of proposals is more than $500$. Figure \ref{fig:recall_coco}(h) and Figure \ref{fig:recall_coco}(i) shows the average recall of all the methods on large and small objects, respectively. On can observe that SPOP-net performs best on detecting small objects in terms of AR, which clearly validates the superiority of SPOP-net in small objects localization.
\begin{table*}
	\caption{Object detection average precision for all the $20$ categories as well as the mean average precision (mAP) on the PASCAL VOC 2007 testing set using the publicly available Fast-RCNN detector trained on VOC 2007 trainval set.}
	\footnotesize
	\renewcommand{\arraystretch}{2}
	\begin{tabular}{|p{2cm}<{\centering}|p{0.26cm}<{\centering}p{0.28cm}<{\centering}p{0.28cm}<{\centering}p{0.27cm}<{\centering}p{0.35cm}<{\centering}p{0.27cm}<{\centering}p{0.28cm}<{\centering}p{0.3cm}<{\centering}p{0.28cm}<{\centering}p{0.3cm}<{\centering}p{0.3cm}<{\centering}p{0.3cm}<{\centering}p{0.3cm}<{\centering}p{0.4cm}<{\centering}p{0.4cm}<{\centering}p{0.3cm}<{\centering}p{0.32cm}<{\centering}p{0.3cm}<{\centering}p{0.3cm}<{\centering}p{0.4cm}<{\centering}|p{0.5cm}<{\centering}|}
		\hline
		& aero& bike& bird& boat& bottle& bus& car& cat& chair& cow& table&
		dog& horse& mbike& person& plant& sheep& sofa& train& tv& mAP \\
		\hline
		Selective Search& $\mathbf{76.1}$& $77.3$& $65.3$& $53.9$& $37.8$& $76.9$& $78.2$& $80.9$& $40.6$& $74.0$& $67.2$& $79.4$& $82.4$& $74.9$& $66.1$& $33.3$& $66.0$& $67.3$& $73.3$& $67.3$& $66.9$ \\
		\hline
		Edge Boxes& $62.8$& $77.3$& $66.2$& $53.6$& $42.9$& $80.6$& $77.7$& $81.5$& $41.4$& $73.5$& $65.3$& $78.1$& $79.5$&  $\mathbf{76.2}$& $67.8$&  $\mathbf{36.7}$& $64.5$& $62.4$& $70.3$& $67.9$& $66.3$ \\
		\hline
		MCG & $69.3$& $72.3$& $62.4$& $54.4$& $39.2$& $77.8$& $70.1$& $80.4$& $40.1$& $67.2$&  $\mathbf{68.7}$& $77.3$& $75.0$& $68.8$& $60.7$& $34.1$& $59.5$& $64.7$& $70.6$& $68.2$& $64.0 $\\
		\hline
		\scriptsize RPN ($1{,}000$ props)& $70.0$& $76.6$& $67.2$& $59.1$& $44.6$& $80.0$& $78.6$& $86.2$& $44.1$& $75.5$& $60.7$& $81.3$& $80.4$& $75.8$& $74.1$& $30.5$& $72.9$& $67.2$& $79.4$& $69.1$& $68.7$ \\
		\hline
		\scriptsize RPN ($300$ props)& $71.8$& $77.4$& $68.0$& $58.9$&  $\mathbf{46.3}$& $81.8$& $79.0$& $86.6$& $45.6$&  $\mathbf{79.4}$& $60.2$& $81.7$& $81.1$& $75.9$&  $\mathbf{74.5}$& $31.7$&  $\textbf{73.6}$& $67.2$& $79.5$&  $\mathbf{70.6}$& $69.5$\\
		\hline
		\textbf{SPOP-net (ours)} & $70.6$&  $\mathbf{78.5}$&  $\mathbf{69.3}$&  $\mathbf{62.5}$& $41.1$&  $\mathbf{82.8}$&  $\mathbf{79.1}$&  $\mathbf{88.6}$&  $\mathbf{47.7}$& $76.6$& $66.5$&  $\mathbf{83.7}$&  $\mathbf{83.6}$& $73.1$& $69.6$& $36.1$& $67.8$&  $\mathbf{72.1}$&  $\mathbf{85.4}$& $69.6$&  $\mathbf{70.2}$ \\
		\hline
		
	\end{tabular}
	\label{tab: map}
\end{table*}
\subsection{Object Detection Performance}
We conduct experiments analyzing object proposals for use with object detectors to evaluate the effects of proposals on the detection quality. We utilize the standard Fast-RCNN~\cite{girshick2015fast} framework as the benchmark. We choose the publicly released VGG $16$-layer \cite{simonyan2014very} detector trained on VOC 2007 trainval set in all the evaluation experiments. The proposals of the proposed SPOP-net, Selective Search, Edge Boxes, MCG and RPN are evaluated. Please note that RPN itself integrates proposal generation and detection in a unified framework, called Faster-RCNN. To be fair, we do not adopt this unified detector for object detection with RPN proposals in our evaluation. This is because this unified detector has a weights sharing mechanism in $13$ layers which are used for both proposal generation and object detection. These layers are trained on the class-specific annotations with object category information that is not employed in training  other methods. For SPOP-net, Selective Search, Edge Boxes and MCG, we select the top $1{,}000$ proposals to pass through the object detectors for post-classification. For the RPN method, considering that it only needs a small number of proposals to achieve high recall, and more proposals do not bring too many improvements to the recall but introduce more false positives, we conduct an extra setting which uses the top $300$ proposals for detection, which is also claimed by~\cite{ren2015faster}. 
\begin{figure*}	
	\subfloat[Recall vs IoU (100 proposals) ]{\includegraphics[width=0.248\linewidth]{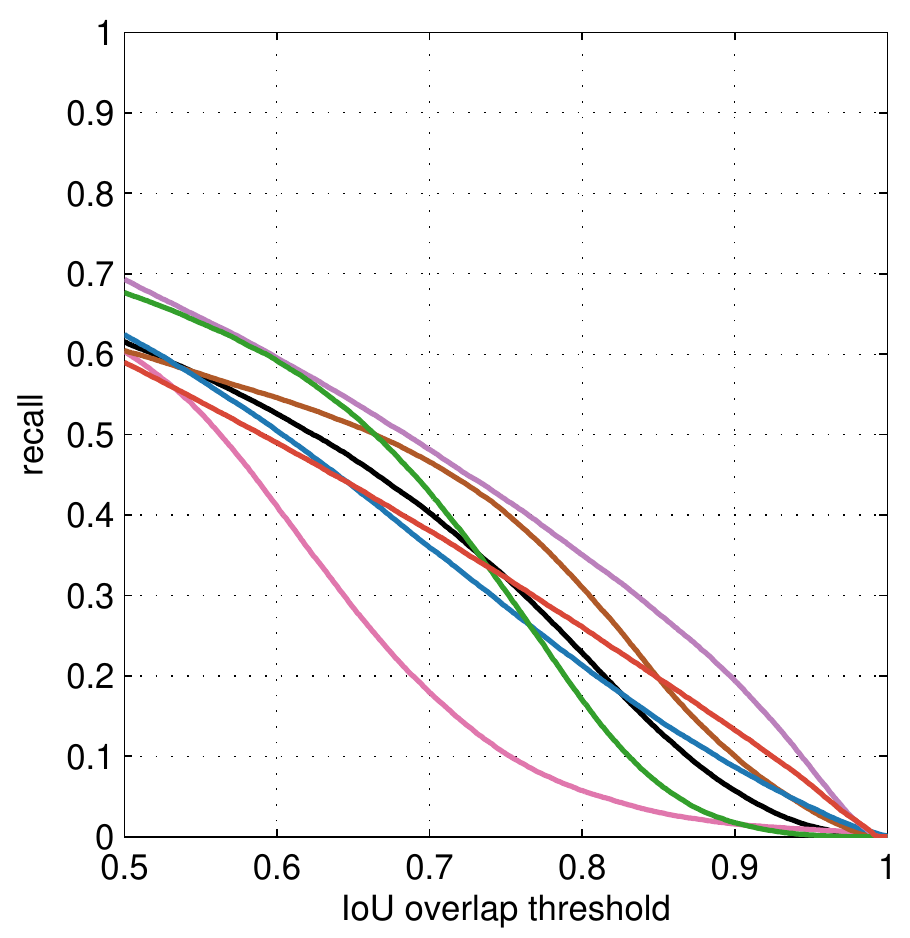}
	}
	\hspace{-0.3cm}
	\subfloat[Recall vs IoU (1000 proposals) ]{\includegraphics[width=0.248\linewidth]{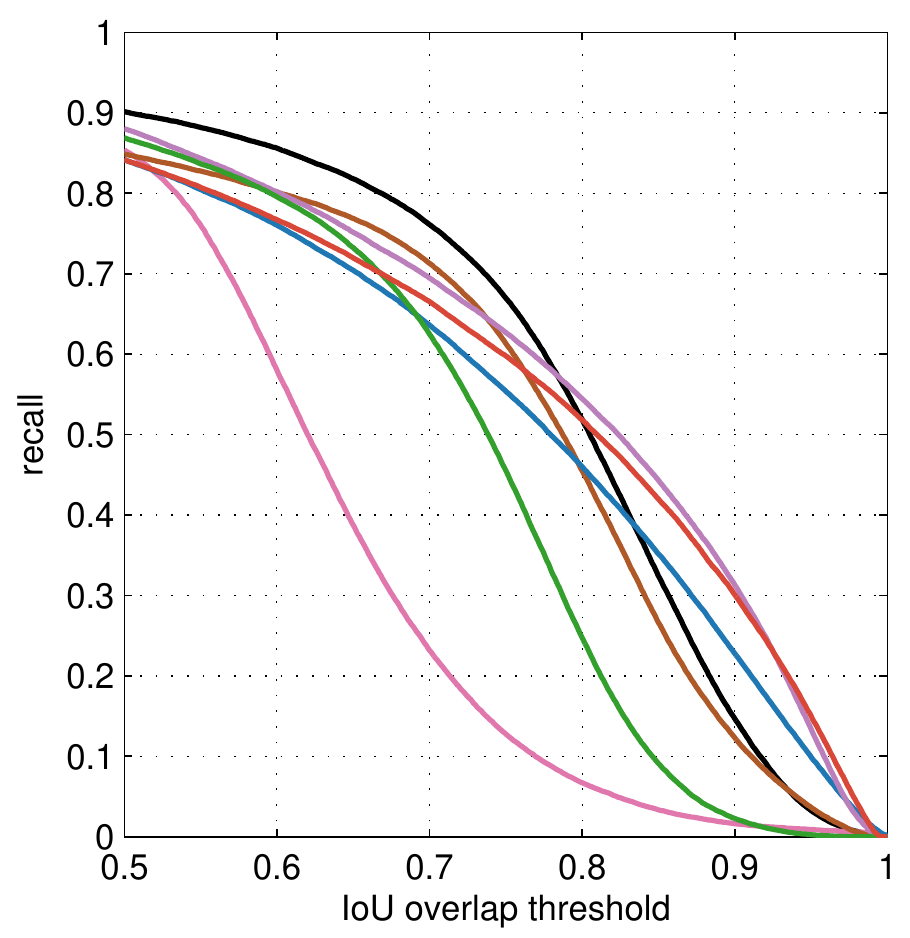}
	} 
	\hspace{-0.3cm}
	\subfloat[AR vs \# proposal (0.5$<$IoU$<$1) ]{\includegraphics[width=0.248\linewidth]{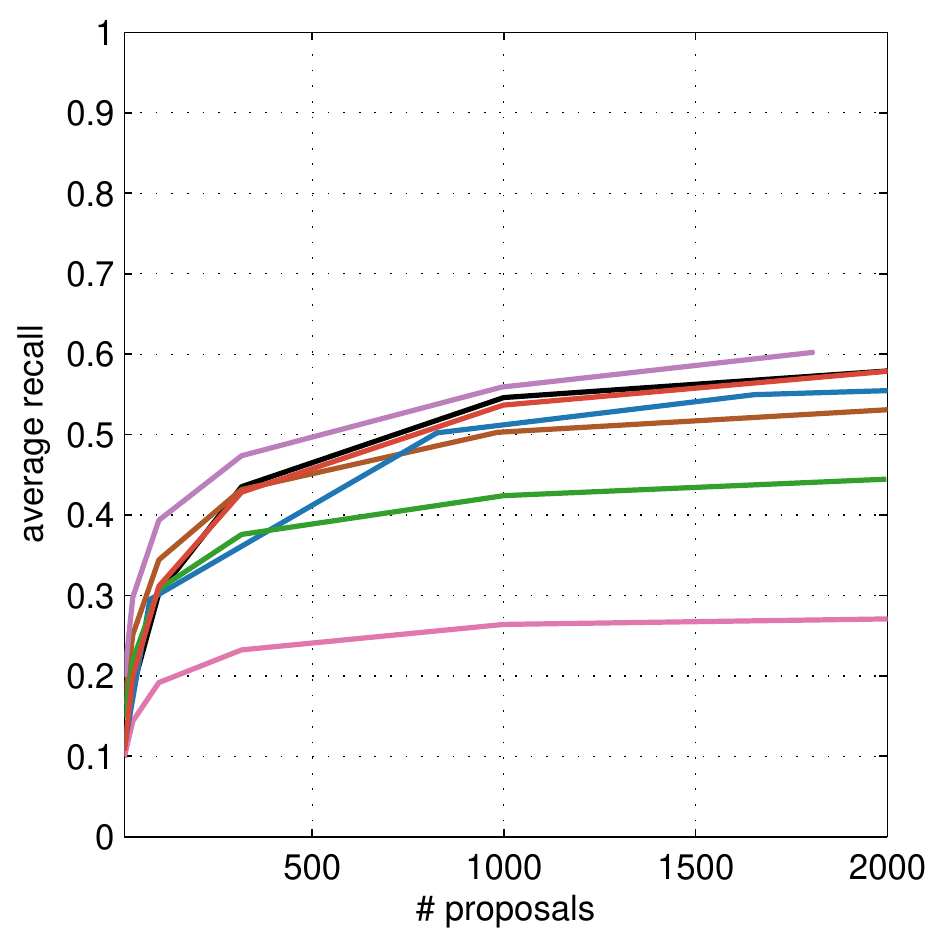}
	}
	\hspace{-0.3cm}
	\subfloat[ABO vs \# proposal  ]{\includegraphics[width=0.248\linewidth]{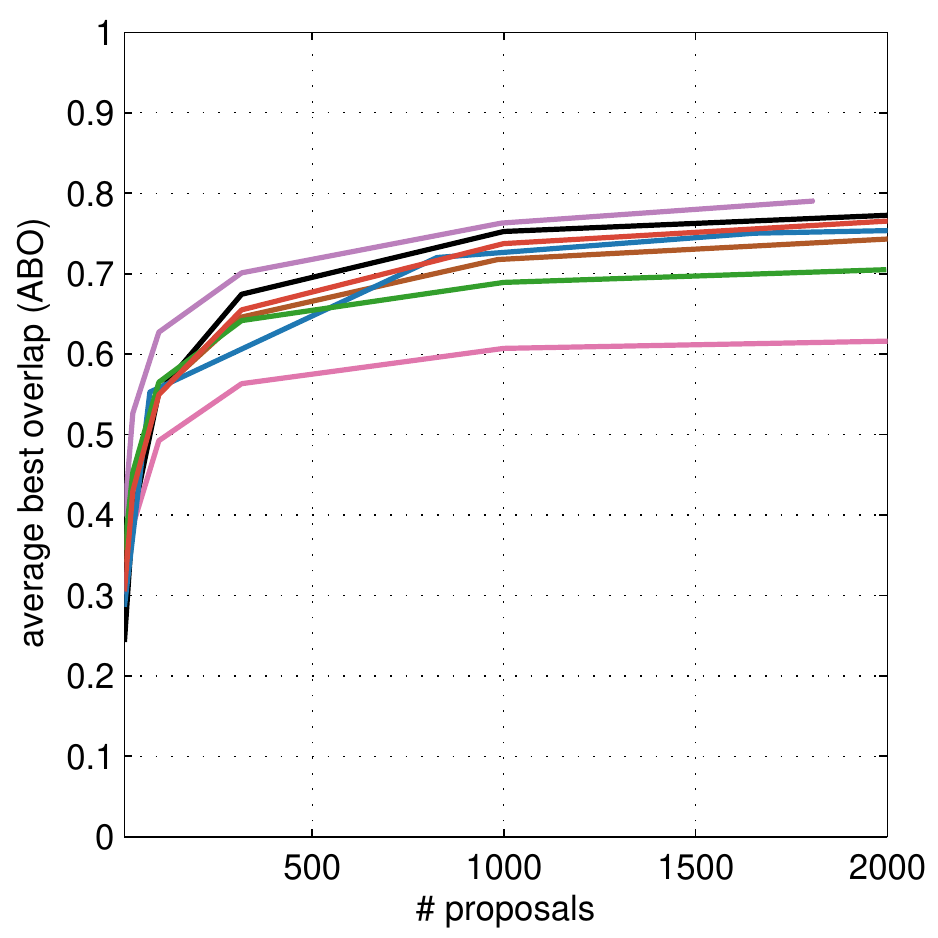}
	}
	\vspace{0.2cm}
	\\	
	\vspace{-0.2cm}
	\hspace{0.45cm}
	\subfloat{\includegraphics[width=0.94\linewidth]{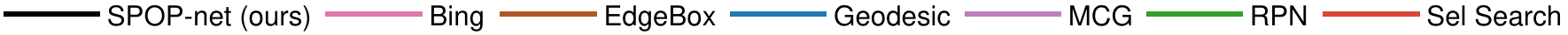}
	}
	\\    
	\caption{Recall and average best overlap (ABO) comparison between our SPOP-net and other state-of-the-art methods on ILSVRC 2013 validation set. }  
	\label{fig:recall_imagenet}
	\vspace{-0.2cm}
\end{figure*}

The detection mean average precision (mAP) as well as the average precision of $20$ categories is presented in Table \ref{tab: map}. It can be seen that the proposed SPOP-net wins on $11$ categories among the $20$ categories of PASCAL VOC 2007 and also achieves the best mAP $70.2\%$. Using $1{,}000$ RPN proposals for detection, $68.7\%$ mAP can be obtained. With only $300$ proposals, RPN achieves a better mAP $69.5\%$ than $1{,}000$ proposals. This verifies the good performance of RPN when generating a small number of proposals.

\subsection{Generalization to Unseen Categories}
The high recall rate which our approach achieves on the PASCAL VOC 2007 testing set does not guarantee it to have learned the generic objectness notion or be able to predict the object proposals for the images containing novel objects in unseen categories. This is because it is possible that the model is highly tuned to the $20$ categories of PASCAL VOC. To investigate whether it is capable of predicting the proposals for the unseen categories beyond training, we evaluate our approach on the ImageNet ILSVRC 2013 validation set which contains more than $20{,}000$ images with around $50{,}000$ annotated objects in $200$ categories. 

From Figure~\ref{fig:recall_imagenet}, the overall trend of the SPOP-net remains consistent with that on the PASCAL VOC 2007. Specifically, with a small number of proposals (\emph{e.g.} $100$ proposals), the SPOP-net does not perform as well as MCG, RPN and Edge Boxes, but shows its superiority when the number of proposals reaches $1{,}000$. See Figure \ref{fig:recall_imagenet}(b). As for average recall (AR) and average best overlap (ABO), the SPOP-net is also one of the best methods across a broad range of proposal numbers. It is worth mentioning that RPN does not perform as well as on PASCAL VOC 2007. An obvious drop is seen under all the evaluation scenarios from Figure~\ref{fig:recall_imagenet} compared to Figure~\ref{fig:recall_SOA}. This may result from the category information employed when training the  layers in the RPN network shared with class-specific detectors. Such class-awareness enables RPN to fit the $20$ categories of PASCAL VOC 2007 better but affects its generalization ability to unseen categories. 

Based on the high recall rate the SPOP-net remains when evaluated on ILSVRC 2013, no significant overfitting towards the PASCAL VOC categories is observed. In other words, the proposed approach has learned a generic notion of objectness and can generalize well to the unseen categories.

\section{Summary and Conclusions}
\label{sec:conclusion}
In this paper, we developed an effective scale-aware pixel-wise localization network for object proposal generation. The network fully exploits the available pixel-wise segmentation  annotations and predicts the proposals pixel-wisely. Each proposal combines two proposals predicted by two networks specialized for different sizes respectively. The combination follows a weighting mechanism utilizing the weighting confidence produced by a large-/small-size object classification model. This strategy is shown to enhance the accuracy of localization on small objects.  
Significant improvements over the state-of-the-art methods were achieved by the proposed SPOP-net  on the PASCAL VOC 2007 testing set.
 The proposals of the SPOP-net  used in Fast-RCNN detector also provide the highest mAP, benefiting from  the high recall rate of the proposed model.
 In the future, we plan to extend our method to deal with both object proposal generation and bounding box regression step to achieve better localization performance.

\ifCLASSOPTIONcaptionsoff
  \newpage
\fi

\bibliographystyle{IEEEtran}
\bibliography{IEEEabrv}

\begin{thebibliography}{10}
\providecommand{\url}[1]{#1}
\csname url@samestyle\endcsname
\providecommand{\newblock}{\relax}
\providecommand{\bibinfo}[2]{#2}
\providecommand{\BIBentrySTDinterwordspacing}{\spaceskip=0pt\relax}
\providecommand{\BIBentryALTinterwordstretchfactor}{4}
\providecommand{\BIBentryALTinterwordspacing}{\spaceskip=\fontdimen2\font plus
\BIBentryALTinterwordstretchfactor\fontdimen3\font minus
  \fontdimen4\font\relax}
\providecommand{\BIBforeignlanguage}[2]{{%
\expandafter\ifx\csname l@#1\endcsname\relax
\typeout{** WARNING: IEEEtran.bst: No hyphenation pattern has been}%
\typeout{** loaded for the language `#1'. Using the pattern for}%
\typeout{** the default language instead.}%
\else
\language=\csname l@#1\endcsname
\fi
#2}}
\providecommand{\BIBdecl}{\relax}
\BIBdecl

\bibitem{girshick2014rich}
R.~Girshick, J.~Donahue, T.~Darrell, and J.~Malik, ``Rich feature hierarchies
  for accurate object detection and semantic segmentation,'' in \emph{Computer
  Vision and Pattern Recognition (CVPR), 2014 IEEE Conference on}.\hskip 1em
  plus 0.5em minus 0.4em\relax IEEE, 2014, pp. 580--587.

\bibitem{he2014spatial}
K.~He, X.~Zhang, S.~Ren, and J.~Sun, ``Spatial pyramid pooling in deep
  convolutional networks for visual recognition,'' in \emph{Computer
  Vision--ECCV 2014}.\hskip 1em plus 0.5em minus 0.4em\relax Springer, 2014,
  pp. 346--361.

\bibitem{girshick2015fast}
R.~Girshick, ``Fast r-cnn,'' in \emph{Computer Vision (ICCV), 2015 IEEE
  International Conference on}.\hskip 1em plus 0.5em minus 0.4em\relax IEEE,
  2015, pp. 1440--1448.

\bibitem{felzenszwalb2010object}
P.~F. Felzenszwalb, R.~B. Girshick, D.~McAllester, and D.~Ramanan, ``Object
  detection with discriminatively trained part-based models,'' \emph{Pattern
  Analysis and Machine Intelligence, IEEE Transactions on}, vol.~32, no.~9, pp.
  1627--1645, 2010.

\bibitem{cheng2014bing}
M.-M. Cheng, Z.~Zhang, W.-Y. Lin, and P.~Torr, ``Bing: Binarized normed
  gradients for objectness estimation at 300fps,'' in \emph{Computer Vision and
  Pattern Recognition (CVPR), 2014 IEEE Conference on}.\hskip 1em plus 0.5em
  minus 0.4em\relax IEEE, 2014, pp. 3286--3293.

\bibitem{zitnick2014edge}
C.~L. Zitnick and P.~Doll{\'a}r, ``Edge boxes: Locating object proposals from
  edges,'' in \emph{Computer Vision--ECCV 2014}.\hskip 1em plus 0.5em minus
  0.4em\relax Springer, 2014, pp. 391--405.

\bibitem{alexe2010object}
B.~Alexe, T.~Deselaers, and V.~Ferrari, ``What is an object?'' in
  \emph{Computer Vision and Pattern Recognition (CVPR), 2010 IEEE Conference
  on}.\hskip 1em plus 0.5em minus 0.4em\relax IEEE, 2010, pp. 73--80.

\bibitem{uijlings2013selective}
J.~R. Uijlings, K.~E. van~de Sande, T.~Gevers, and A.~W. Smeulders, ``Selective
  search for object recognition,'' \emph{International journal of computer
  vision}, vol. 104, no.~2, pp. 154--171, 2013.

\bibitem{manen2013prime}
S.~Manen, M.~Guillaumin, and L.~Van~Gool, ``Prime object proposals with
  randomized prim's algorithm,'' in \emph{Computer Vision (ICCV), 2013 IEEE
  International Conference on}.\hskip 1em plus 0.5em minus 0.4em\relax IEEE,
  2013, pp. 2536--2543.

\bibitem{zhang2011proposal}
Z.~Zhang, J.~Warrell, and P.~H. Torr, ``Proposal generation for object
  detection using cascaded ranking svms,'' in \emph{Computer Vision and Pattern
  Recognition (CVPR), 2011 IEEE Conference on}.\hskip 1em plus 0.5em minus
  0.4em\relax IEEE, 2011, pp. 1497--1504.

\bibitem{arbelaez2014multiscale}
P.~Arbelaez, J.~Pont-Tuset, J.~Barron, F.~Marques, and J.~Malik, ``Multiscale
  combinatorial grouping,'' in \emph{Computer Vision and Pattern Recognition
  (CVPR), 2014 IEEE Conference on}.\hskip 1em plus 0.5em minus 0.4em\relax
  IEEE, 2014, pp. 328--335.

\bibitem{Erhan2013Scalable}
D.~Erhan, C.~Szegedy, A.~Toshev, and D.~Anguelov, ``Scalable object detection
  using deep neural networks,'' in \emph{Computer Vision and Pattern
  Recognition (CVPR), 2014 IEEE Conference on}.\hskip 1em plus 0.5em minus
  0.4em\relax IEEE, 2014, pp. 2147--2154.

\bibitem{ren2015faster}
S.~Ren, K.~He, R.~Girshick, and J.~Sun, ``Faster r-cnn: Towards real-time
  object detection with region proposal networks,'' in \emph{Advances in Neural
  Information Processing Systems}, 2015, pp. 91--99.

\bibitem{jieobject}
Z.~Jie, W.~F. Lu, S.~Sakhavi, Y.~Wei, E.~H.~F. Tay, and S.~Yan, ``Object
  proposal generation with fully convolutional networks,'' \emph{Circuits and
  Systems for Video Technology, IEEE Transactions on}, 2016.

\bibitem{everingham2014pascal}
M.~Everingham, S.~A. Eslami, L.~Van~Gool, C.~K. Williams, J.~Winn, and
  A.~Zisserman, ``The pascal visual object classes challenge: A
  retrospective,'' \emph{International Journal of Computer Vision}, vol. 111,
  no.~1, pp. 98--136, 2014.

\bibitem{russakovsky2014imagenet}
O.~Russakovsky, J.~Deng, H.~Su, J.~Krause, S.~Satheesh, S.~Ma, Z.~Huang,
  A.~Karpathy, A.~Khosla, M.~Bernstein \emph{et~al.}, ``Imagenet large scale
  visual recognition challenge,'' \emph{International Journal of Computer
  Vision}, vol. 115, no.~3, pp. 211--252, 2015.

\bibitem{chen2014semantic}
L.-C. Chen, G.~Papandreou, I.~Kokkinos, K.~Murphy, and A.~L. Yuille, ``Semantic
  image segmentation with deep convolutional nets and fully connected crfs,''
  in \emph{International Conference on Learning Representations}, 2015.

\bibitem{long2014fully}
J.~Long, E.~Shelhamer, and T.~Darrell, ``Fully convolutional networks for
  semantic segmentation,'' in \emph{Computer Vision and Pattern Recognition
  (CVPR), 2015 IEEE Conference on}, 2015, pp. 3431--3440.

\bibitem{liang2015proposal}
X.~Liang, Y.~Wei, X.~Shen, J.~Yang, L.~Lin, and S.~Yan, ``Proposal-free network
  for instance-level object segmentation,'' \emph{arXiv preprint
  arXiv:1509.02636}, 2015.

\bibitem{alexe2012measuring}
B.~Alexe, T.~Deselaers, and V.~Ferrari, ``Measuring the objectness of image
  windows,'' \emph{Pattern Analysis and Machine Intelligence, IEEE Transactions
  on}, vol.~34, no.~11, pp. 2189--2202, 2012.

\bibitem{krahenbuhl2014geodesic}
P.~Kr{\"a}henb{\"u}hl and V.~Koltun, ``Geodesic object proposals,'' in
  \emph{Computer Vision--ECCV 2014}.\hskip 1em plus 0.5em minus 0.4em\relax
  Springer, 2014, pp. 725--739.

\bibitem{kk-lpo-15}
P.~Kr{\"{a}}henb{\"{u}}hl and V.~Koltun, ``Learning to propose objects,'' in
  \emph{Computer Vision and Pattern Recognition (CVPR), 2015 IEEE Conference
  on}.\hskip 1em plus 0.5em minus 0.4em\relax IEEE, 2015, pp. 1574--1582.

\bibitem{krizhevsky2012imagenet}
A.~Krizhevsky, I.~Sutskever, and G.~E. Hinton, ``Imagenet classification with
  deep convolutional neural networks,'' in \emph{Advances in neural information
  processing systems}, 2012, pp. 1097--1105.

\bibitem{wei2015hcp}
Y.~Wei, W.~Xia, M.~Lin, J.~Huang, B.~Ni, J.~Dong, Y.~Zhao, and S.~Yan, ``Hcp: A
  flexible cnn framework for multi-label image classification,'' \emph{Pattern
  Analysis and Machine Intelligence, IEEE Transactions on}, 2016.

\bibitem{szegedy2014going}
C.~Szegedy, W.~Liu, Y.~Jia, P.~Sermanet, S.~Reed, D.~Anguelov, D.~Erhan,
  V.~Vanhoucke, and A.~Rabinovich, ``Going deeper with convolutions,'' in
  \emph{Computer Vision and Pattern Recognition (CVPR), 2015 IEEE Conference
  on}.\hskip 1em plus 0.5em minus 0.4em\relax IEEE, 2015, pp. 1--9.

\bibitem{liang2015towards}
X.~Liang, S.~Liu, Y.~Wei, L.~Liu, L.~Lin, and S.~Yan, ``Towards computational
  baby learning: A weakly-supervised approach for object detection,'' in
  \emph{Computer Vision (ICCV), 2015 IEEE International Conference on}.\hskip
  1em plus 0.5em minus 0.4em\relax IEEE, 2015, pp. 999--1007.

\bibitem{wei2015stc}
Y.~Wei, X.~Liang, Y.~Chen, X.~Shen, M.-M. Cheng, Y.~Zhao, and S.~Yan, ``Stc: A
  simple to complex framework for weakly-supervised semantic segmentation,''
  \emph{arXiv preprint arXiv:1509.03150}, 2015.

\bibitem{wei2016learning}
Y.~Wei, X.~Liang, Y.~Chen, Z.~Jie, Y.~Xiao, Y.~Zhao, and S.~Yan, ``Learning to
  segment with image-level annotations,'' \emph{Pattern Recognition}, 2016.

\bibitem{liang2015reversible}
X.~Liang, Y.~Wei, X.~Shen, Z.~Jie, J.~Feng, L.~Lin, and S.~Yan, ``Reversible
  recursive instance-level object segmentation,'' in \emph{Computer Vision and
  Pattern Recognition (CVPR), 2016 IEEE Conference on}.\hskip 1em plus 0.5em
  minus 0.4em\relax IEEE, 2016.

\bibitem{ghodrati2015deepproposal}
A.~Ghodrati, A.~Diba, M.~Pedersoli, T.~Tuytelaars, and L.~Van~Gool,
  ``Deepproposal: Hunting objects by cascading deep convolutional layers,'' in
  \emph{Computer Vision (ICCV), 2015 IEEE International Conference on}.\hskip
  1em plus 0.5em minus 0.4em\relax IEEE, 2015, pp. 2578--2586.

\bibitem{huang2015densebox}
L.~Huang, Y.~Yang, Y.~Deng, and Y.~Yu, ``Densebox: Unifying landmark
  localization with end to end object detection,'' \emph{arXiv preprint
  arXiv:1509.04874}, 2015.

\bibitem{zeiler2014visualizing}
M.~D. Zeiler and R.~Fergus, ``Visualizing and understanding convolutional
  networks,'' in \emph{Computer Vision--ECCV 2014}.\hskip 1em plus 0.5em minus
  0.4em\relax Springer, 2014, pp. 818--833.

\bibitem{Chen2015Improving}
X.~Chen, H.~Ma, X.~Wang, and Z.~Zhao, ``Improving object proposals with
  multi-thresholding straddling expansion,'' in \emph{Computer Vision and
  Pattern Recognition (CVPR), 2015 IEEE Conference on}.\hskip 1em plus 0.5em
  minus 0.4em\relax IEEE, 2015, pp. 2587--2595.

\bibitem{achanta2012slic}
R.~Achanta, A.~Shaji, K.~Smith, A.~Lucchi, P.~Fua, and S.~Susstrunk, ``Slic
  superpixels compared to state-of-the-art superpixel methods,'' \emph{Pattern
  Analysis and Machine Intelligence, IEEE Transactions on}, vol.~34, no.~11,
  pp. 2274--2282, 2012.

\bibitem{hariharan2011semantic}
B.~Hariharan, P.~Arbel{\'a}ez, L.~Bourdev, S.~Maji, and J.~Malik, ``Semantic
  contours from inverse detectors,'' in \emph{Computer Vision (ICCV), 2011 IEEE
  International Conference on}.\hskip 1em plus 0.5em minus 0.4em\relax IEEE,
  2011, pp. 991--998.

\bibitem{jia2014caffe}
Y.~Jia, E.~Shelhamer, J.~Donahue, S.~Karayev, J.~Long, R.~Girshick,
  S.~Guadarrama, and T.~Darrell, ``Caffe: Convolutional architecture for fast
  feature embedding,'' in \emph{ACM International Conference on
  Multimedia}.\hskip 1em plus 0.5em minus 0.4em\relax ACM, 2014, pp. 675--678.

\bibitem{Hosang2015arXiv}
J.~Hosang, R.~Benenson, P.~Doll\'ar, and B.~Schiele, ``What makes for effective
  detection proposals?'' \emph{Pattern Analysis and Machine Intelligence, IEEE
  Transactions on}, vol.~38, no.~4, pp. 814--830, 2016.

\bibitem{lin2014Microsoft}
T.-Y. Lin, M.~Maire, S.~Belongie, L.~Bourdev, R.~Girshick, J.~Hays, P.~Perona,
  D.~Ramanan, C.~L. Zitnick, and P.~Dollár, ``Microsoft coco: Common objects
  in context,'' in \emph{Computer Vision--ECCV 2014}.\hskip 1em plus 0.5em
  minus 0.4em\relax Springer, 2014, pp. 740--755.

\bibitem{simonyan2014very}
K.~Simonyan and A.~Zisserman, ``Very deep convolutional networks for
  large-scale image recognition,'' in \emph{International Conference on
  Learning Representations}, 2015.

\end{thebibliography}

\end{document}